\documentclass[logo]{moonmath}

\usepackage[numbers,sort&compress]{natbib}   
\usepackage[utf8]{inputenc}
\usepackage[T1]{fontenc}
\usepackage{hyperref}
\usepackage{url}
\usepackage{booktabs}
\usepackage{amsmath,amssymb}
\usepackage{graphicx}
\graphicspath{{../figures/}}
\usepackage{subcaption}
\usepackage{multirow}
\usepackage{xcolor}
\usepackage{colortbl}
\usepackage{array}
\usepackage{microtype}
\usepackage{enumitem}
\usepackage{float}       % [H] placement
\usepackage[section]{placeins}  % FloatBarrier at every \section boundary
\newcommand{\worldjen}{\textsc{WorldJen}}
\newcommand{\veo}{Veo~3.1~Fast}
\newcommand{\kling}{Kling~v2.6~Pro}
\newcommand{\ltx}{LTX-2}
\newcommand{\wanb}{Wan~v2.2~A14B}
\newcommand{\hunyuan}{Hunyuan~v1.5}
\newcommand{\wans}{Wan~2.1~1.3B}

\title{\worldjen: An End-to-End Multi-Dimensional Benchmark\\
for Generative Video Models}

\author{%
  Karthik Inbasekar\thanks{Correspondence: \texttt{karthik@moonmath.ai}} \quad
  Guy Rom \quad Omer Shlomovits\\
  \normalsize moonmath.ai
}

\begin{document}

% ─────────────────────────────────────────────────────────────────────────────
\begin{abstract}
Evaluating generative video models remains an open problem. Reference-based metrics
such as Structural Similarity Index Measure (SSIM) and Peak Signal to Noise Ratio (PSNR) 
reward pixel fidelity over semantic correctness, while Frechet Video Distance (FVD)
favors distributional textures over physical plausibility. Binary Visual Question Answering (VQA)
based benchmarks like VBench~2.0 are prone to yes-bias and rely on low-resolution auditors that
miss temporal failures. Moreover, their prompts target a single dimension at a time, multiplying
the number of videos required while still not guaranteeing reliable results.

\worldjen\ addresses these limitations directly. Binary VQA is replaced with Likert-scale
questionnaires graded by a VLM that receives frames at native video resolution. Video generation costs are addressed by using 
adversarially curated prompts that are designed to exercise up to 16 quality dimensions simultaneously. The framework is built around two interlocking contributions.
First, A \textbf{blind human preference study} is conducted, accumulating (2{,}696 pairwise annotations from 7 annotators with 100\% pair coverage over 50 of the curated prompts $\times$ 6 state-of-the-art video models. A mean inter-annotator agreement of 66.9\% is achieved and the study establishes a human ground-truth Bradley-Terry (BT) rating with a three-tier structure.
Second, A \textbf{VLM-as-a-judge evaluation engine} using prompt-specific,
dimension-specific Likert questionnaires (10 questions per dimension,
47{,}160 scored responses) judges the videos and reproduces the
human-established three-tier BT rating structure independently. The VLM achieves a
Spearman $\hat{\rho}=1.000,~p=0.0014$ that is interpreted as tier agreement with the human results. Six focused ablation studies validate the robustness of the VLM evaluation framework. The project page, including video examples and an interactive scoring playground, is available at \url{https://moonmath.ai/worldjen/}.
\end{abstract}

\maketitle

% ─────────────────────────────────────────────────────────────────────────────
%\FloatBarrier
\section{Introduction}
\label{sec:intro}
% ─────────────────────────────────────────────────────────────────────────────

Generative video models have undergone a remarkable transformation. In just a few years,
the field has moved from academic prototypes producing a handful of frames to commercial
systems capable of generating minutes of photorealistic, temporally coherent footage on
demand~\citep{veo2025,kling2024,wan2025}. This rapid progress raises a fundamental
question: \emph{what does ``better'' mean for a generative video model, and how can we
measure it reliably?}

Traditional computer vision employed reference-based metrics that compare a generated video
against a ground truth video. SSIM~\citep{wang2004ssim} and PSNR measure pixel-level structural
similarity and signal fidelity, respectively. Both these metrics are gameable by generating blurry but
mathematically close pixels, and neither can detect logic, reasoning, physics, semantic adherence
and other more complex features. LPIPS~\citep{zhang2018lpips} operates on learned perceptual features 
on a per-frame basis, while it is an improvement for capturing human
perception, it still misses the temporal components intrinsic to video generation.
FVD~\citep{unterthiner2019fvd} addresses temporality by
comparing the statistical distribution of generated clips to real-world video embeddings,
but this biases style over substance, rewarding texture that resembles training data while
ignoring physical implausibility.

Benchmarks built specifically for video generation take a different approach. VBench~\citep{huang2024vbench}
pioneered the decomposition of quality into dimensions using task-specific metrics such as
DINO~\citep{caron2021dino} and CLIP~\citep{radford2021clip} for subject/background
feature similarity, RAFT~\citep{teed2020raft} optical flow for motion
smoothness, GRiT~\citep{wu2022grit} for object-class detection, and
Tag2Text~\citep{huang2023tag2text} for scene tagging.
VBench++~\citep{vbenchpp_2024} extended this to image-to-video (i2v) generation and trustworthiness
evaluation while retaining the same methodology. However, these dimension-specific tools rely on fixed shallow features and cannot assess
higher-level semantic faithfulness or physical plausibility.
VBench~2.0~\citep{vbench2_2025} represents significant advancement by replacing specialist
detectors with a generalist VLM (LLaVA-Video-7B) that performs binary Visual Question
Answering (VQA) on each video. Nevertheless, four limitations remain:
\begin{itemize}[leftmargin=1.5em,topsep=2pt,itemsep=0pt]
  \item VBench's underlying metrics (DINO, CLIP, RAFT) explicitly resize frames to
    $224{\times}224$ pixels before feature extraction, a resolution at which
    fine-grained temporal artifacts, physics violations, and subtle subject drift
    fall below the pixel threshold and are averaged away. This produces noisy scores, with poor resolution.
  \item Binary VQA suffers from yes-bias, when asked ``Is the physics correct?'' a VLM
    tends to answer ``Yes'' unless failure is catastrophic~\citep{zheng2023mtbench},
    lacking the granularity to detect subtle violations or distinguish degrees of quality. 
    Score compression follows directly as binary outputs collapse nuanced quality
    differences into pass/fail judgements, producing near-ceiling scores across models
    and eliminating the discrimination needed to rank them reliably. This observation forced us to rethink to a granular scoring mechanism. 
  \item Prompts are crafted as narrow specialists for individual dimensions, failing to
    reflect the complexity of real user queries, which typically overlap across multiple
    quality axes simultaneously, and while models sometimes pass these individual dimension tests, their performance in complex situations are unpredictable.
  \item A more pressing problem is evaluation cost and scalability.
    A full VBench run requires generating \textbf{6,230 videos} across 946 prompts
    ($~$5 videos per prompt, 25 for temporal flickering due to static-scene filtering),
    while VBench~2.0 requires \textbf{3,209 videos} across 1,013 prompts
    (3 per prompt and 20 for the Diversity dimension).
    Evaluating a single new model therefore demands thousands of video generations
    before a single score is produced, creating a prohibitive barrier for rapid
    iterative model development. 
\end{itemize}

Inspired by Goodhart's Law~\citep{goodhart1975}: ``when a measure becomes a target, it
ceases to be a good measure'' we design \worldjen\ video benchmarking framework to resist gaming through full resolution video processing, Likert-scale granularity, and complex prompts designed to test multiple benchmarking dimensions, while simultaneously
reducing number of videos required for the scoring system. The main contributions of this work are
\begin{enumerate}[leftmargin=1.5em,topsep=2pt,itemsep=2pt]
\item \textbf{Prompt curation.} A corpus of 3{,}754 human-authored prompts filtered
    from VidProM~\citep{vidprom2024} and judged on 16 quality dimensions, yielding
    prompts with wide coverage and complex cross-dimension interactions. The prompts are also LLM enhanced without changing the core theme of the prompt. 
 \item \textbf{Human evals.} A blind pairwise preference study
    across 750 pairs consisting of videos generated by 6 state-of-the-art video generation models from 50 curated prompts. The study accumulated 2{,}696 weighted annotations
    from 7 annotators achieving 100\% pair coverage, and 66.9\% mean inter-annotator agreement. The study establishes a human ground-truth Bradley-Terry
    rating \cite{bradley1952} with a clear \textbf{three-tier structure} and a statistically unambiguous separation between tiers. 
 \item \textbf{Prompt-specific Likert questionnaires for VLM evaluation:} For each prompt 10
    evaluation questions per dimension are generated via a structured LLM pipeline, ensuring every question targets features specific to the given prompt and dimension, with a 1--5 Likert rubric covering expected events, failure modes, success modes, and adversarial probes.
  \item \textbf{VLM evaluation engine.} Videos are evaluated by a VLM judge \texttt{gemini-3-flash-preview} at full video resolution using the prompt-specific, dimension-specific
    Likert questionnaires. Question-level scores (1--5) feed the BT model to produce
    stable BT ratings that agree with human ground-truth on all three statistical tiers achieving a Spearman $\hat{\rho}=1.000,~p=0.0014$, with clear tier separation. 
\end{enumerate}
\noindent\textbf{Open source.}
To support reproducibility and community use, the source code, prompts, evaluation
questionnaires, VLM scoring results, the dataset of
420 generated videos, and the anonymized human annotation dataset
(2{,}696 pairwise preference votes with confidence labels) are released. \footnote{
\url{https://github.com/moonmath-ai/WorldJen-benchmarking-subsystem}}\footnote{
\url{https://huggingface.co/datasets/ik6626/WorldJen-benchmarking-subsystem}}

\begin{figure}[tbp]
  \centering
  \includegraphics[width=0.55\linewidth]{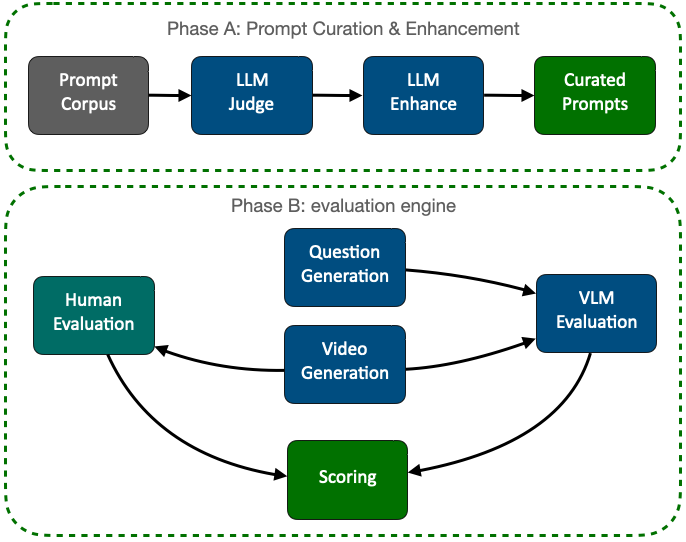}
  \caption{\worldjen\ framework overview. Phase~A is standard, curates and enhances prompts Appendix~\S~\ref{app:prompt_curation};
    Phase~B runs the VLM evaluation engine producing BT ratings and 
    PHAS (Predicted Human Alignment Score) rankings introduced in this paper.}
  \label{fig:overview}
\end{figure}

Figure~\ref{fig:overview}, gives a high-level view of the two-phase pipeline used in this work. Prompt curation (Phase~A) follows a standard multi-dimensional filtering and scoring procedure, it mostly serves as background material, readers interested in the full details can find them in Appendix~\S~\ref{app:prompt_curation}.
The main body of the paper focuses on the two primary contributions in Phase B,
the human preference study \S~\ref{sec:human_study} and the VLM evaluation engine \S~\ref{sec:vlm_eval}.

The paper is structured as follows.
\S~\ref{sec:related} reviews closely related work.
\S~\ref{sec:framework} details the \worldjen\ framework.
\S~\ref{sec:shared_setup} details the datasets used for the subsequent experiment sections.
\S~\ref{sec:human_study} reports the human preference study that establishes
ground-truth BT ratings and the three tier structure.
\S~\ref{sec:vlm_eval} describes the \worldjen\ VLM evaluation engine and shows that it reproduces human preferences exactly. 
\S~\ref{sec:phas} presents a Predicted Human Alignment Score, which on caliberation with the human results, is 
capable of reproducing the human/VLM BT rankings with far fewer prompts.
Core ablation studies \S~\ref{sec:ablation} validate framework robustness through VLM auditor cross validation \S~\ref{sec:abl_cross_vlm}, VLM run reliability/hallucinations and \S~\ref{sec:abl_reliability}, open source VLM alternatives \S~\ref{sec:abl_gemma4}. Comparison of human preferences with VBench and the \worldjen\ framework is reported in \S~\ref{sec:vbench}. 
\S~\ref{sec:limitations} discusses limitations and future work.
\S~\ref{sec:conclusion} concludes. The appendices \S~\ref{app:human} details the human annotation protocol, \S~\ref{app:prompt_curation} discusses prompt curation  and \S~\ref{app:vlm} the VLM questionaire generation process. \S~\ref{app:casestudies} presents two explicit examples of the pipeline. \S~\ref{sec:abl_enhance} discusses prompt enhancement effects on the VLM. Effects of number of questions and number of prompts are reported in \S~\ref{app:abl_qcount} and \S~\ref{app:abl_nprompts} respectively.
% ─────────────────────────────────────────────────────────────────────────────
%
\section{Related Work}
\label{sec:related}
% ─────────────────────────────────────────────────────────────────────────────
% A concrete illustration of this methodological gap is as follows, Wan~2.1~1.3B scores near the
% top of the VBench leaderboard~\citep{vbench_leaderboard} on \emph{Motion Smoothness}
% (0.97) and \emph{Temporal Flickering} (0.99). Note that the metrics are computed via RAFT optical
% flow magnitude and CLIP inter-frame cosine similarity, which both reward low-motion,
% perceptually stable outputs.\footnote{VBench leaderboard scores are for the
% \texttt{Wan2.1-T2V-1.3B} entry as accessed March~2026.}
% A model can therefore maximise these scores by generating near-static scenes, which
% satisfies the metric while failing to satisfy prompt intent. This is a typical problem with
% RAFT/CLIP which often fail to distinguish between intentional motion and unintentional noise.
% In particular they capture \emph{pixel stability} i.e unconditional, whereas a
% prompt-conditioned VLM Likert score captures \emph{motion fidelity to the requested
% action} i.e conditional, a distinction that becomes critical when prompts explicitly demand
% complex multi-step dynamics. This gap motivates our prompt enhancement pipeline and the use of VLM-based
% evaluation.

Some closely related works are: 
EvalCrafter~\citep{liu2024evalcrafter} provides a 700-prompt benchmark with human quality
annotations. T2V-CompBench~\citep{sun2024t2vcompbench} targets compositional evaluation.
VideoPhy~\citep{bansal2024videophy} specifically evaluates physical commonsense in video
generation, a dimension we also identify as the primary current gap. VBVR~\citep{wang2026vbvr} introduces a large-scale video reasoning dataset spanning 200
curated reasoning tasks and over one million video clips, with a verifiable evaluation
framework using rule-based, human-aligned scorers. VQQA~\citep{song2026vqqa} proposes a closely related multi-agent framework that
dynamically generates visual questions and uses VLM critiques as \emph{semantic gradients}
to iteratively refine video generation prompts at test time, achieving +11.57\% on
T2V-CompBench over vanilla generation. While VQQA and \worldjen\ share the core insight of dynamic VQA question generation and
VLM-based scoring, their objectives are orthogonal. VQQA is a closed-loop
\emph{generation optimizer} that improves a single model's output across iterations,
whereas \worldjen\ is an \emph{evaluation benchmark} that comparatively ranks multiple
models using enhanced prompts across 16
fine-grained dimensions. 

\paragraph{LLM/VLM-as-a-judge.}
MT-Bench~\citep{zheng2023mtbench} and Chatbot Arena~\citep{chiang2024chatbot} pioneered
the use of LLMs as evaluators and BT-rating-based rankings for language models.
AlpacaEval~\citep{li2023alpacaeval} and WildBench~\citep{lin2024wildbench} extended this
framework to instruction following. \worldjen\ applies the the VLM-as-a-judge philosophy to \emph{comparative benchmarking},
combining prompt-specific Likert-scale questionnaires with BT ratings and a
human preference anchor across 16 video benchmarking dimensions.
% ─────────────────────────────────────────────────────────────────────────────
%\FloatBarrier
\section{The \worldjen\ Framework}
\label{sec:framework}
% ─────────────────────────────────────────────────────────────────────────────

\begin{figure}[tbp]
  \centering
  \includegraphics[width=0.65\linewidth]{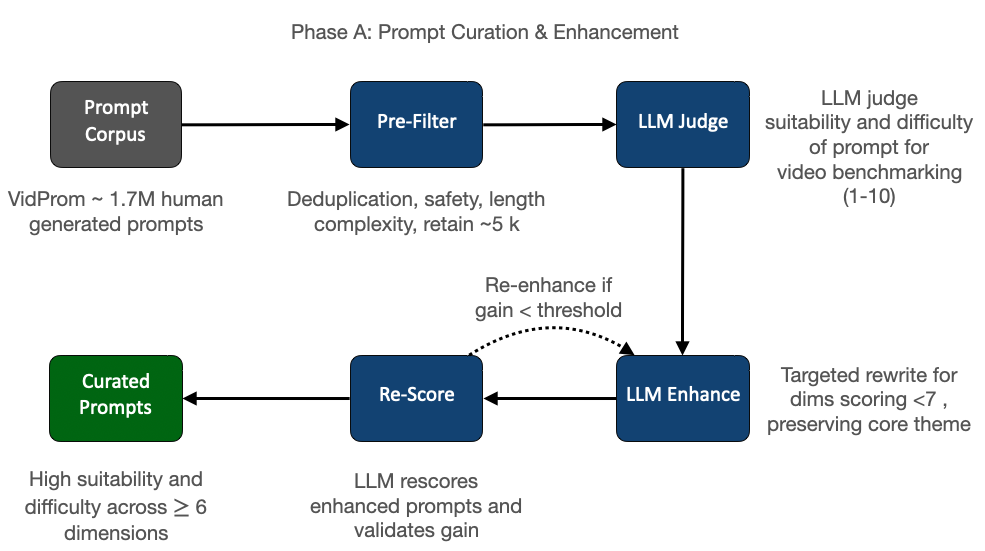}
  \caption{Phase~A detailed pipeline. 
    The dashed feedback arrow indicates re-enhancement when the rescore gain falls
    below threshold.}
  \label{fig:phase_a}
\end{figure}

\begin{figure}[tbp]
  \centering
  \includegraphics[width=0.55\linewidth]{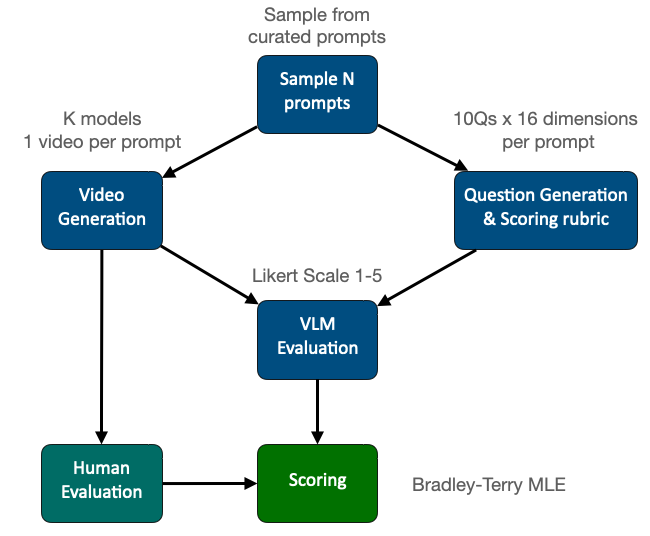}
  \caption{Phase~B evaluation engine. The curated prompt pool fans out
    in parallel into prompt-specific question generation and video generation.
    Both outputs feed the VLM Evaluator; scores aggregate into Bradley-Terry ratings combining into a final leaderboard.}
  \label{fig:phase_b}
\end{figure}

A high-level view of the two-phase pipeline used in this work is provided in Figure~\ref{fig:overview}. Phase~A focuses on prompt curation, it produces a reusable, multi-dimensionally enriched 
prompt pool. Phase~B takes that pool together with any set of video models and produces
scored, ranked results using the benchmarking system.
Figures~\ref{fig:phase_a} and~\ref{fig:phase_b} detail each phase respectively.

% ══════════════════════════════════════════════════════════════════════════════
\subsection{Phase A: Prompt Curation}
\label{sec:phase_a}
% ══════════════════════════════════════════════════════════════════════════════

\subsubsection{Prompt Curation}
\label{sec:curation}

\paragraph{Source and pre-filtering.}
Prompts are sourced from VidProM~\citep{vidprom2024}, a corpus of approximately 1.7M
human-authored video generation prompts. One main reason to switch to human authored prompts
is that using an LLM to generate a prompt corpus without any context or feedback loop often results in prompts with
high cosine similarity (mode collapse). Human seeded prompts, although imperfect have enough entropy
to counter this effect. A filtering process is applied to the VidProM dataset that leads to approximately 5000 prompts retained ($\sim$0.3\%
of the original corpus). See Appendix~\ref{app:filtering} for more details.

\paragraph{Multi-dimensional judging.}
16 video evaluation dimensions (\S~\ref{app:defs}) are defined and organized into four groups.

\textbf{Group~A: Motion \& Stability} (Subject Consistency, Scene Consistency, Motion
Smoothness, Temporal Flickering, Inertial Consistency);

\textbf{Group~B: Logic \& Physics} (Physical Mechanics, Object Permanence, Human Fidelity,
Dynamic Degree);

\textbf{Group~C: Instruction Adherence} (Semantic Adherence, Spatial Relationship,
Semantic Drift);

\textbf{Group~D: Aesthetic Quality} (Composition \& Framing, Lighting \& Volumetric,
Color Harmony, Structural Gestalt).

Each prompt from the filtered prompt set, is scored on all applicable dimensions using Gemini~3.1~Flash-Lite~\citep{geminiteam2023}
on a \emph{suitability} scale (1--10: how well-suited is this prompt for testing this
dimension?) and a \emph{difficulty} scale (1--10: how challenging is this prompt for a
typical video model?). (see \S \ref{app:system_prompts_curation})

Of the 5,000 pre-filtered prompts, 276 (7.4\%) were additionally flagged during judging
as requiring copyright or safety review (e.g., references to trademarked characters or
public figures) and were subsequently remediated. Furthermore, manual pruning was used to 
remove veiled content. The final curated set contains
\textbf{3,754 unique prompts}, which are available for public use.
Prompts are also LLM-enhanced for cleanup and to increase difficulty and suitability for each benchmarking dimension. 

\subsection{Phase B: Evaluation Engine}
\label{sec:phase_b}
% ══════════════════════════════════════════════════════════════════════════════

\subsubsection{Prompt-Specific Question Generation}
\label{sec:questions}

For each prompt and each applicable dimension, 10 evaluation questions are generated via
a structured Gemini~3~Flash prompt (\S~\ref{app:system_prompts_vlm} )
\begin{itemize}[leftmargin=1.5em,topsep=2pt,itemsep=0pt]
  \item \textbf{Context:} the original prompt text and the target dimension definition,
    including scope, exclusions, and focus areas.
  \item \textbf{Scale anchors:} a 1--5 Likert rubric with anchor descriptions at each
    level (1 = major failure, 3 = mediocre/passable, 5 = flawless execution).
  \item \textbf{Coverage types:} questions must cover four categories. They are expected events and prompt details, potential failure modes, success modes, and adversarial probes
    checking for subtle inconsistencies. In particular, generic or near-duplicate questions are explicitly discouraged.
\end{itemize}
Some example question sets are provided in \S~\ref{app:casestudies}.

\subsubsection{Video Generation}
\label{sec:videogen}

Using the curated prompts one video per (model, prompt) pair is generated via API
or local inference (see \S~\ref{sec:shared_setup}). All models run with default settings for the scope of the paper and no prompt-specific tuning is applied. 

\subsubsection{VLM Evaluation Engine}
\label{sec:vlm}

This section describes the instantiation of the VLM engine used in this paper.
Each video is evaluated by Gemini~3~Flash (\texttt{gemini-3-flash-preview}) in a multimodal call that receives a sequence
of extracted frames and the dimension-specific VQA questions (with rubrics) for that dimension.
The original prompt that generated the video is \emph{not} passed directly, instead its semantic content is already embedded in the per-prompt, per-dimension VQA questions generated during curation, making the evaluation criteria explicit and auditable rather than leaving them to the VLM's open-ended interpretation of free text.

Frame extraction follows a \emph{dimension-aware sampling strategy}
with three modes, chosen to match the temporal granularity needed for each dimension:

\begin{itemize}[topsep=3pt,itemsep=2pt]
  \item \textbf{Holistic mode} (32 frames, uniform stride): used for dimensions whose
    assessment requires a global view of the full clip, Semantic Adherence, Composition
    \& Framing, Aesthetic Quality, Lighting \& Volumetric, Color Harmony, Structural
    Gestalt, Dynamic Degree, and Semantic Drift.  The 32 frames are drawn at positions
    $\lfloor i \cdot T/32 \rfloor$ for $i \in \{0,\ldots,31\}$, where $T$ is the total
    frame count, giving uniform coverage from first to last frame. 

  \item \textbf{Sampled mode} (16 frames, uniform stride): used for dimensions that
    require checking temporal \emph{consistency} across the clip without needing
    fine-grained motion, Scene Consistency, Object Permanence, and Subject Consistency.
    Frames are placed at $\lfloor i \cdot T/16 \rfloor$ for $i \in \{0,\ldots,15\}$.
    Sixteen frames are sufficient to catch identity or environment drift while keeping
    inference cost low.

  \item \textbf{Micro mode} (dense prefix, $\approx$12 frames): used for dimensions
    that are most sensitive to \emph{frame-to-frame} artifacts in short intervals, Motion
    Smoothness, Temporal Flickering, Inertial Consistency, Physical Mechanics, and Human
    Fidelity.  Frames are extracted at every 5th frame index starting from 0, up to a
    maximum of 60 frames (i.e.\ the first ${\approx}2$ seconds at 30\,fps), yielding
    roughly 12 frames.  Focusing on the clip's opening seconds is justified because
    physical plausibility and motion artifacts are most apparent at the start of action,
    before the generator can accumulate temporal self-consistency.
\end{itemize}

The VLM outputs an integer Likert score (1--5) and a free-text justification for each
question. Chain-of-thought prompting is avoided to avoid length bias~\citep{zheng2023mtbench}. The justifications serve as a post audit trail for quality control. All Gemini~3~Flash calls use the API default sampling temperature ($T{=}1.0$, stochastic);
no \texttt{generation\_config} is specified, so scores have inherent run-to-run variance
(quantified in Ablation~A5, \S~\ref{sec:abl_reliability}). 

\subsubsection{Scoring System}
\label{sec:scoring}

\paragraph{Dimension scores.}
Raw Likert scores (1--5) are averaged across the 10 questions for a given (prompt,
model, dimension) triple, then averaged across all prompts to produce a model-level
dimension score. Prompts for which a dimension is marked as unsuitable during curation (i.e.\ a
\emph{null dimension}) are excluded from that dimension's average; each model-level
score is therefore computed over only the subset of prompts for which the dimension
is applicable.

\paragraph{BT rating via Bradley-Terry.}
The dimension scores above aggregate across prompts to produce model-level averages
(used in dimension analysis). For the BT rating, the aggregation goes in the other direction:
the per-prompt dimension scores are averaged across all applicable (non-null) dimensions
to yield a single per-model per-prompt score. For every prompt, all $\binom{6}{2}=15$ model pairs are compared:
the model with the higher per-prompt score wins that matchup ($+1$);
an exact tie awards each model $+0.5$. In practice, ties are negligible given
floating-point averages across multiple dimensions and questions for the VLM, while for the humans, it is averted by using confidence labels. This yields $15 \times 50 = 750$ prompt-level matchups in total.

Let $W_{ij}$ denote the total wins of model $i$ over model $j$, and
$n_{ij} = W_{ij}+W_{ji}$ the total comparisons between them.
The Bradley-Terry model~\citep{bradley1952} assigns each model a latent strength
$p_i > 0$ such that the probability that $i$ beats $j$ is
\begin{equation}
  P(i \succ j) = \frac{p_i}{p_i + p_j}.
  \label{eq:bt}
\end{equation}
Strengths are estimated via the Minorization-Maximization (MM) algorithm~\citep{hunter2004mm}, whose update rule is
\begin{equation}
  p_i^{(t+1)} = \frac{W_i}{\displaystyle\sum_{j \neq i} \dfrac{n_{ij}}{p_i^{(t)} + p_j^{(t)}}},
  \qquad W_i = \sum_j W_{ij},
  \label{eq:mm}
\end{equation}
iterated until $\max_i|p_i^{(t+1)}-p_i^{(t)}| < 10^{-8}$ (at most 1{,}000 steps).
Strengths are normalised to unit geometric mean and converted to a 1500-centred log-odds scale via
\begin{equation}
  \text{BT rating}_i = 1500 + 400\log_{10}\!\left(\frac{p_i}{\bar{p}_{\mathrm{geom}}}\right),
  \qquad \bar{p}_{\mathrm{geom}} = \exp\!\left(\tfrac{1}{K}\textstyle\sum_j \ln p_j\right),
  \label{eq:bt_rating}
\end{equation}
where $K=6$ is the number of models being compared (anchoring the scale so the
geometric-mean model sits at 1500). We employ the Bradley-Terry model to estimate the latent skill parameters of the models 
based on the pairwise comparison matrix. To improve interpretability, 
the resulting worth parameters were transformed into a scale with a mean of 1500. 
It is important to note that this is a global optimization of the complete dataset, 
distinct from the iterative, sequential updates characteristic of the ELO rating system.

Uncertainty is quantified by 1{,}000 prompt-level bootstrap resamples (with
replacement); the 2.5th and 97.5th percentiles of the bootstrap distribution
form the reported 95\% confidence interval (CI). The canonical point estimate is the 
deterministic BT-MLE (Eq.~\eqref{eq:bt_rating}) and not the bootstrap mean. Rank correlations are measured with Spearman's $\hat{\rho}$ or Kendall's $\tau$ \footnote{\url{https://library.virginia.edu/data/articles/correlation-pearson-spearman-and-kendalls-tau}}.

\paragraph{Predicted Human Alignment Score (PHAS).}
In addition to BT ratings, a weighted average of dimension scores calibrated
against human pairwise preference annotations (details and calibrated weights in
\S~\ref{sec:phas}) is introduced.

% ─────────────────────────────────────────────────────────────────────────────
%\FloatBarrier
\section{Dataset and Models}
\label{sec:shared_setup}
% ─────────────────────────────────────────────────────────────────────────────

Both the human preference study \S~\ref{sec:human_study} and the VLM evaluation
\S~\ref{sec:vlm_eval} operate on the same fixed set of prompts and videos described here.
\paragraph{Prompt selection.}
\textbf{50 prompts} (suitability $>8$ on $\geq 5$ dimensions are sampled from the curated
pool of 3{,}754 prompts. Per-dimension suitability scores
from the re-scoring phase determine which dimensions are applicable for VQA questionnaire
generation (null-suitability dimensions are skipped). This full prompt set is used both in
the human study \S~\ref{sec:human_study} and VLM study \S~\ref{sec:vlm_eval} for data collection.

For the PHAS section \S~\ref{sec:phas}, the 50 prompts are partitioned into two non-overlapping subsets: \textbf{20 validation
prompts} (mean suitability $9.05$, mean difficulty $8.22$) and \textbf{30 calibration
prompts} (mean suitability $9.14$, mean difficulty $8.32$).
The two subsets are statistically indistinguishable in prompt quality, confirming no
systematic bias. This split is used \emph{exclusively} for PHAS weight calibration
(\S~\ref{sec:phas}).
\paragraph{Models and video generation.}
Six SOTA video models are evaluated: \veo~\citep{veo2025}, \kling~\citep{kling2024},
\ltx~\citep{ltx2025}, \wanb~\citep{wan2025},
\hunyuan~\citep{hunyuanvideo2024}, and \wans~\citep{wan2025} (local open-source baseline).
Commercial models are accessed via the fal.ai~\citep{fal2024} inference API:
\begin{itemize}[noitemsep,topsep=2pt]
  \item \texttt{fal-ai/veo3.1/fast} — \textit{generated March~2026}
  \item \texttt{fal-ai/kling-video/v2.6/pro/text-to-video} — \textit{generated March~2026}
  \item \texttt{fal-ai/wan/v2.2-a14b/text-to-video} — \textit{generated March~2026}
  \item \texttt{fal-ai/ltx-2/text-to-video} — \textit{generated March~2026}
  \item \texttt{fal-ai/hunyuan-video-v1.5/text-to-video} — \textit{generated March~2026}
\end{itemize}
As FAL API endpoints may be updated silently, all generated videos are released with the
dataset to ensure exact reproducibility independent of future API changes.
\wans\ was run locally across 8~H200 GPUs.
One video is generated per (model, prompt) pair, yielding \textbf{300 videos} in total.
Veo~3.1~Fast produces ${\approx}8$\,s clips; all other five models produce ${\approx}5$\,s
clips. Some of the videos come with audio, but our current framework does not test audio, \emph{both} in the human/VLM evaluation phases. Frame sampling is \emph{duration-relative} (frames drawn at fixed percentages of
total length), so the same frame count is analysed regardless of clip duration.
Implications for holistic, sampled, and micro dimensions described in \S~\ref{sec:vlm} are discussed in the
limitations (\S~\ref{sec:limitations}).

% ─────────────────────────────────────────────────────────────────────────────
%\FloatBarrier
\section{Human Preference Study}
\label{sec:human_study}

This section establishes human ground-truth
rankings using the shared dataset described in \S~\ref{sec:shared_setup}
(50 prompts, 6 models, 300 videos).
These rankings serve as the validation target for \S~\ref{sec:vlm_eval}: if the VLM
engine reproduces them, that provides evidence that automated evaluation is a reliable
proxy for human judgement.

\paragraph{Setup.}
A blind pairwise preference study is conducted using a custom web application
(Figure~\ref{fig:evalui}; Appendix~\ref{app:human_protocol}) built on Google Apps
Script and served over Google Drive.  For each prompt, the evaluator interface
generates all $\binom{6}{2}=15$ pairwise model comparisons, producing a total pool
of \textbf{750 comparison pairs}.  Annotators are identified by self-reported email
at session start; the interface resumes from each annotator's last completed pair
on return.  The dataset replaces all email addresses with opaque IDs
(\texttt{A1}--\texttt{A7}) to protect annotator privacy; affiliation categories
and per-annotator contribution counts are reported in Table~\ref{tab:annotators}.

\textbf{Blinding.}  Model identities are masked as ``Video~A'' and ``Video~B'';
left/right assignment is re-randomised independently for every pair.  Annotators
are not told the number of models or their names.

\textbf{Prompt presentation.}  A short \emph{prompt summary} is displayed as an
italic tag above the video arena.  Hovering the tag reveals the full enhanced prompt
in a tooltip overlay (Figure~\ref{fig:evalui}, bottom), allowing annotators to verify
exact wording, attribute lists, and spatial relationships on demand without the full
text cluttering the comparison view.

\textbf{Response format.}  Annotators make a forced binary choice with a
three-level confidence signal, producing six effective response classes
(Video~A / Video~B $\times$ Much / Clearly / Slightly better).  There is no
\emph{tie} option; ``Slightly better'' acts as the forced best-guess for marginal
or indistinguishable pairs.  This prevents abstention while still capturing
preference strength.  The three confidence levels (Much=3, Clearly=2, Slightly=1)
are stored as annotation weights and used as sample weights in the PHAS logistic
regression calibration (\S~\ref{sec:phas}); Note that the BT model uses only
the binary win direction.

\textbf{Interface design.}  Vote buttons are disabled until both videos have loaded
and begun playing simultaneously, preventing premature votes on an incomplete pair.
A \textbf{10-second watch timer} (animated progress bar beneath each video card)
further enforces a minimum viewing window before the vote buttons unlock, ensuring
annotators observe at least one full loop of each clip before judging.
Videos are delivered as base64-encoded blobs from Google Drive and cached client-side;
the next two pairs are prefetched in the background while the annotator watches the
current pair. A 12-second timeout triggers an automatic cache-busted reload for
stuck streams.  Keyboard shortcuts (\texttt{Space} = pause/play both, \texttt{R}
= restart both) support efficient review.

\textbf{Coverage design.}  Each annotator works through their own personal queue of
all 750~pairs they have not yet judged, sorted by ascending global coverage so the
least-reviewed pairs are served first.  Different annotators therefore provide
complementary coverage across the 750~pairs.  A break overlay appears every 50~pairs
to reduce fatigue.  All annotations are
pooled before fitting a single Bradley-Terry model (unweighted: each vote contributes one win/loss), which handles incomplete
tournament graphs natively. Only graph \emph{connectivity} is required, which is met as long as
each model appears in at least one comparison, and not full coverage.  The BT rating uncertainty is captured through 1{,}000-sample bootstrap CIs on the Bradley-Terry fit.
Each additional annotator improves coverage and reduces bootstrap variance.

\begin{figure}[t]
  \centering
  \includegraphics[width=0.95\linewidth]{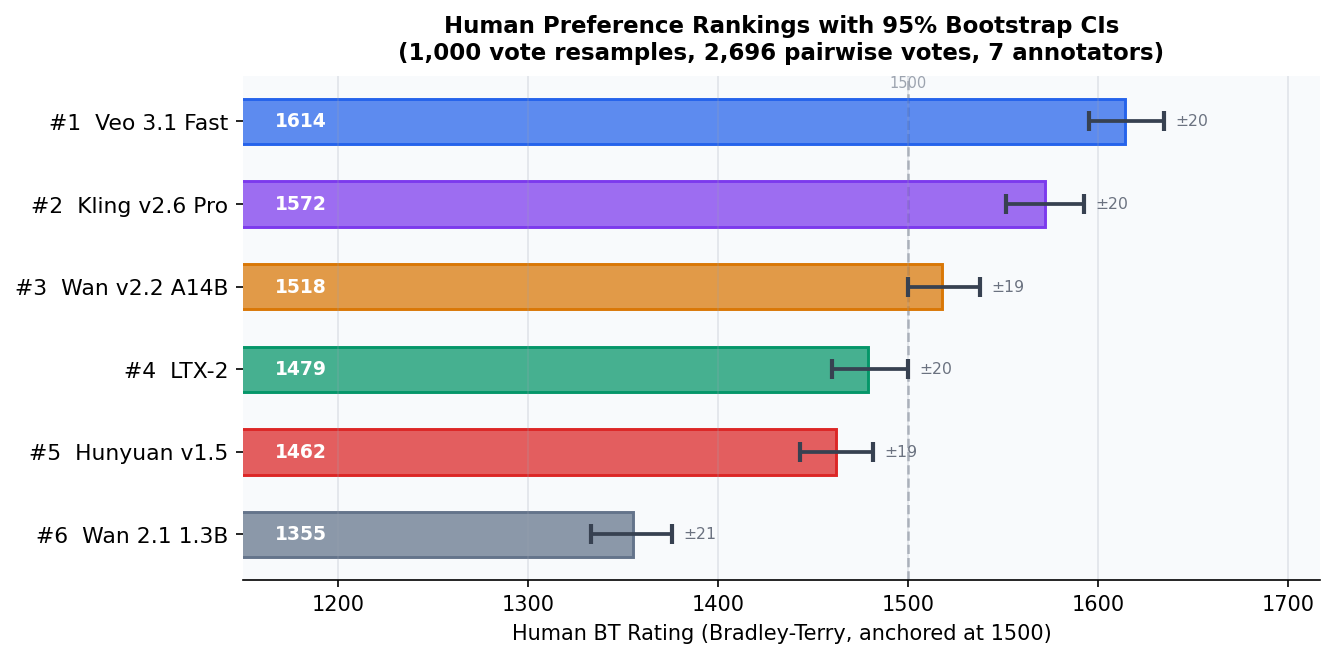}
  \caption{Human BT ratings with 95\% bootstrap CIs
    (1{,}000 vote resamples of the 2{,}696 pairwise votes, 7 annotators, 100\% pair coverage).
    Three-tier structure: \veo\ and \kling\ form a clear top tier;
    \wanb, \ltx, and \hunyuan\ form a statistically indistinguishable mid-tier cluster;
    \wans\ trails by more than 100 points.}
  \label{fig:human_bt}
\end{figure}

\paragraph{Results.}
With 2{,}696 pairwise annotations from 7 annotators covering all 750 model-pair
slots (100\% coverage), the human Bradley-Terry model produces stable rankings across
all 6 models. Mean inter-annotator agreement across 750 shared pairs is 66.9\%
with Krippendorff $\alpha = 0.273$~\citep{krippendorff2018} in the ``Fair'' tier~\citep{landis1977} for binary preference data.
Table~\ref{tab:rankcompare} reports the human BT ratings along with their CI intervals.
Figure~\ref{fig:human_bt} shows the human BT ratings with 95\% bootstrap CIs. 

\begin{table}[t]
\centering
\caption{Human Bradley-Terry rankings (2{,}696 pairwise votes, 7 annotators,
  100\% pair coverage).
  Bradley-Terry MLE rankings; 95\% bootstrap CIs from 1{,}000 vote resamples.
  BT strength = log-odds relative to mean (higher is better).
  Within-tier ordering is consistent but not statistically significant given overlapping CIs.}
\label{tab:rankcompare}
\small
\setlength{\tabcolsep}{6pt}
\begin{tabular}{lcccc}
\toprule
\textbf{Model} & \textbf{Rank} & \textbf{Human BT [95\% CI]} & \textbf{BT Strength} & \textbf{Tier} \\
\midrule
\veo      & 1 & 1614 \scriptsize{[1595, 1635]} & $+0.66$ & Top \\
\kling    & 2 & 1572 \scriptsize{[1552, 1592]} & $+0.41$ & Top \\
\midrule
\wanb     & 3 & 1518 \scriptsize{[1500, 1538]} & $+0.10$ & Mid \\
\ltx      & 4 & 1479 \scriptsize{[1460, 1500]} & $-0.12$ & Mid \\
\hunyuan  & 5 & 1462 \scriptsize{[1443, 1482]} & $-0.22$ & Mid \\
\midrule
\wans     & 6 & 1355 \scriptsize{[1333, 1376]} & $-0.83$ & Bottom \\
\bottomrule
\end{tabular}
\end{table}

\paragraph{Three-tier structure.}
The human BT rating reveals a clear three-tier hierarchy that will later serve as
ground truth for evaluating the VLM pipeline (\S~\ref{sec:leaderboard_vlm}):
\begin{itemize}[leftmargin=1.5em,topsep=2pt,itemsep=1pt]
  \item \textbf{Top tier}: \veo\ (Human BT~1614, BT strength $+0.66$) and
    \kling\ (1572, $+0.41$), the ${\approx}54$-point gap from \kling\ down to
    the mid-tier exceeds the bootstrap CI half-width (${\approx}20$ points),
    making tier membership unambiguous.
    Within this tier, \veo\ leads \kling\ by $+0.25$ BT log-odds; both models
    sit well clear of the mid-tier cluster.
  \item \textbf{Mid tier}: \wanb\ (1518), \ltx\ (1479), and \hunyuan\ (1461):
    human annotators place all three models within a 57-point band,
    confirming that these models are of comparable quality.
    The canonical human BT ordering within this tier (\wanb\ $>$ \ltx\ $>$ \hunyuan)
    is consistent but not statistically significant given the overlapping bootstrap CIs;
    these three models should be treated as a tied cluster.
  \item \textbf{Bottom tier}: \wans\ (1355, BT strength $-0.83$):
    separated from the mid-tier by ${\approx}106$ Human BT points, well outside
    the bootstrap CI. The top-to-bottom BT log-odds span (\veo\ to \wans) of $1.49$ is the largest pairwise log-odds gap in the ranking, reflecting near-universal human consensus
    that this model trails the field.
\end{itemize}

\paragraph{Annotator composition and bias mitigations.}
The 7 annotators comprise 4 domain experts (Expert~1--4, where Expert~1 is a co-author)
and 3 independent external annotators with academic backgrounds in mathematics (A5),
social sciences (A6), and visual art (A7); see Table~\ref{tab:annotators}.
Effort was made to get more external votes, and the 3 external annotators contribute 1{,}647 votes (61.1\%).

\begin{table}[t]
\centering
\caption{Annotator breakdown for the human preference study.
  Confidence levels: \textbf{Mb} = Much better, \textbf{Cl} = Clearly better,
  \textbf{Sl} = Slightly better.}
\label{tab:annotators}
\small
\setlength{\tabcolsep}{5pt}
\begin{tabular}{clcrrrr}
\toprule
\textbf{ID} & \textbf{Role} & \textbf{N} & \textbf{\%} & \textbf{Mb\%} & \textbf{Cl\%} & \textbf{Sl\%} \\
\midrule
A1 & Expert~1 (co-author)         & 750 & 27.8 & 56.5 & 25.9 & 17.6 \\
A2 & Expert~2                     & 150 &  5.6 & 28.7 & 48.7 & 22.7 \\
A3 & Expert~3                     &  99 &  3.7 & 16.2 & 39.4 & 44.4 \\
A4 & Expert~4                     &  50 &  1.9 & 40.0 & 28.0 & 32.0 \\
A5 & External (mathematics)       & 750 & 27.8 &  5.2 & 61.2 & 33.6 \\
A6 & External (social sciences)   & 750 & 27.8 & 37.5 & 21.6 & 40.9 \\
A7 & External (visual art)        & 147 &  5.5 & 62.6 & 20.4 & 17.0 \\
\midrule
\multicolumn{2}{l}{\textbf{Total}} & \textbf{2696} & \textbf{100} & & & \\
\bottomrule
\end{tabular}
\end{table}

Expert annotator participation (Expert~1--4) introduces a potential for unconscious bias.
We employ four design choices that mitigate this risk.
First, model identities were fully masked throughout the study, for example annotators saw
only ``Video~A'' and ``Video~B'' with left/right assignment re-randomised per
pair, making it infeasible to preferentially score a specific model.
Second, the 66.9\% mean inter-annotator agreement and Krippendorff $\alpha=0.273$
are consistent with the inherent subjectivity of pairwise video preference tasks
and in particular, with the inherent subjectivity of aesthetic preference judgements more broadly. Importantly, this variance does not impede the efficacy of the BT scoring, as is evident from the robustness of the 95\% bootstrap CIs Table~\ref{tab:rankcompare}, Figure~\ref{fig:human_bt}. Third, 3 external annotators contributed 1{,}647 votes (61.1\%) which significantly works against expert bias. Fourth, annotator \texttt{A1} completed the full 750-pair annotation twice under separate accounts (\emph{test-retest} self-consistency check) and achieved agreement across the two passes with \textbf{89.7\%} (673/750 pairs). This is substantially higher than the 66.9\% mean cross-annotator agreement and well above the 50\% chance baseline.

\paragraph{Confidence-scale heterogeneity.}
Table~\ref{tab:annotators} reveals substantial variation in how annotators
use the three confidence labels, independent of their directional choices.
\texttt{A5} (external, mathematics) marks only 5.2\% of comparisons as
``Much better'' the lowest of any annotator, compared to 62.6\% for
\texttt{A7} (external, visual art) and 56.5\% for \texttt{A1}.
This is not evidence of directional disagreement: \texttt{A5}'s pairwise exact agreement with \texttt{A2}
the annotator with the most similar confidence distribution is 76.0\% (``good'').
Thus, \texttt{A5}'s pattern reflects a systematic preference for the
``Clearly better'' label as a ceiling: their distribution is bimodal between
``Clearly'' (61.2\%) and ``Slightly'' (33.6\%), effectively collapsing the
three-level scale to two.

The Bradley-Terry rating is entirely unaffected, since it ingests only the
\emph{direction} of each vote (who wins the pair), not the confidence label.
\texttt{A5}'s 750 votes therefore contribute equally to the BT fit as
\texttt{A1}'s or \texttt{A7}'s.
Confidence labels (Much$=$3, Clearly$=$2, Slightly$=$1) are used solely as
sample weights in the PHAS logistic regression calibration, as described in
\S~\ref{sec:phas}.

% ─────────────────────────────────────────────────────────────────────────────
%\FloatBarrier
\section{VLM Evaluation}
\label{sec:vlm_eval}
% ─────────────────────────────────────────────────────────────────────────────

We now ask whether \worldjen's VLM-as-a-judge pipeline reproduces the human
ground-truth ranking automatically.

\subsection{Evaluation Protocol}
\label{sec:eval_protocol}
\label{sec:setup}

The prompt set, model pool, and generated videos are identical to \S~\ref{sec:human_study} and are described in \S~\ref{sec:shared_setup}. This section covers the VLM-specific evaluation procedure.

\paragraph{VQA and VLM evaluation.}
For each prompt, 10 Likert-scale questions per applicable dimension are generated
by Gemini~3~Flash (offline) from the enhanced prompt text, and stored in per-prompt
question files.
Videos are evaluated by Gemini~3~Flash (VLM) using dimension-specific frame-sampling
modes (holistic, sampled, micro), yielding approximately
$16 \times 10 \times 300 = 48{,}000$ theoretical Likert responses,
reduced to \textbf{47{,}160} after excluding 84 null (prompt, dimension) pairs
(78 Human Fidelity nulls on prompts containing no human subjects;
6 Object Permanence nulls on prompts with no trackable objects).
BT ratings for the main leaderboard are derived as described in \S~\ref{sec:scoring}.

\subsection{Main Results: VLM BT Rating Leaderboard}
\label{sec:leaderboard_vlm}

Table~\ref{tab:leaderboard} reports the canonical VLM BT rating (point estimate), overall average Likert score for
all six models across 50 prompts. The interpretation is that this is the most likely true ordering
given the data we have. The leaderboard table also estimates a confidence interval, by resampling
the 50 prompts data 1000 times, to estimate how much the BT rating would shift if one picked a slightly
different prompt set. A wide CI interpretation is that the estimate is sensitive to which prompts
are included. 

\begin{table}[t]
\centering
\caption{%
  \worldjen\ leaderboard (50 prompts $\times$ 6 models $=$ 300 videos). Canonical point wise
  VLM BT Eq.~\eqref{eq:bt_rating} via Bradley-Terry MLE~\citep{bradley1952} and 95\% CI from 1{,}000
  prompt-level bootstrap resamples.
}
\label{tab:leaderboard}
\small
\setlength{\tabcolsep}{5pt}
\begin{tabular}{lccccc}
\toprule
\textbf{Model} & \textbf{Canonical VLM BT} & \textbf{95\% CI} & \textbf{VLM Rank} & \textbf{Avg Score} &\textbf{Tier} \\
\midrule
\veo      & \textbf{1652} & [1590, 1728] & 1 & \textbf{4.32} & Top\\
\kling    & 1628          & [1571, 1697] & 2 & 4.27  & Top        \\
\midrule
\wanb     & 1509          & [1454, 1569] & 3 & 4.07 & Mid         \\
\ltx      & 1504          & [1444, 1560] & 4 & 4.10  & Mid        \\
\hunyuan  & 1433          & [1374, 1485] & 5 & 4.01 & Mid\\
\midrule
\wans     & \underline{1274} & [1212, 1323] & 6 & \underline{3.82}  & Bottom \\
\bottomrule
\end{tabular}
\end{table}

\begin{figure}[t]
  \centering
  \includegraphics[width=0.95\linewidth]{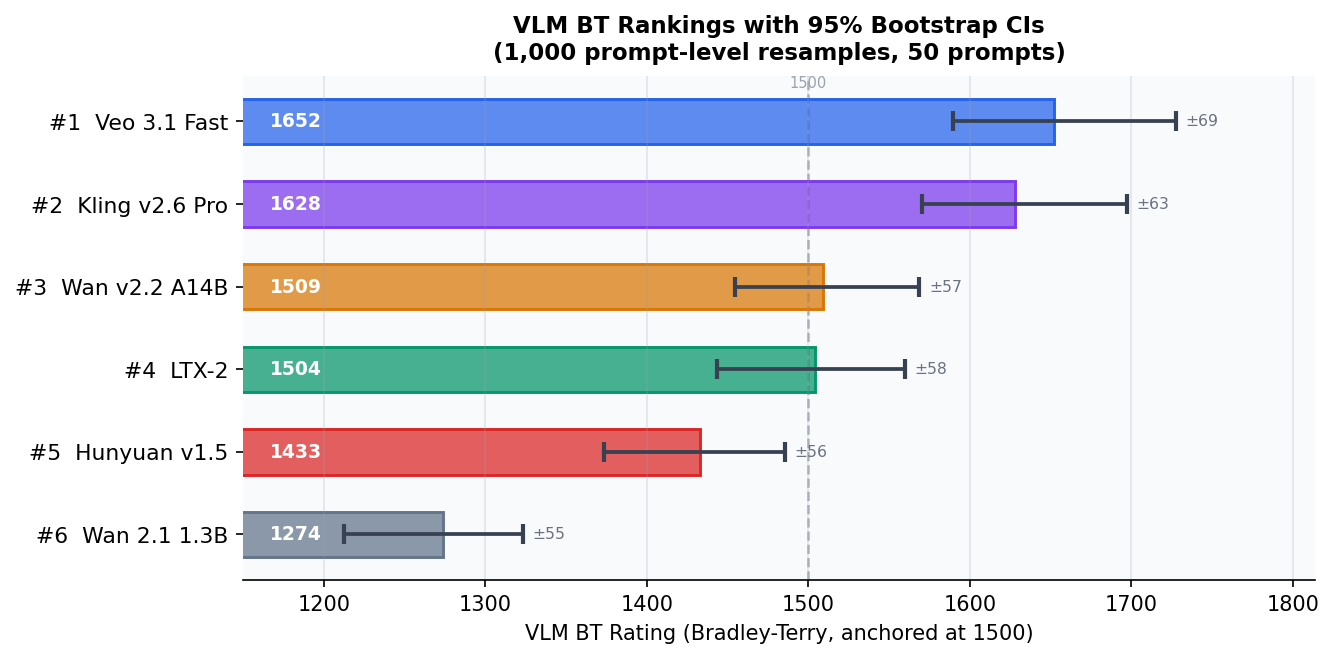}
  \caption{VLM BT rating (Gemini~3~Flash) with 95\% bootstrap confidence intervals
    (1{,}000 prompt-level resamples, 50 prompts).
    Rank labels are embedded in the model names (\#1 = best).
    Canonical BT values are shown inside each bar; CI half-widths ($\pm$) appear after each whisker.
    Veo~3.1~Fast and Kling~v2.6~Pro form a clear top tier;
    LTX-2, Wan~v2.2, and Hunyuan form a statistically indistinguishable mid-tier cluster
    with fully overlapping CIs.
    A comparison with the human BT rating is in Figure~\ref{fig:vlm_human_comparison} (\S~\ref{sec:vlm_human_comparison}).}
  \label{fig:bt_ci}
\end{figure}

\paragraph{Three-tier structure.}
The BT rating distribution reveals a clear three-tier hierarchy.
Bootstrap resampling (1{,}000 prompt-level draws; Figure~\ref{fig:bt_ci}) confirms
the structure is stable: tier membership is invariant across resamples, with
inter-tier gaps that exceed per-model CI half-widths ($\pm$69~pts for the top tier).
\begin{itemize}[leftmargin=1.5em,topsep=2pt,itemsep=1pt]
  \item \textbf{Top tier}: \veo\ (1652) and \kling\ (1628) are
    separated from the mid-tier by ${\approx}110$ BT points (roughly
    $1.7\times$ the per-model bootstrap CI half-width, $\pm$65~pts).
    Within this tier the Veo/Kling ordering is statistically
    unresolved at 50~prompts (24-pt gap $<$ CI).
  \item \textbf{Mid tier}: \wanb\ (1509), \ltx\ (1504), and \hunyuan\ (1433), again form
    a statistically indistinguishable cluster with fully overlapping CIs.
    Within-tier rank differences are noise at the current sample size. 
  \item \textbf{Bottom tier}: \wans\ (1274):
    separated from the mid-tier by ${\approx}158$ BT points (${\approx}2.4\times$ CI),
    confirming that the 1.3B open-source model is a clear step behind the
    larger commercial and high-capacity open-source models. Note that in table \ref{tab:leaderboard} \wanb\ has a marginally lower average score yet a marginally higher BT rating: this is
expected because the BT rating is determined by pairwise \emph{win counts}, not score magnitude.
\end{itemize}
\paragraph{Interpreting the confidence intervals.}
CI overlap indicates that the \emph{score estimates} of the two models in comparison are close
relative to prompt-level sampling noise. The correct diagnostic for rank reliability in this case,
is pairwise concordance across bootstrap draws. Ablation A3 (\S~\ref{app:abl_nprompts}) 
shows that at $N{=}50$ prompts the full six-model rank is recovered in \textbf{100\%} of 500 independent
bootstrap subsamples, which is a perfect concordance rate despite the overlapping CIs
visible in Figure~\ref{fig:bt_ci}. This gives us confidence to interpret the within-tier pairs 
(Veo/Kling; Wan~A14B/LTX-2/Hunyuan) as \emph{genuine near-ties in quality} and that 
the benchmark reflects that closeness.

The inter-tier gaps (${\approx}110$--$158$~pts, ${\approx}1.7$--$2.4\times$ the
CI half-width) are large enough that tier membership is stable under any
realistic increase in prompt count. This can clearly be interpreted as difference in quality. 
Figure~\ref{fig:bt_ci} visualises each model's VLM BT rating with 95\% bootstrap
confidence intervals (1{,}000 prompt-level resamples). 

\subsection{VLM--Human Rank Comparison}
\label{sec:vlm_human_comparison}
\begin{figure}[t]
  \centering
  \includegraphics[width=\linewidth]{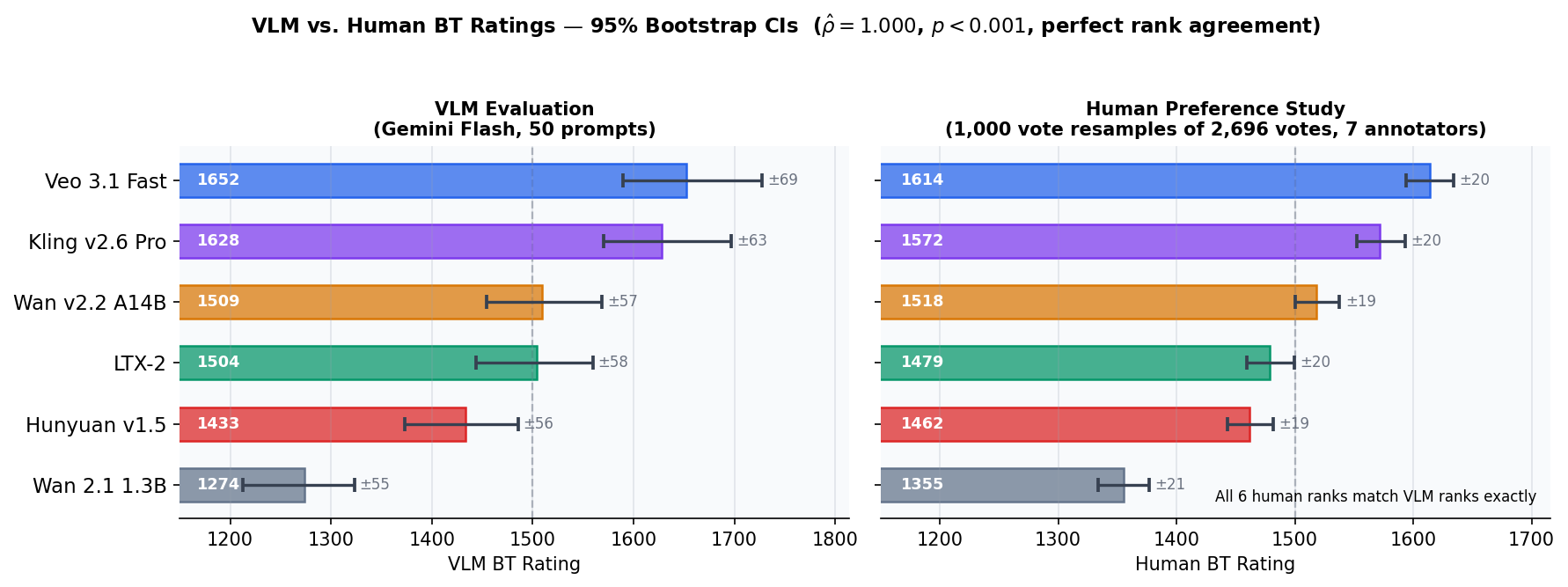}
  \caption{Side-by-side BT ratings with 95\% bootstrap CIs.
    \textbf{Left:} VLM BT (Gemini~3~Flash, 1{,}000 prompt-level resamples, 50 prompts).
    \textbf{Right:} Human BT (1{,}000 bootstrap resamples of the 2{,}696 pairwise votes,
    7 annotators, 100\% pair coverage).
    The VLM reproduces the human-established three-tier partition.
    Spearman $\hat{\rho}=1.000$ ($n{=}6$; tier-level agreement).}
  \label{fig:vlm_human_comparison}
\end{figure}

% ─────────────────────────────────────────────────────────────────────────────
In Figure~\ref{fig:vlm_human_comparison} we place the human ground-truth ranking established in \S~\ref{sec:human_study} alongside the VLM BT rating computed in \S~\ref{sec:leaderboard_vlm}.
The three-tier structure first established by human judges (\S~\ref{sec:human_study})
is reproduced in full by the VLM evaluation. We achieve a Spearman $\hat{\rho}=1.000,~p=0.0014,~n{=}6$ and Kendall $\tau=1.000,~p=0.003$.
Because the six models form three well-separated tiers with large between-tier BT rating gaps,
any method that correctly assigns tier membership will mechanically achieve $\hat{\rho}=1.000$;
the result should therefore be interpreted as tier-level concordance
rather than evidence of fine-grained within-tier discrimination.
That said, the BT rating rests on a dense signal: 160 prompt-specific VQA questions per video
(10 questions $\times$ 16 dimensions) aggregated across 50 prompts, yielding 47{,}160 scored
responses in total. The consistency of the tier ordering across all 15 model pairs (15/15
concordant) and across all 16 individual dimensions (\S~\ref{sec:dimensions}) provides
additional confidence that the individual model ranking reflects genuine quality differences rather than
sampling noise.

The VLM evaluation pipeline reproduces the human-established three-tier partition
without any shared signal, providing the primary empirical validation of \worldjen. This confirms that VLM-as-a-judge can be used as a proxy for human preference.

% ─────────────────────────────────────────────────────────────────────────────
% %\FloatBarrier
\subsection{Dimension-Wise Analysis}
\label{sec:dimensions}

\begin{figure}[tbp]
  \centering
  \includegraphics[width=0.97\linewidth]{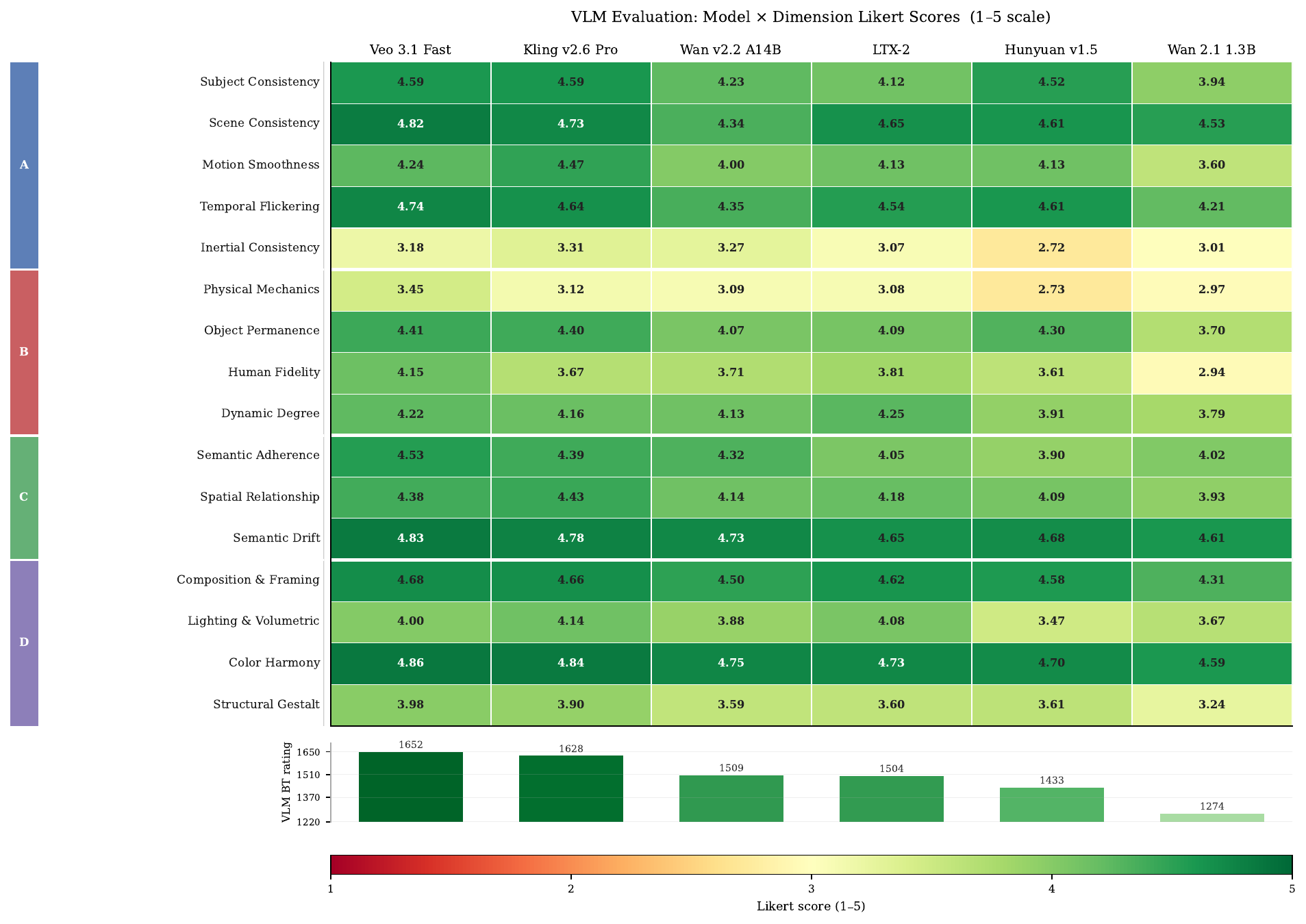}
  \caption{Model $\times$ dimension Likert scores (1--5 scale, averaged over 50 prompts
    and 10 questions per dimension). Models are sorted left-to-right by VLM BT ratings
    (bottom panel). Inertial consistency and physical mechanics are visibly lighter across all
    models, confirming the physics gap across all evaluated models. \veo\ achieves the highest BT rating
    (1652) and leads or ties on 12 of 16 dimensions.}
  \label{fig:heatmap}
\end{figure}

Table~\ref{tab:dimensions} shows per-model VLM scores across all 16 dimensions,
organised vertically by the four evaluation groups. 
Figure~\ref{fig:heatmap} visualises the same scores as a green--red heatmap,
making cross-model patterns immediately apparent, such as the physics gap (Group~B),
the aesthetic ceiling (Group~D).

\begin{table}[t]
\centering
\caption{Per-model per-dimension Likert scores (1--5 scale, averaged over 50 prompts
  and 10 questions per dimension; null-suitability dimensions excluded per prompt).
  \textbf{Bold} = best per dimension; \underline{underline} = worst.}
\label{tab:dimensions}
\footnotesize
\setlength{\tabcolsep}{4pt}
\renewcommand{\arraystretch}{1.1}
\begin{tabular}{llcccccc}
\toprule
\textbf{Group} & \textbf{Dimension} &
  \textbf{Veo 3.1} & \textbf{Kling} & \textbf{LTX-2} &
  \textbf{Wan A14B} & \textbf{Hunyuan} & \textbf{Wan 1.3B} \\
\midrule
\multirow[t]{5}{*}{A: Motion}
  & Subject Consistency  & \textbf{4.59} & 4.59 & 4.12 & 4.23 & 4.52 & \underline{3.94} \\
  & Scene Consistency    & \textbf{4.82} & 4.73 & 4.65 & \underline{4.34} & 4.61 & 4.53 \\
  & Motion Smoothness    & 4.24 & \textbf{4.47} & 4.13 & 4.00 & 4.13 & \underline{3.60} \\
  & Temporal Flickering  & \textbf{4.74} & 4.64 & 4.54 & 4.35 & 4.61 & \underline{4.21} \\
  & Inertial Consistency & 3.18 & \textbf{3.31} & 3.07 & 3.27 & \underline{2.72} & 3.01 \\
\midrule
\multirow[t]{4}{*}{B: Logic \& Physics}
  & Physical Mechanics   & \textbf{3.45} & 3.12 & 3.08 & 3.09 & \underline{2.73} & 2.97 \\
  & Object Permanence    & \textbf{4.41} & 4.40 & 4.09 & 4.07 & 4.30 & \underline{3.70} \\
  & Human Fidelity       & \textbf{4.15} & 3.67 & 3.81 & 3.71 & 3.61 & \underline{2.94} \\
  & Dynamic Degree       & 4.22 & 4.16 & \textbf{4.25} & 4.13 & 3.91 & \underline{3.79} \\
\midrule
\multirow[t]{3}{*}{C: Adherence}
  & Semantic Adherence   & \textbf{4.53} & 4.39 & 4.05 & 4.32 & \underline{3.90} & 4.02 \\
  & Spatial Relationship & 4.38 & \textbf{4.43} & 4.18 & 4.14 & 4.09 & \underline{3.93} \\
  & Semantic Drift       & \textbf{4.83} & 4.78 & 4.65 & 4.73 & 4.68 & \underline{4.61} \\
\midrule
\multirow[t]{4}{*}{D: Aesthetic}
  & Composition \& Framing    & \textbf{4.68} & 4.66 & 4.62 & 4.50 & 4.58 & \underline{4.31} \\
  & Lighting \& Volumetric    & 4.00 & \textbf{4.14} & 4.08 & 3.88 & \underline{3.47} & 3.67 \\
  & Color Harmony             & \textbf{4.86} & 4.84 & 4.73 & 4.75 & 4.70 & \underline{4.59} \\
  & Structural Gestalt        & \textbf{3.98} & 3.90 & 3.60 & 3.59 & 3.61 & \underline{3.24} \\
\midrule
\multicolumn{2}{l}{\textbf{Overall Average}} &
  \textbf{4.32} & 4.27 & 4.10 & 4.07 & 4.01 & \underline{3.82} \\
\bottomrule
\end{tabular}
\end{table}

\paragraph{Physics is the primary gap.}
Inertial Consistency (range: 2.72--3.31) and Physical Mechanics (range: 2.73--3.45)
are the two lowest-scoring dimensions across all models, confirmed on 50 prompts.
Even the top-ranked \veo\ scores only 3.18 and 3.45 respectively, near the scale
midpoint and far below the 4.0+ scores typical of aesthetic and coherence dimensions.
\hunyuan\ is the weakest on Inertial Consistency (2.72) and Physical Mechanics (2.73),
while \veo\ leads on Physical Mechanics (3.45) and \kling\ leads on Inertial Consistency (3.31).
This pattern aligns with the findings of VideoPhy~\citep{bansal2024videophy}.

\paragraph{Aesthetics are strong and uniform.}
Color Harmony (range: 4.59--4.86) and Semantic Drift (range: 4.61--4.83) show the
smallest cross-model variance, suggesting that modern video models have largely converged
on aesthetic quality and prompt coherence.

\paragraph{Human Fidelity is a differentiator.}
\veo\ leads on Human Fidelity (4.15) while \wans\ scores only 2.94, a gap of 1.21
points on the same scale. This dimension has the highest practical relevance for
commercial video applications. Prompts without human subjects are correctly excluded
from this dimension's average.

\paragraph{Lighting anomaly.}
\hunyuan\ scores substantially lower on Lighting \& Volumetric (3.47) than all other
models (range: 3.67--4.14), suggesting a specific architectural or training weakness
in rendering light interaction. This gap is consistent across all 50 prompts.

\paragraph{Semantic adherence: cross-method validation.}
To validate our VLM-based Semantic Adherence scores independently, we computed
prompt video cosine similarity using \textbf{Gemini Embedding~2}~\citep{geminiteam2023}%
\footnote{Specifically \texttt{gemini-embedding-exp-03-07}; cited under the Gemini family paper as no dedicated technical report is available.}
as a reference-free proxy: each video and its corresponding enhanced prompt are
embedded jointly, and alignment is measured by cosine similarity.
Table~\ref{tab:sa_validation} compares the two methods across all 300 video--prompt
pairs (50 prompts $\times$ 6 models).
The per-model rankings agree strongly (Spearman $\hat{\rho} = 0.943$,
$p = 0.005$, $n{=}6$ models), differing only in the \veo/\kling\ swap at ranks 1--2.

\begin{table}[t]
\centering
\caption{Cross-method validation of Semantic Adherence (50 prompts $\times$ 6 models).
  \emph{Emb SA} = mean prompt--video cosine similarity via Gemini Embedding~2
  (reference-free, averaged over 50 prompts).
  \emph{VLM SA} = mean Likert score for the Semantic Adherence dimension from our VLM questionnaire (1--5 scale, averaged over 50 prompts).
  Rankings agree strongly (Spearman $\hat{\rho} = 0.943$, $p = 0.005$).}
\label{tab:sa_validation}
\small
\setlength{\tabcolsep}{7pt}
\begin{tabular}{lcccc}
\toprule
\textbf{Model} & \textbf{Emb SA} & \textbf{Emb Rank} & \textbf{VLM SA} & \textbf{VLM Rank} \\
\midrule
\kling   & \textbf{0.500} & 1 & 4.39          & 2 \\
\veo     & 0.497          & 2 & \textbf{4.53} & 1 \\
\wanb    & 0.486          & 3 & 4.32          & 3 \\
\ltx     & 0.473          & 4 & 4.05          & 4 \\
\wans    & 0.469          & 5 & 4.02          & 5 \\
\hunyuan & \underline{0.468} & 6 & \underline{3.90} & 6 \\
\midrule
\multicolumn{5}{l}{\small Spearman $\hat{\rho} = 0.943$,\ $p = 0.005$,\ $n{=}6$ models} \\
\bottomrule
\end{tabular}
\end{table}

\section{Predicted Human Alignment Score (PHAS)}
\label{sec:phas}
% ─────────────────────────────────────────────────────────────────────────────

In this section, we present an alternative scoring method which derives robustness from caliberated weights of the dimensions based on human preferences. This strongly 
depends on the kind of survey, in this paper \S~\ref{sec:human_study} we have relied on blind pairwise preference study which makes the weights non-interpretable in a meaningful way. We merely use the caliberated weights as an intermediate step to recover the PHAS model score. 

We remind the reader that our $50$ prompt set is split into $30$ caliberation and $20$ validation sets. PHAS (Eq.~\ref{eq:phas_prompt}) is a calibration-set-fitted weighted average of VLM dimension scores. A non-negative ridge logistic regression is fit on a 30-prompt calibration-set human pairwise annotations to learn
dimension importance weights $w_d$. These weights are then applied to VLM scores on a held-out 20-prompt validation set, to compute the PHAS score. 

\paragraph{PHAS definition.}
Let $s_{m,p,d}$ be the mean Likert score (over 10 questions) for model $m$, prompt $p$,
dimension $d$, and $w_d$ the calibrated weight for dimension $d$.  Let
$\mathcal{D}_p \subseteq \mathcal{D}$ be the applicable dimensions for prompt $p$
(null-suitability dimensions are excluded per prompt).  PHAS is computed per
(model, prompt) pair and averaged across prompts:
\begin{equation}
  \mathrm{PHAS}(m) = \frac{1}{|\mathcal{P}|}\sum_{p\in\mathcal{P}}\,
  \frac{\displaystyle\sum_{d\in\mathcal{D}_p} w_d\,s_{m,p,d}}
       {\displaystyle\sum_{d\in\mathcal{D}_p} w_d}
  \;\times\;\lambda(m,p), \label{eq:phas_prompt}
\end{equation}
where the denominator $\sum_{d\in\mathcal{D}_p}w_d$ proportionally redistributes
weight across whichever dimensions are applicable, ensuring that prompts where
certain dimensions are unsuitable (e.g.\ no humans present $\Rightarrow$
\emph{Human Fidelity} is null) are not artificially deflated relative to prompts
where all dimensions apply.
$\lambda(m,p)$ is a variance penalty (capped at 30\%) that down-weights models with high within-dimension
question variance (signalling VLM uncertainty):
\begin{align}
\lambda(m,p) &= 1 - \min\!\left(0.30,\;
  0.05 \cdot \frac{1}{|\mathcal{D}_p|}\sum_{d \in \mathcal{D}_p} \sigma^{2}_{m,p,d}
  \right), \label{eq:phas-penalty}
\end{align}
where $\sigma^{2}_{m,p,d}$ is the Bessel-corrected sample variance ($\delta{=}1$)
across the 10 individual question scores within dimension $d$, for model $m$ on
prompt $p$ (i.e.\ how much the VLM's 10 Likert scores disagree with each other for
the same video clip and dimension). High $\sigma^2$ indicates VLM inconsistency
within a dimension, interpreted as a signal of uncertainty or hallucination rather than
cross-prompt variance. Because every (model, prompt, dimension) triple is always evaluated with exactly
10 questions, the Bessel correction factor ($\tfrac{N}{N-1}{=}\tfrac{10}{9}$)
is a universal constant that scales all $\sigma^2$ values equally and therefore
does not affect relative rankings, but does correct for the downward bias in
sample variance estimates at small~$N$. The 0.05 scaling factor calibrates the penalty to observed variance magnitudes. The mean per-dimension within-run $\sigma^2$ across the 6 models ranges from ~0.81 to ~1.20. Multiplying by 0.05 maps this range into a raw penalty of ~4–6\% as reported in Table~\ref{tab:variance_penalty}. This keeps the penalty a secondary signal, a small but real correction and not a dominant factor. 

\paragraph{Calibration.}
For each pairwise annotation $(m_A, m_B, p)$:
\begin{itemize}[noitemsep,topsep=2pt]
  \item \textbf{Feature vector} $\mathbf{x} \in \mathbb{R}^{16}$: the per-dimension
    VLM score difference $x_d = s_{m_A,p,d} - s_{m_B,p,d}$, i.e.\ how much higher
    model~$A$ scored than model~$B$ on each dimension for that prompt.
  \item \textbf{Label} $y \in \{0,1\}$: whether a human annotator preferred model~$A$
    ($y{=}1$) or model~$B$ ($y{=}0$) in the blind pairwise comparison.
  \item \textbf{Sample weight}: confidence level of the vote
    (Much/Clearly/Slightly better $\to$ 3/2/1).
    These enter the L-BFGS-B objective as per-sample multipliers on the binary
    cross-entropy term, so a ``Much better'' vote contributes $3\times$ to the
    gradient update; they do \emph{not} appear as additional input features.
    Annotators differ substantially in how they use the confidence scale
    (Table~\ref{tab:annotators}; \S~\ref{sec:human_study}):
    e.g.\ \texttt{A5} marks only 5.2\% of votes as ``Much better'' vs.\ 62.6\%
    for \texttt{A7}.
    This heterogeneity is handled correctly: the weighting
    is applied \emph{per comparison}, not \emph{per annotator}, so no annotator
    is globally excluded or down-weighted, only their least-confident individual
    votes receive less loss weight.
\end{itemize}
The regression learns which dimension score gaps most reliably predict human
preference and the fitted non-negative coefficients become the dimension weights~$w_d$.
We fit a non-negative constrained ridge logistic regression on the
1{,}653 calibration annotations and apply an L2 penalty on the pre-normalisation weights
$\tilde{w}_d$. This prevents overfitting across the 16 parameters. The inverse regularisation
strength $C{=}1/\lambda{=}0.1$ is selected by 5-fold cross validation from the set
$\{0.001,0.01,0.1,1,10,100\}$ following standard practice (smaller $C$ leads to stronger shrinkage).
We employ a standard L-BFGS-B solver, which natively enforces $\tilde{w}_d \geq 0$ via
bound constraints. A dimension only receives $\tilde{w}_d > 0$ if scoring higher on it is associated with winning human pairwise comparisons; dimensions with no such signal are automatically set to $\tilde{w}_d{=}0$ rather than assigned a spurious negative weight.
The fitted $\tilde{w}_d$ are then normalised to sum to~1 to produce the final
dimension weights $w_d$ used in Eq.~\eqref{eq:phas_prompt}.

The fitted weights are evaluated in two ways.
First, \textbf{pairwise prediction accuracy}: given a held-out human vote on the
validation set, does the model correctly predict which video won?
Accuracy on the 1{,}043 held-out validation-set annotations is \textbf{62.6\%}
(5-fold cross validation accuracy on the calibration set: 64.4\%; chance baseline: 50\%).
Second, \textbf{PHAS ranking}: the calibrated $w_d$ are applied to VLM dimension
scores on the held-out 20 validation prompts to produce per-model PHAS scores.
Evaluating on the held-out split in both cases ensures the weights are not
overfit to the data used to learn them. Table~\ref{tab:phas_weights_calibrated} reports the full calibrated weight vector. 

\paragraph{Recover BT ranking from PHAS scores}
The full annotation pool (all 50 prompts, 2{,}696 votes) has Krippendorff
$\alpha=0.273$ (which is only considered ``Fair''~\citep{landis1977} for binary preference data), the calibration
subset (30 prompts, 1{,}653 votes) drives the weight fitting.
While this is numerically low and it is difficult to place importance on the individual extracted weights themselves in table ~\ref{tab:phas_weights_calibrated}, the PHAS formula \eqref{eq:phas_prompt} with the fitted weights reproduces the canonical BT ranking (ordering in decreasing order of BT ratings) on the validation set of 20 prompts. This reproduction of the BT ranking by PHAS is the main result of this section, summarized in Table~\ref{tab:variance_penalty}. 

\paragraph{Scene Consistency, Object Permanence, Human Fidelity, Dynamic Degree, and Spatial Relationship collapse to $w=0$.} 
This does not mean that these dimensions are irrelevant to video quality, but that they did not provide a
statistically significant marginal signal for human preference beyond what the non-zero dimensions already
captured, and are correctly suppressed rather than assigned spurious negative weights.
This zeroing is \emph{robust to the choice of optimiser}: replacing the L2 penalty with an elastic-net
penalty (L1$+$L2 blend, mixing ratio tuned by 5-fold cross-validation) yields the same five zero-weight dimensions and
nearly identical non-zero weights, so this is not an artifact of 
a specific regularisation method.

\begin{table}[h]
\centering
\caption{Calibrated PHAS dimension weights from non-negative ridge logistic regression
  (calibration-set annotations, 5-fold cross validation, val.\ accuracy 62.6\%).
  Dimensions with $w=0$ received negative raw coefficients and were clipped.}
\label{tab:phas_weights_calibrated}
\small
\setlength{\tabcolsep}{5pt}
\begin{tabular}{llc}
\toprule
\textbf{Group} & \textbf{Dimension} & \textbf{Calibrated $w$} \\
\midrule
\multirow{5}{*}{A: Motion \& Stability}
  & Subject Consistency     & 0.091 \\
  & Scene Consistency       & 0.000 \\
  & Motion Smoothness       & 0.074 \\
  & Temporal Flickering     & 0.014 \\
  & Inertial Consistency    & 0.054 \\
\midrule
\multirow{4}{*}{B: Logic \& Physics}
  & Physical Mechanics      & 0.058 \\
  & Object Permanence       & 0.000 \\
  & Human Fidelity          & 0.000 \\
  & Dynamic Degree          & 0.000 \\
\midrule
\multirow{3}{*}{C: Instruction Adherence}
  & Semantic Adherence & 0.264 \\
  & Spatial Relationship    & 0.000 \\
  & Semantic Drift          & 0.126 \\
\midrule
\multirow{4}{*}{D: Aesthetic Quality}
  & Composition \& Framing  & 0.130 \\
  & Lighting \& Volumetric  & 0.053 \\
  & Color Harmony           & 0.082 \\
  & Structural Gestalt      & 0.054 \\
\midrule
& \textit{Sum} & \textit{1.000} \\
\bottomrule
\end{tabular}
\end{table}

\begin{table}[t]
\centering
\caption{PHAS scores (validation-20 prompts, calibrated weights), variance penalty
  $\lambda(m)$, and VLM BT rating (all 50 prompts) for comparison.
  PHAS rank order using the caliberated weights ~\ref{tab:phas_weights_calibrated} evaluated on 20 validation prompts reproduces the VLM BT rank.}
\label{tab:variance_penalty}
\small
\setlength{\tabcolsep}{5pt}
\begin{tabular}{lcccccc}
\toprule
\textbf{Model} & \textbf{PHAS} & \textbf{PHAS Rank} & \textbf{VLM BT (50)} & \textbf{VLM BT Rank} & $\boldsymbol{\Delta}$ & \textbf{Var Penalty} \\
\midrule
\veo     & \textbf{4.229} & 1 & \textbf{1652} & 1 & \textbf{0} & 4.1\% \\
\kling   & 4.124          & 2 & 1628          & 2 & \textbf{0} & 4.5\% \\
\wanb    & 4.063          & 3 & 1509          & 3 & \textbf{0} & 5.0\% \\
\ltx     & 3.959          & 4 & 1504          & 4 & \textbf{0} & 5.1\% \\
\hunyuan & 3.771          & 5 & 1433          & 5 & \textbf{0} & 5.3\% \\
\wans    & \underline{3.705} & 6 & \underline{1274} & 6 & \textbf{0} & 6.0\% \\
\bottomrule
\end{tabular}
\end{table}

\paragraph{Variance penalty is inversely correlated with model quality.}
The per-model variance penalties in Table~\ref{tab:variance_penalty} range from
4.1\% (\veo) to 6.0\% (\wans), and are monotonically ordered by model rank.
The best model (\veo) is also the most \emph{consistently} evaluated by the VLM
auditor, while the weakest (\wans) shows the highest within-dimension score variance.
This negative correlation between quality and evaluator uncertainty suggests that
as generative models improve, VLMs find their outputs easier to score consistently. 
This is a useful property for benchmark calibration and a secondary validation of the
variance penalty as a meaningful signal beyond pure score averaging.

\paragraph{PHAS vs.\ BT rating: interpretability vs.\ competitiveness.}
PHAS and BT ratings encode complementary notions of model quality.
The BT rating is derived from a \emph{log-odds} model where small differences in
head-to-head win-rate are amplified into large point spreads, hence the 378-point
gap between Veo and Wan~1.3B (table \ref{tab:leaderboard}).
PHAS is a \emph{weighted arithmetic mean} on the original 1--5 Likert scale: the
same models differ by only 0.52 PHAS points, a more conservative and directly
interpretable gap.
For ranking purposes both are equivalent due to zero rank discordances. PHAS is more intuitive because it stays 
on the native evaluation scale. Because PHAS is a weighted average of VLM scores, its ranking is already stable at 20 validation prompts consistent with A3 ablation, Appendix~\ref{app:abl_nprompts}. 

% ─────────────────────────────────────────────────────────────────────────────
%\FloatBarrier
% ─────────────────────────────────────────────────────────────────────────────
\section{Ablation Studies}
\label{sec:ablation}
% ─────────────────────────────────────────────────────────────────────────────

We present three ablation studies that stress-test the core design decisions
of \worldjen. \S~\ref{sec:abl_cross_vlm} covers VLM auditor cross validation across closed source API's. \S~\ref{sec:abl_reliability} covers reruns and variance tests while \S~\ref{sec:abl_gemma4} explores open source alternatives which are structurally close to the main auditor. Prompt-enhancement impact~(A1), question-count sensitivity~(A2),
and BT rating stability over prompt count~(A3) are addressed in
Appendix~\ref{sec:abl_enhance}, \ref{app:abl_qcount}, and \ref{app:abl_nprompts} respectively.

\subsection{A4 — VLM Auditor Cross-Validation (Gemini vs.\ Claude)}
\label{sec:abl_cross_vlm}
% ─────────────────────────────────────────────────────────────────────────────

\textbf{Motivation.}
\worldjen\ uses Gemini~3~Flash as its VLM auditor throughout.
A natural question is whether rankings are overfit to Gemini's specific
hallucination patterns or perceptual biases.  If a different frontier VLM
produces the same relative ordering of models, the framework is
\emph{auditor-independent}.

\textbf{Setup.}
We re-ran VLM evaluation on 20 validation prompts $\times$ 6 models
(120 videos) using Claude Sonnet (\texttt{claude-sonnet-4-6}) via the Anthropic
API.  The 20-prompt validation subset was chosen to manage API cost while
remaining sufficient for rank comparison: A3 (\S~\ref{app:abl_nprompts})
shows that $N{=}20$ yields a mean Spearman $\hat{\rho}=0.944$ vs.\ the full-50
gold ranking, which is adequate for detecting rank disagreements between auditors.  
All experimental variables that could affect relative rankings were held
fixed, such as identical VQA questions, identical
dimension-aware sampling modes (holistic / sampled / micro, with the same
per-dimension assignments as the main study), and a system prompt deliberately
aligned with Gemini's wording.  Null-suitability dimensions were skipped per
prompt, mirroring the main evaluation protocol.

One necessary technical difference arises from API design: Gemini's
\texttt{generate\_content} API accepts PIL \texttt{Image} objects natively,
while Claude's \texttt{messages} API requires frames to be base64-encoded JPEG
strings.  Both pipelines share the same upstream frame extraction code
(OpenCV~\texttt{cv2.VideoCapture} $\to$ PIL~\texttt{Image}); only the
serialisation step differs.  This has no effect on which frames are selected or
on the score scale, as both auditors receive the same pixel content. In Table~\ref{tab:abl_cross_vlm} we compare the BT rating ordering of Gemini vs.\ Claude on the same validation prompts. In table ~\ref{tab:abl_cross_vlm_dims} we compare the per dimension scores across both the auditors. 

\begin{table}[t]
\centering
\caption{A4: Claude vs.\ Gemini average scores and BT rating on the same 20 validation prompts.
$\Delta_\text{avg}$ = Claude$-$Gemini average score.
BT MLE \eqref{eq:bt_rating} are computed independently within each auditor using the same
Bradley-Terry MLE as in earlier sections.}
\label{tab:abl_cross_vlm}
\small
\begin{tabular}{lrrrrrr}
\toprule
Model & \multicolumn{2}{c}{Avg score} & $\Delta_\text{avg}$ & \multicolumn{2}{c}{BT rating} & Rank \\
\cmidrule(lr){2-3}\cmidrule(lr){5-6}
 & Claude & Gemini & & Claude & Gemini & (C / G) \\
\midrule
Veo 3.1          & 3.25 & 4.28 & $-$1.03 & 1685 & 1673 & 1 / 1 \\
Kling v2.6 Pro   & 3.21 & 4.21 & $-$1.00 & 1603 & 1605 & 2 / 2 \\
Wan v2.2 14B     & 3.17 & 4.09 & $-$0.92 & 1579 & 1518 & 3 / 3 \\
LTX-2            & 3.17 & 4.08 & $-$0.92 & 1565 & 1512 & 4 / 4 \\
HunyuanVideo     & 2.95 & 3.92 & $-$0.98 & 1302 & 1405 & 5 / 5 \\
Wan 1.3B         & 2.97 & 3.74 & $-$0.78 & 1266 & 1288 & 6 / 6 \\
\bottomrule
\end{tabular}
\end{table}

\begin{table}[h]
\centering
\caption{A4: per-model, per-dimension scores for Claude vs.\ Gemini on
the same 20 validation prompts.
$\Delta$ = Claude$-$Gemini.  $\rho$ = Spearman rank correlation across the 6 models
within each dimension ($K{=}6$; $p < 0.05$ marked $^*$).
}
\label{tab:abl_cross_vlm_dims}
\resizebox{\linewidth}{!}{%
\scriptsize
\setlength{\tabcolsep}{4pt}
\begin{tabular}{llrrrrrrrrrrrrrrrrrr}
\toprule
 & & \multicolumn{3}{c}{V (Veo 3.1)} & \multicolumn{3}{c}{K (Kling)} & \multicolumn{3}{c}{L (LTX-2)} & \multicolumn{3}{c}{W14 (Wan 14B)} & \multicolumn{3}{c}{H (Hunyuan)} & \multicolumn{3}{c}{W1 (Wan 1.3B)} \\
\cmidrule(lr){3-5}\cmidrule(lr){6-8}\cmidrule(lr){9-11}\cmidrule(lr){12-14}\cmidrule(lr){15-17}\cmidrule(lr){18-20}
Dimension & $\rho$ & C & G & $\Delta$ & C & G & $\Delta$ & C & G & $\Delta$ & C & G & $\Delta$ & C & G & $\Delta$ & C & G & $\Delta$ \\
\midrule
\multicolumn{20}{l}{\textit{Group A — Motion \& Stability}} \\
\midrule
Subject Cons. & $0.94^*$ & 3.25 & 4.54 & $-$1.29 & 3.25 & 4.70 & $-$1.45 & 3.12 & 4.31 & $-$1.19 & 3.19 & 4.33 & $-$1.14 & 3.21 & 4.62 & $-$1.41 & 2.98 & 4.22 & $-$1.24 \\
Scene Cons.   & $0.54$   & 3.31 & 4.74 & $-$1.43 & 3.27 & 4.66 & $-$1.39 & 3.20 & 4.47 & $-$1.27 & 3.17 & 4.14 & $-$0.97 & 2.91 & 4.49 & $-$1.58 & 3.13 & 4.53 & $-$1.40 \\
Motion Smooth.& $0.84^*$ & 3.24 & 4.15 & $-$0.91 & 3.23 & 4.19 & $-$0.96 & 3.21 & 4.15 & $-$0.94 & 3.13 & 3.85 & $-$0.72 & 2.91 & 4.04 & $-$1.13 & 2.89 & 3.30 & $-$0.41 \\
Temp. Flicker.& $0.66$   & 3.10 & 4.67 & $-$1.57 & 3.34 & 4.71 & $-$1.37 & 3.17 & 4.54 & $-$1.37 & 3.15 & 4.39 & $-$1.24 & 3.19 & 4.59 & $-$1.40 & 3.04 & 4.05 & $-$1.01 \\
Inertial Cons.& $0.60$   & 2.81 & 3.33 & $-$0.52 & 2.57 & 3.02 & $-$0.45 & 2.65 & 2.96 & $-$0.31 & 2.56 & 3.27 & $-$0.71 & 2.40 & 2.62 & $-$0.22 & 2.45 & 3.00 & $-$0.55 \\
\midrule
\multicolumn{20}{l}{\textit{Group B — Logic \& Physics}} \\
\midrule
Phys. Mech.   & $0.89^*$ & 2.69 & 3.40 & $-$0.71 & 2.55 & 2.83 & $-$0.28 & 2.64 & 2.91 & $-$0.27 & 2.58 & 3.10 & $-$0.52 & 2.35 & 2.45 & $-$0.10 & 2.58 & 2.82 & $-$0.24 \\
Object Perm.  & $0.37$   & 2.90 & 4.16 & $-$1.26 & 2.89 & 4.06 & $-$1.17 & 2.93 & 4.14 & $-$1.21 & 2.86 & 4.03 & $-$1.17 & 2.82 & 4.25 & $-$1.43 & 2.72 & 3.36 & $-$0.64 \\
Human Fid.    & $0.49$   & 2.97 & 3.99 & $-$1.02 & 2.97 & 4.01 & $-$1.04 & 3.05 & 3.84 & $-$0.79 & 2.98 & 3.95 & $-$0.97 & 2.90 & 3.65 & $-$0.75 & 2.59 & 2.99 & $-$0.40 \\
Dynamic Deg.  & $0.77$   & 3.17 & 4.35 & $-$1.18 & 3.04 & 4.19 & $-$1.15 & 3.33 & 4.25 & $-$0.92 & 2.96 & 4.03 & $-$1.07 & 2.84 & 3.83 & $-$0.99 & 2.99 & 3.71 & $-$0.72 \\
\midrule
\multicolumn{20}{l}{\textit{Group C — Instruction Adherence}} \\
\midrule
Semantic Adh. & $0.94^*$ & 3.76 & 4.54 & $-$0.78 & 3.61 & 4.41 & $-$0.80 & 3.53 & 4.08 & $-$0.55 & 3.59 & 4.42 & $-$0.83 & 3.18 & 3.74 & $-$0.56 & 3.23 & 4.01 & $-$0.78 \\
Spatial Rel.  & $0.83$   & 3.22 & 4.33 & $-$1.11 & 3.17 & 4.29 & $-$1.12 & 3.18 & 4.00 & $-$0.82 & 3.15 & 4.13 & $-$0.98 & 2.98 & 3.71 & $-$0.73 & 2.98 & 3.67 & $-$0.69 \\
Semantic Drft.& $0.89^*$ & 3.90 & 4.81 & $-$0.91 & 3.80 & 4.79 & $-$0.99 & 3.58 & 4.58 & $-$1.00 & 3.87 & 4.73 & $-$0.86 & 3.42 & 4.58 & $-$1.16 & 3.42 & 4.62 & $-$1.20 \\
\midrule
\multicolumn{20}{l}{\textit{Group D — Aesthetic Quality}} \\
\midrule
Comp.\&Frame. & $-0.03$  & 3.74 & 4.55 & $-$0.81 & 3.69 & 4.53 & $-$0.84 & 3.50 & 4.74 & $-$1.24 & 3.56 & 4.56 & $-$1.00 & 3.32 & 4.69 & $-$1.37 & 3.25 & 4.11 & $-$0.86 \\
Lighting.     & $0.83$   & 3.18 & 4.08 & $-$0.90 & 3.30 & 4.17 & $-$0.87 & 3.14 & 4.09 & $-$0.95 & 3.16 & 3.93 & $-$0.77 & 2.75 & 3.35 & $-$0.60 & 2.96 & 3.61 & $-$0.65 \\
Color Harm.   & $0.83$   & 3.67 & 4.90 & $-$1.23 & 3.65 & 4.89 & $-$1.24 & 3.55 & 4.78 & $-$1.23 & 3.71 & 4.80 & $-$1.09 & 3.30 & 4.61 & $-$1.31 & 3.46 & 4.66 & $-$1.20 \\
Struct. Gest. & $0.77$   & 3.15 & 3.91 & $-$0.76 & 3.00 & 3.83 & $-$0.83 & 2.91 & 3.46 & $-$0.55 & 3.06 & 3.68 & $-$0.62 & 2.70 & 3.58 & $-$0.88 & 2.78 & 3.19 & $-$0.41 \\
\bottomrule
\end{tabular}%
}
\end{table}

\textbf{Findings.}
\begin{enumerate}

\item \textbf{Relative ordering preserved.} Table~\ref{tab:abl_cross_vlm} shows that
both auditors produce the same ordering relative to the three-tier partition produced by the human evals/Gemini VLM \S~\ref{sec:vlm_human_comparison}. Spearman $\hat{\rho} = 1.000, p = 0.0014$ on BT ratings and
$\rho = 0.886,~ p = 0.019$ on average scores confirms this is not an artifact of the BT aggregation. Claude preserves the correct tier \emph{ordering} but with a compressed BT rating spread,
suggesting lower per-prompt discriminability. This is consistent with 20-prompt bootstrap CIs being too wide to statistically
resolve tier boundaries for either auditor at this sample size.

\item \textbf{Claude is systematically stricter.}
Table~\ref{tab:abl_cross_vlm_dims} shows that Claude scores are uniformly lower
than Gemini's across all 16 dimensions ($\Delta \in [-1.34, -0.36]$).  The
largest gaps appear in \emph{Scene Consistency} ($\Delta=-1.34$) and
\emph{Temporal Flickering} ($\Delta=-1.33$), which are the high-ceiling dimensions where
Gemini awards near-perfect scores more readily.  The smallest gaps are in
\emph{Physical Mechanics} ($\Delta=-0.36$) and \emph{Inertial Consistency}
($\Delta=-0.46$), dimensions where both auditors agree models are near-floor,
leaving less room for disagreement.  Notably, the physics dimensions (Group
B) have the smallest absolute deltas, confirming that both auditors share a
consensus on the most challenging aspects of video generation. 

\item \textbf{Dimension-level rank agreement.}
Per-dimension Spearman correlations (6 model scores per dimension) reveal a
clear pattern (Table~\ref{tab:abl_cross_vlm_dims}).
Semantically structured dimensions achieve the highest inter-auditor agreement:
\emph{Semantic Adherence} ($\rho=0.943$, $p=0.005$) and
\emph{Subject Consistency} ($\rho=0.943$, $p=0.005$).
Physics dimensions also agree strongly: \emph{Physical Mechanics} ($\rho=0.886^*$)
and \emph{Semantic Drift} ($\rho=0.886^*$).
Two Group~D aesthetic dimensions show moderate but non-significant agreement: \emph{Lighting}
($\rho=0.829$, $p=0.058$) and \emph{Color Harmony} ($\rho=0.829$, $p=0.058$), but
\emph{Composition \& Framing} is the notable exception ($\rho=-0.03$,
$p=0.957$), suggesting that the relevant scores are sensitive to the specific prompt sample at this scale.
The weakest agreement is in Object Permanence ($\rho=0.37$), Human Fidelity
($\rho=0.49$), and Scene Consistency ($\rho=0.54$). These are all dimensions where absolute score levels differ the most between auditors.
Cross-referencing A5 (Table~\ref{tab:hallucination}): \emph{Structural Gestalt}
($\rho=0.77$, $p=0.072$) shows borderline agreement here \emph{and}
4th-highest within-run variance in A5 (43.1\% of instances with $\sigma^2{>}1.5$),
consistent with genuine perceptual ambiguity.
\emph{Temporal Flickering} ($\rho=0.66$, $p=0.156$) remains non-significant
but improved compared to the prior 50-prompt estimate, consistent with moderate
agreement that stabilises with more data.

\item \textbf{Implication for rankings.}
The scale offset between the two VLMs is consistent and global; it cancels out
in the pairwise Bradley-Terry comparisons that drive BT ratings. The \worldjen\ leaderboard is therefore \emph{robust to auditor choice} at the relative ordering of BT ratings and in general at the \emph{tier level} as shown in Table~\ref{tab:abl_cross_vlm}. In particular, the only practical consequence is a uniform shift in absolute Likert scores (~$-$1.0\,pt with Claude) that does not alter any tier boundary.

\end{enumerate}

\textbf{Conclusion.}
Claude \textbf{preserves the full three-tier structure} with rank order identical to Gemini
($\rho = 1.00$, $p < 0.001$; Table~\ref{tab:abl_cross_vlm}).
The systematic strictness offset ($\Delta_\text{avg}\approx -1.0$\,pt) shifts absolute
Likert scores uniformly but does not move any model across a tier boundary.
Claude's inter-tier BT rating gaps are smaller (${\approx}24$~pts T/M vs.\ $87$~pts for Gemini;
$36$~pts M/B vs.\ $117$~pts), confirming lower per-prompt discriminability but identical
tier assignments. The \worldjen\ leaderboard is therefore robust to auditor choice.

% ─────────────────────────────────────────────────────────────────────────────
\subsection{A5 — VLM Evaluation Uncertainty}
\label{sec:abl_reliability}
% ─────────────────────────────────────────────────────────────────────────────

VLM evaluator uncertainty manifests on two axes: (i)~\emph{within-run
question variance} which estimates inconsistency across the 10 questions on the same video
in the same run and (ii)~\emph{between-run reproducibility} that measures score drift
across multiple VLM evaluations.  We quantify both below.

\subsubsection{A5 (i) Within-run question variance.}
For each (model, prompt, dimension) triple, the 10 Likert scores are used to compute
$\sigma^2$ (Bessel-corrected).  We report the \emph{percentage of (model, prompt)
instances} where $\sigma^2 > 1.5$ per dimension (Table~\ref{tab:hallucination}).

Within-run spread arises from two distinct sources.
\textbf{(i) Partial prompt adherence.}
The VQA generator (available in the code release) is explicitly instructed
to produce questions spanning \emph{distinct} aspects of each dimension: expected
prompt elements, failure modes, success modes, and adversarial probes.
For a complex prompt, each of the 10 questions targets a different requirement producing high spread that
accurately reflects \emph{partial adherence} rather than evaluator confusion.
This is the dominant effect for multi-element dimensions such as Semantic Adherence
(54.2\%) and Physical Mechanics (52.8\%), as can be explicitly checked from the VQA questionnaire in the open sourced dataset.
\textbf{(ii) True VLM oscillation (hallucination).}
For some specific (model, prompt) pairs, the VLM genuinely oscillates on the
\emph{same} property, scoring 5 on one phrasing of a question and 1 on another
about the identical observable. These extreme cases ($\sigma^2 > 4$, see for example Appendix~\ref{app:casestudy_hallucination}) reflect evaluator instability
on ambiguous or occluded content.

Both effects are captured by the $\sigma^2 > 1.5$ threshold, but source~(i) dominates
the aggregate statistics. The dimension mean used in PHAS faithfully aggregates the partial-adherence signals
from source~(i), as confirmed by the strong VLM--human rank correlation
(\S~\ref{sec:human_study}); source~(ii) contributes noise that the 10-question mean
mitigates by averaging over the oscillation.

\begin{table}[h]
\centering
\caption{A5(i)-- Per-dimension within-run question variance.
  \textbf{\% instances}: percentage of (model, prompt) pairs where
  within-run score variance $\sigma^2 > 1.5$ (Bessel-corrected, $n{=}599$
  pairs per dimension).  High rates primarily reflect partial prompt-element
  adherence across the 10 distinct VQA questions;
  Group A = Motion \& Stability; B = Logic \& Physics;
  C = Instruction Adherence; D = Aesthetic Quality.}
\label{tab:hallucination}
\small
\setlength{\tabcolsep}{5pt}
\begin{tabular}{llr}
\toprule
\textbf{Group} & \textbf{Dimension} & \textbf{\% instances} \\
\midrule
C & Semantic Adherence    & 54.2 \\
B & Physical Mechanics    & 52.8 \\
B & Dynamic Degree        & 47.2 \\
D & Structural Gestalt    & 43.1 \\
D & Lighting \& Volumetric & 40.6 \\
C & Spatial Relationship  & 38.5 \\
A & Inertial Consistency  & 33.6 \\
B & Human Fidelity        & 28.9 \\
C & Semantic Drift        & 25.9 \\
B & Object Permanence     & 25.6 \\
A & Subject Consistency   & 22.7 \\
D & Composition \& Framing & 20.4 \\
A & Motion Smoothness     & 14.0 \\
A & Scene Consistency     & 11.5 \\
A & Temporal Flickering   &  9.8 \\
D & Color Harmony         &  6.0 \\
\bottomrule
\end{tabular}
\end{table}

\begin{table}[h]
\centering
\small
\caption{A5(ii)-- True VLM oscillation per-dimension counts of extreme within-run
  variance ($\sigma^2 > 4.0$, Bessel-corrected).  Only 43 of 4{,}716 total
  (model, prompt, dimension) instances (0.9\%) exceed this threshold,
  confirming that genuine VLM self-contradiction is rare.
  Dimensions with zero extreme cases are omitted.
  Group A = Motion \& Stability; B = Logic \& Physics;
  C = Instruction Adherence; D = Aesthetic Quality.}
\label{tab:extreme_variance}
\small
\setlength{\tabcolsep}{5pt}
\begin{tabular}{llrr}
\toprule
\textbf{Group} & \textbf{Dimension} & \textbf{Count ($\sigma^2{>}4$)} & \textbf{\% of instances} \\
\midrule
B & Physical Mechanics    & 9 & 3.0 \\
C & Spatial Relationship  & 8 & 2.7 \\
C & Semantic Adherence    & 5 & 1.7 \\
D & Lighting \& Volumetric & 5 & 1.7 \\
B & Object Permanence     & 3 & 1.0 \\
B & Dynamic Degree        & 3 & 1.0 \\
C & Semantic Drift        & 3 & 1.0 \\
D & Structural Gestalt    & 3 & 1.0 \\
A & Subject Consistency   & 1 & 0.3 \\
A & Motion Smoothness     & 1 & 0.3 \\
A & Inertial Consistency  & 1 & 0.3 \\
\midrule
\multicolumn{2}{l}{\textbf{Total}} & \textbf{43} & \textbf{0.9} \\
\bottomrule
\end{tabular}
\end{table}

Three patterns emerge from Table~\ref{tab:hallucination}.
First, \textbf{Group~A} motion dimensions (Scene Consistency 11.5\%,
Motion Smoothness 14.0\%, Temporal Flickering 9.8\%) have the lowest rates
because the VQA questions for these dimensions probe a relatively binary,
perceptually unambiguous signal, either the scene warps or it does not,
either frames skip or they do not, leaving little room for partial adherence.
Second, \textbf{Group~B physics dimensions} and \textbf{Semantic Adherence}
cluster at the top (Physical Mechanics 52.8\%, Semantic Adherence 54.2\%):
these dimensions involve many distinct sub-requirements per prompt, so partial
adherence across those sub-requirements is the expected outcome for current models.
Third, even the highest-ranked model (\veo) reaches 58\% on Physical Mechanics
and 35\% on Inertial Consistency, confirming that high within-run spread on
physics dimensions is a \emph{task-level} property of complex prompts,
not a weakness of any particular model.

Across all dimensions, true VLM oscillation (source~ii) is rare: only 43 of
4{,}716 instances (0.9\%) exceed $\sigma^2{>}4$
(Table~\ref{tab:extreme_variance}), with Physical Mechanics (9~cases, 3.0\%)
and Spatial Relationship (8~cases, 2.7\%) most affected.
The hallucination case studies in Appendix~\ref{app:casestudy_hallucination}
are drawn from this tail.

\subsubsection{A5 (ii) -- Between-run reproducibility.}
VLM evaluators are also stochastic across API calls: the same pipeline rerun
on a different occasion may return slightly different scores.  To quantify
this, we ran the full VLM evaluation pipeline twice on all 50
prompts $\times$ 6 models (300 videos, 16 dimensions each), using the same
pinned model (\texttt{gemini-3-flash-preview}), VQA questions, and
sampling parameters, with no shared random seed between runs.

\begin{table}[t]
\centering
\small
\caption{A5(ii): Run-to-run reliability: per-model mean Likert scores and rankings
  for Run~1 and Run~2 (50 prompts $\times$ 6 models $\times$ 16 dimensions).
  Paired $t$-test $p$-values reported per model; all non-significant at $\alpha{=}0.01$, signaling VLM rerun reliability.}
\label{tab:abl_reliability_model}
\setlength{\tabcolsep}{6pt}
\begin{tabular}{@{}lrrrrrr@{}}
\toprule
\textbf{Model} & \textbf{Run 1} & \textbf{Run 2} & $\Delta$ & \textbf{$p$-value} & \textbf{Rank 1} & \textbf{Rank 2} \\
\midrule
\veo     & 4.320 & 4.325 & $+$0.005 & 0.829 & 1 & 1 \\
\kling   & 4.275 & 4.279 & $+$0.004 & 0.848 & 2 & 2 \\
\wanb    & 4.074 & 4.084 & $+$0.010 & 0.669 & 4 & 3 \\
\ltx     & 4.110 & 4.078 & $-$0.032 & 0.190 & 3 & 4 \\
\hunyuan & 4.016 & 4.001 & $-$0.015 & 0.549 & 5 & 5 \\
\wans    & 3.831 & 3.884 & $+$0.053 & 0.044 & 6 & 6 \\
\bottomrule
\end{tabular}
\end{table}

\begin{table}[t]
\centering
\small
\caption{A5(ii): Bradley-Terry rating with 95\% bootstrap CI (1,000 prompt-level
  resamples, same MM algorithm as canonical analysis) for each independent run.
  Both runs preserve the three-tier point-estimate ordering.
  Tier separation is expressed as the mean gap between adjacent tiers relative
  to the CI half-width of the boundary model (${\approx}{\pm}60$~pts):
  Run~1 T/M $= 119$~pts (${\approx}1.9{\times}$~CI), M/B $= 159$~pts (${\approx}2.9{\times}$~CI), matching the canonical analysis;
  Run~2 T/M $= 76$~pts (${\approx}1.3{\times}$~CI), M/B $= 101$~pts (${\approx}1.7{\times}$~CI), weaker but positive, consistent with run-to-run VLM stochasticity.}
\label{tab:abl_reliability_bt}
\setlength{\tabcolsep}{5pt}
\begin{tabular}{@{}lcrrrrrrr@{}}
\toprule
\textbf{Model} & \textbf{Tier} &
  \multicolumn{3}{c}{\textbf{Run 1 }} &
  \multicolumn{3}{c}{\textbf{Run 2 }} \\
\cmidrule(lr){3-5}\cmidrule(lr){6-8}
 & & BT rating & \multicolumn{2}{c}{95\% CI} & BT rating & \multicolumn{2}{c}{95\% CI} \\
\midrule
\veo     & T & 1652.4 & [1590, & 1728] & 1640.0 & [1584, & 1712] \\
\kling   & T & 1627.9 & [1571, & 1697] & 1596.1 & [1541, & 1659] \\
\wanb    & M & 1509.0 & [1454, & 1569] & 1520.4 & [1458, & 1587] \\
\ltx     & M & 1503.8 & [1444, & 1560] & 1438.5 & [1383, & 1492] \\
\hunyuan & M & 1432.7 & [1374, & 1485] & 1467.7 & [1411, & 1520] \\
\wans    & B & 1274.2 & [1212, & 1323] & 1337.2 & [1273, & 1391] \\
\bottomrule
\end{tabular}
\end{table}

\begin{table}[h]
\centering
\small
\caption{A5(ii): Run-to-run reliability: per-dimension paired $t$-test between Run~1
  and Run~2 (4{,}710 matched (model, prompt, dimension) pairs).
  Only two of 16 dimensions reach $p < 0.05$; all effect sizes ($|\Delta|$) are
  below 0.09 on the 1--5 scale.  ``ns'' = not significant ($\alpha{=}0.05$).}
\label{tab:abl_reliability_dim}
\setlength{\tabcolsep}{5pt}
\begin{tabular}{@{}llrrrr@{}}
\toprule
\textbf{Group} & \textbf{Dimension} & \textbf{Run 1} & \textbf{Run 2} & $\Delta$ & \textbf{Sig.} \\
\midrule
\multirow{5}{*}{A} & Subject Consistency   & 4.330 & 4.273 & $-$0.057 & ns \\
                   & Scene Consistency     & 4.614 & 4.592 & $-$0.022 & ns \\
                   & Motion Smoothness     & 4.096 & 4.118 & $+$0.021 & ns \\
                   & Temporal Flickering   & 4.516 & 4.556 & $+$0.039 & ns \\
                   & Inertial Consistency  & 3.093 & 3.180 & $+$0.087 & ns \\
\midrule
\multirow{4}{*}{B} & Physical Mechanics    & 3.073 & 3.062 & $-$0.010 & ns \\
                   & Object Permanence     & 4.162 & 4.172 & $+$0.010 & ns \\
                   & Human Fidelity        & 3.650 & 3.668 & $+$0.018 & ns \\
                   & Dynamic Degree        & 4.077 & 4.067 & $-$0.010 & ns \\
\midrule
\multirow{3}{*}{C} & Semantic Adherence    & 4.209 & 4.148 & $-$0.061 & $*$ \\
                   & Spatial Relationship  & 4.190 & 4.253 & $+$0.063 & ns \\
                   & Semantic Drift        & 4.713 & 4.697 & $-$0.016 & ns \\
\midrule
\multirow{4}{*}{D} & Composition \& Framing  & 4.562 & 4.546 & $-$0.016 & ns \\
                   & Lighting \& Volumetric  & 3.875 & 3.963 & $+$0.088 & $*$ \\
                   & Color Harmony           & 4.742 & 4.729 & $-$0.013 & ns \\
                   & Structural Gestalt      & 3.653 & 3.605 & $-$0.048 & ns \\
\bottomrule
\end{tabular}
\end{table}

\textbf{Overall agreement.}
Across 4{,}710 matched (model, prompt, dimension) triplets, the two runs
achieve Pearson $r = 0.756$ and Spearman $\hat{\rho} = 0.743$, with
MAE $= 0.44$ and RMSE $= 0.67$ on the 1--5 scale.
The intraclass correlation coefficient ICC(3,1) $= 0.870$, exceeding the
conventional ``good reliability'' threshold of 0.75~\citep{koo2016guideline}.
Global means are near-identical: $4.104$ (Run~1) vs.\ $4.109$ (Run~2).

\textbf{Dimension stability.}
Table~\ref{tab:abl_reliability_dim} shows that 14 of 16 dimensions have
non-significant divergences under a paired $t$-test ($\alpha = 0.05$).  The two
significant dimensions, \emph{Semantic Adherence} ($\Delta = -0.06$,
$p = 0.018$) and \emph{Lighting \& Volumetric} ($\Delta = +0.09$,
$p = 0.021$) have effect sizes below 0.1 points, far below the threshold
for practical significance on a 5-point scale.

\textbf{Rank stability.}
Table~\ref{tab:abl_reliability_model} shows that model-level scores are
virtually identical across runs (all $|\Delta| \leq 0.053$; 5 of 6 models
non-significant at $\alpha = 0.01$).  Model rankings are stable at positions
1, 2, 5, and 6; the only change across the two runs is a swap between \ltx\ and \wanb\
at ranks 3 and 4, a statistically indistinguishable mid-tier pair.  Rank-order
Spearman $\hat{\rho} = 0.943$ ($p = 0.005$) across the two runs.

Table~\ref{tab:abl_reliability_bt} confirms the three-tier structure in
Bradley-Terry ratings with bootstrap CIs.
Both runs assign identical tier memberships in point estimates: \veo\ and \kling\
lead (top tier), \wans\ trails (bottom tier), and the three mid-tier models occupy
ranks 3--5.
Following the same metric used in the canonical analysis
(\S~\ref{sec:vlm_eval}), tier separation is measured as the mean BT rating gap between
adjacent tiers relative to the CI half-width.
Run~1 reproduces the canonical result: T/M gap $\approx 119$~pts
(${\approx}1.9{\times}$ CI, matching the canonical $1.9{\times}$);
M/B gap $\approx 159$~pts (${\approx}2.9{\times}$ CI).
Run~2 shows the same tier ordering in point estimates with somewhat weaker tier
separation: T/M gap $\approx 76$~pts (${\approx}1.3{\times}$ CI),
M/B gap $\approx 101$~pts (${\approx}1.7{\times}$ CI), consistent with
run-to-run VLM stochasticity.

\textbf{Conclusion.}
Independent reruns \textbf{preserve the full three-tier structure} with identical tier
membership across point estimates (Table~\ref{tab:abl_reliability_bt}).
Run~2 shows somewhat weaker tier separation due to run-to-run VLM stochasticity,
but no tier-boundary violations occur in either run.
ICC$(3,1) = 0.87$ confirms that between-tier score differences reflect genuine
quality differences rather than VLM sampling noise.

% ─────────────────────────────────────────────────────────────────────────────
\subsection{A6 — Open-Source VLM Auditor: Gemma~4 vs.\ Gemini~3~Flash}
\label{sec:abl_gemma4}
% ─────────────────────────────────────────────────────────────────────────────

\textbf{Motivation.}
The prior ablations (A4--A5) establish that \worldjen\ is robust to the choice
between two \emph{closed-source} frontier VLMs (Gemini, Claude).  A more
practically important question is whether the pipeline can be replicated without
any proprietary API dependency.  We therefore run the full evaluation on all
50 prompts $\times$ 6 models using \textbf{Gemma~4} (31B; \texttt{gemma-4-31b-it}),
a state-of-the-art open-weight multimodal model, and compare its BT ratings
and PHAS scores against Gemini~3~Flash and the human BT ground truth.
The experiment was conducted on 8$\times$H200 GPUs in batch mode and completed in ${\approx}5.15$\,hrs.

\textbf{Implementation note.}  Because Gemma~4 runs locally without API rate limits, inference
uses greedy decoding (\texttt{do\_sample=False}, $T{=}0$), making scores fully deterministic
and reproducible across runs.  This contrasts with Gemini~3~Flash, which uses stochastic
sampling ($T{=}1.0$) via the API and the resulting run-to-run score variance in Gemini was
characterized in the previous section (\S~\ref{sec:abl_reliability}).  All other evaluation settings
(frame extraction, VQA questions, rubrics, dimension modes) are identical between the two auditors.

\textbf{Results.}

\begin{table}[h]
\centering
\small
\caption{A6: Open-source (Gemma~4) vs.\ closed-source (Gemini~3~Flash) vs.\ Human BT.
BT ratings are BT-MLE point estimates and the 95\% CI from 1,000 prompt-level bootstrap resamples.
Rows are ordered by Human BT rank.
Spearman $\hat{\rho}$: Gemma~4 vs.\ Human $= 0.771$ ($p = 0.072$);
Gemini vs.\ Human $= 1.000$ ($p < 0.001$).
Tier labels: \textbf{T}op, \textbf{M}id, \textbf{B}ottom.}
\label{tab:abl_gemma4}
\setlength{\tabcolsep}{4pt}
\resizebox{\linewidth}{!}{%
\begin{tabular}{@{}lcrrrrrrrrr@{}}
\toprule
\textbf{Model} & \textbf{Tier} &
  \multicolumn{3}{c}{\textbf{Gemma~4 BT}} &
  \multicolumn{3}{c}{\textbf{Gemini BT}} &
  \textbf{Human BT} &
  \textbf{PHAS} & \textbf{PHAS} \\
\cmidrule(lr){3-5}\cmidrule(lr){6-8}
 & & BT rating & \multicolumn{2}{c}{95\% CI} & BT rating & \multicolumn{2}{c}{95\% CI} & BT rating & G4 & Gem \\
\midrule
\veo    & T & 1613.2 & [1567, & 1661] & 1652 & [1590, & 1728] & 1614.2 & 4.188 & 4.229 \\
\kling  & T & 1613.3 & [1568, & 1665] & 1628 & [1571, & 1697] & 1571.8 & 4.077 & 4.124 \\
\wanb   & M & 1461.9 & [1412, & 1507] & 1509 & [1454, & 1569] & 1517.8 & 3.945 & 4.063 \\
\ltx    & M & 1483.7 & [1445, & 1522] & 1504 & [1444, & 1560] & 1479.1 & 3.934 & 3.959 \\
\hunyuan& M & 1477.0 & [1432, & 1519] & 1433 & [1374, & 1485] & 1461.7 & 3.744 & 3.771 \\
\wans   & B & 1350.9 & [1295, & 1398] & 1274 & [1212, & 1323] & 1355.4 & 3.684 & 3.705 \\
\bottomrule
\end{tabular}}
\end{table}

\begin{table}[h]
\centering
\caption{A6: Per-model, per-dimension scores for Gemini~3~Flash (G) and Gemma~4 (G4), with
$\Delta = \text{G4} - \text{G}$.  Models ordered by human BT rank.
$\rho$ = Spearman rank correlation of the 6-model ordering between G and G4 for that dimension
($^*p < 0.05$).}
\label{tab:abl_gemma4_dims}
\resizebox{\linewidth}{!}{%
\scriptsize
\setlength{\tabcolsep}{4pt}
\begin{tabular}{llrrrrrrrrrrrrrrrrrr}
\toprule
 & & \multicolumn{3}{c}{V (Veo 3.1)} & \multicolumn{3}{c}{K (Kling)} & \multicolumn{3}{c}{W14 (Wan v2.2)} & \multicolumn{3}{c}{L (LTX-2)} & \multicolumn{3}{c}{H (Hunyuan)} & \multicolumn{3}{c}{W1 (Wan 2.1)} \\
\cmidrule(lr){3-5}\cmidrule(lr){6-8}\cmidrule(lr){9-11}\cmidrule(lr){12-14}\cmidrule(lr){15-17}\cmidrule(lr){18-20}
Dimension & $\rho$ & G & G4 & $\Delta$ & G & G4 & $\Delta$ & G & G4 & $\Delta$ & G & G4 & $\Delta$ & G & G4 & $\Delta$ & G & G4 & $\Delta$ \\
\midrule
\multicolumn{20}{l}{\textit{Group A — Motion \& Stability}} \\
\midrule
Subject Cons. & $0.83$   & 4.59 & 4.21 & $-$0.38 & 4.59 & 4.01 & $-$0.58 & 4.23 & 3.87 & $-$0.36 & 4.12 & 3.74 & $-$0.38 & 4.52 & 4.14 & $-$0.38 & 3.94 & 3.59 & $-$0.35 \\
Scene Cons.   & $0.94^*$ & 4.82 & 3.98 & $-$0.85 & 4.73 & 3.96 & $-$0.78 & 4.34 & 3.48 & $-$0.86 & 4.65 & 3.60 & $-$1.04 & 4.61 & 3.51 & $-$1.10 & 4.53 & 3.56 & $-$0.97 \\
Motion Smooth.& $0.89^*$ & 4.24 & 3.90 & $-$0.35 & 4.47 & 3.87 & $-$0.61 & 4.00 & 3.32 & $-$0.68 & 4.13 & 3.56 & $-$0.56 & 4.13 & 3.64 & $-$0.48 & 3.60 & 3.12 & $-$0.48 \\
Temp. Flicker.& $0.94^*$ & 4.74 & 4.89 & $+$0.15 & 4.64 & 4.71 & $+$0.06 & 4.35 & 4.46 & $+$0.10 & 4.54 & 4.70 & $+$0.16 & 4.61 & 4.80 & $+$0.19 & 4.21 & 4.32 & $+$0.11 \\
Inertial Cons.& $0.66$   & 3.18 & 2.99 & $-$0.19 & 3.31 & 2.89 & $-$0.42 & 3.27 & 2.79 & $-$0.47 & 3.07 & 2.86 & $-$0.21 & 2.72 & 2.64 & $-$0.09 & 3.01 & 2.63 & $-$0.38 \\
\midrule
\multicolumn{20}{l}{\textit{Group B — Physics \& Objects}} \\
\midrule
Phys. Mech.   & $0.94^*$ & 3.45 & 3.00 & $-$0.45 & 3.12 & 2.81 & $-$0.32 & 3.09 & 2.69 & $-$0.41 & 3.08 & 2.72 & $-$0.36 & 2.73 & 2.48 & $-$0.24 & 2.97 & 2.51 & $-$0.46 \\
Object Perm.  & $0.54$   & 4.41 & 3.65 & $-$0.77 & 4.40 & 3.77 & $-$0.63 & 4.07 & 3.41 & $-$0.66 & 4.09 & 3.29 & $-$0.80 & 4.30 & 3.97 & $-$0.33 & 3.70 & 3.43 & $-$0.27 \\
Human Fid.    & $0.89^*$ & 4.15 & 3.47 & $-$0.68 & 3.67 & 3.22 & $-$0.45 & 3.71 & 3.31 & $-$0.40 & 3.81 & 3.26 & $-$0.55 & 3.61 & 3.24 & $-$0.38 & 2.94 & 2.74 & $-$0.21 \\
Dynamic Deg.  & $0.60$   & 4.22 & 3.22 & $-$1.00 & 4.16 & 3.48 & $-$0.68 & 4.13 & 3.30 & $-$0.83 & 4.25 & 3.78 & $-$0.47 & 3.91 & 3.33 & $-$0.58 & 3.79 & 3.14 & $-$0.65 \\
\midrule
\multicolumn{20}{l}{\textit{Group C — Semantic Fidelity}} \\
\midrule
Semantic Adh. & $0.94^*$ & 4.53 & 4.54 & $+$0.01 & 4.39 & 4.49 & $+$0.09 & 4.32 & 4.43 & $+$0.10 & 4.05 & 4.32 & $+$0.27 & 3.90 & 4.10 & $+$0.20 & 4.02 & 4.09 & $+$0.07 \\
Spatial Rel.  & $0.77$   & 4.38 & 4.29 & $-$0.09 & 4.43 & 4.20 & $-$0.22 & 4.14 & 4.09 & $-$0.04 & 4.18 & 4.16 & $-$0.01 & 4.09 & 4.03 & $-$0.06 & 3.93 & 4.09 & $+$0.17 \\
Semantic Drft.& $0.77$   & 4.83 & 4.70 & $-$0.13 & 4.78 & 4.63 & $-$0.14 & 4.73 & 4.59 & $-$0.14 & 4.65 & 4.50 & $-$0.15 & 4.68 & 4.64 & $-$0.04 & 4.61 & 4.55 & $-$0.06 \\
\midrule
\multicolumn{20}{l}{\textit{Group D — Aesthetic Quality}} \\
\midrule
Comp.\&Frame. & $0.60$   & 4.68 & 4.67 & $-$0.01 & 4.66 & 4.59 & $-$0.07 & 4.50 & 4.40 & $-$0.09 & 4.62 & 4.28 & $-$0.35 & 4.58 & 4.36 & $-$0.22 & 4.31 & 4.28 & $-$0.03 \\
Lighting.     & $0.94^*$ & 4.00 & 3.81 & $-$0.19 & 4.14 & 3.94 & $-$0.20 & 3.88 & 3.59 & $-$0.29 & 4.08 & 3.60 & $-$0.48 & 3.47 & 3.32 & $-$0.15 & 3.67 & 3.38 & $-$0.28 \\
Color Harm.   & $0.83$   & 4.86 & 4.78 & $-$0.08 & 4.84 & 4.74 & $-$0.10 & 4.75 & 4.63 & $-$0.12 & 4.73 & 4.60 & $-$0.13 & 4.70 & 4.65 & $-$0.05 & 4.59 & 4.47 & $-$0.12 \\
Struct. Gest. & $0.94^*$ & 3.98 & 3.30 & $-$0.68 & 3.90 & 3.11 & $-$0.78 & 3.59 & 2.94 & $-$0.64 & 3.60 & 2.90 & $-$0.70 & 3.61 & 2.94 & $-$0.67 & 3.24 & 2.85 & $-$0.39 \\
\bottomrule
\end{tabular}%
}
\end{table}

\begin{enumerate}

\item \textbf{Gemma~4 achieves moderate rank agreement with human ground truth.}
Spearman $\hat{\rho}(\text{Gemma~4 BT, Human BT}) = 0.771$ ($p = 0.072$),
compared to $\hat{\rho} = 1.000$ ($p < 0.001$) for Gemini.
Gemma~4 preserves the three-tier structure: both top-tier models (\veo, \kling) occupy
ranks~1--2, all three mid-tier models (\wanb, \ltx, \hunyuan) occupy ranks~3--5, and
the bottom-tier model (\wans) remains at rank~6.
Within all the tiers the spread of the models are much smaller, within the CI half widths which is the expected behaviour for models with fully overlapping CIs.

\item \textbf{PHAS scores are consistent across auditors.}
Independently recalibrating PHAS weights for Gemma~4 (same 30-prompt human annotations) yields
$\hat{\rho}(\text{PHAS}_{\text{Gemma4}}, \text{PHAS}_{\text{Gemini}}) = 1.000$ ($p < 0.001$),
with comparable absolute score ranges (Gemma~4: $3.68$--$4.19$; Gemini: $3.71$--$4.23$).

\item \textbf{Open-source gap is concentrated in temporal and physics dimensions.}
Table~\ref{tab:abl_gemma4_dims} reveals two complementary failure modes.
First, Gemma~4's \emph{absolute} scores are substantially lower in Group~A
(Scene Consistency $\bar{\Delta} \approx -0.92$, Motion Smoothness $\bar{\Delta} \approx -0.53$)
and Group~B (Dynamic Degree $\bar{\Delta} \approx -0.73$; Object Permanence $\bar{\Delta} \approx -0.57$).
Second, the per-dimension Spearman $\rho$
between Gemma~4 and Gemini model rankings is lowest for Object Permanence ($\rho = 0.54$),
Dynamic Degree ($\rho = 0.60$), Inertial Consistency ($\rho = 0.66$), and
Composition \& Framing ($\rho = 0.60$). These are all the dimensions requiring fine-grained temporal or physical reasoning.

\item \textbf{Tier boundaries are statistically supported by CIs.}
Following the same metric used in the canonical analysis
(\S~\ref{sec:vlm_eval}), tier separation is the mean BT rating gap between adjacent
tiers relative to the CI half-width of the boundary model as shown in Table~\ref{tab:abl_gemma4}.

\end{enumerate}

\textbf{Conclusion.}
Gemma~4 \textbf{preserves the full three-tier structure} with no tier-boundary violations
(Table~\ref{tab:abl_gemma4}).
However, only 8/16 dimensions achieve statistically significant correlation with Gemini
($\hat{\rho}\geq 0.886$, $p<0.05$; \citep{zar1972spearman}); the 8 non-significant dimensions
are the harder temporal, physics, and spatial ones (Inertial Consistency, Object Permanence,
Dynamic Degree, Spatial Relationship, Semantic Drift, Composition \& Framing, Human Fidelity,
Motion Smoothness), suggesting Gemma~4 reproduces global rankings but is weaker on
fine-grained temporal and spatial reasoning.
% ─────────────────────────────────────────────────────────────────────────────
%\FloatBarrier
\section{Comparison with VBench}
\label{sec:vbench_comparison}
\label{sec:vbench}
% ─────────────────────────────────────────────────────────────────────────────
% ─────────────────────────────────────────────────────────────────────────────

\begin{figure}[t]
\centering
\includegraphics[width=0.95\linewidth]{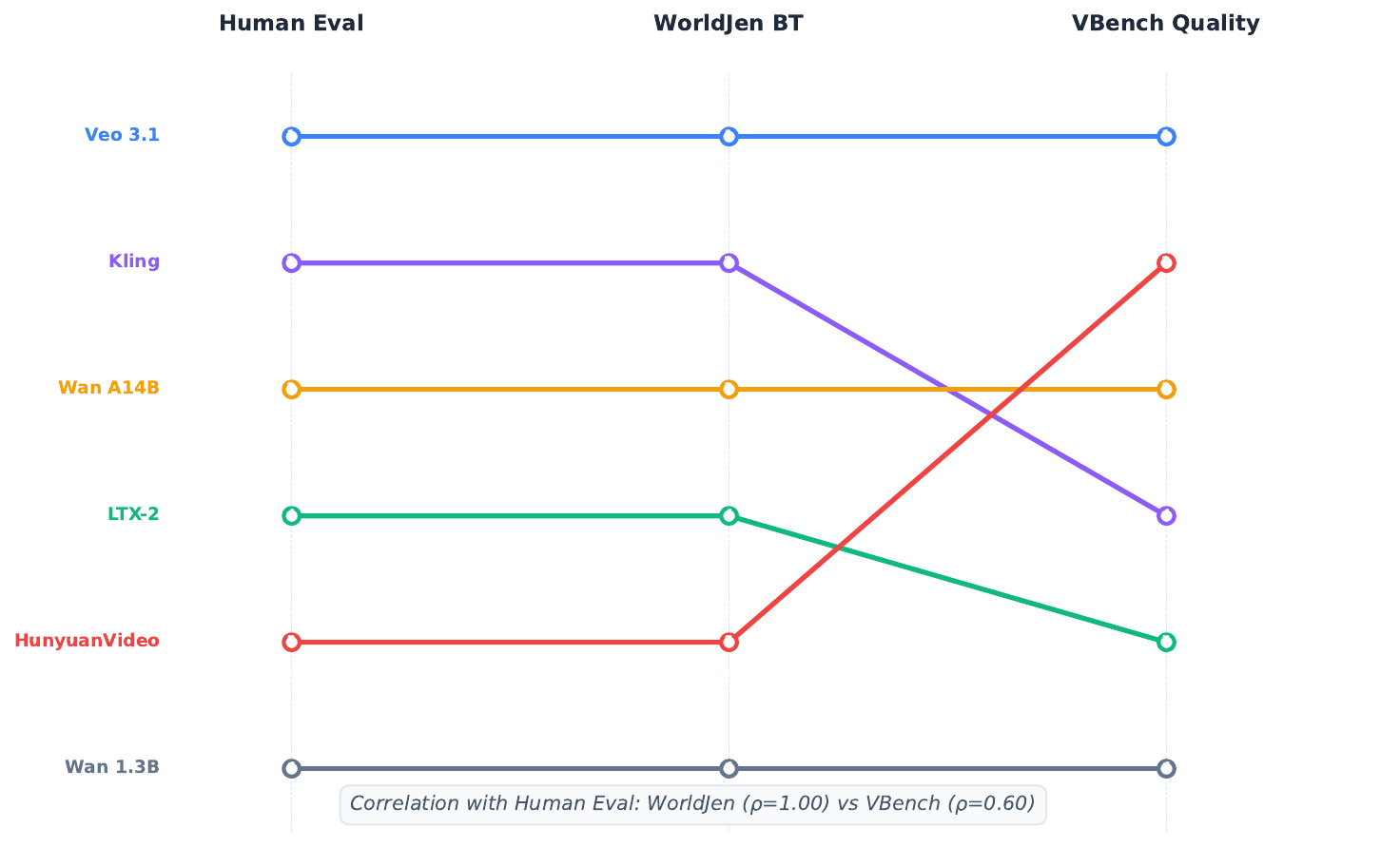}
\caption{\textbf{Model rankings across evaluation methods (bump chart).}
  Each line traces one model across three ranking columns.
  \worldjen\ BT ranking (green background) matches the human-established three-tier structure
  exactly for all six models (15/15 pairwise orderings correct;
  Table~\ref{tab:leaderboard}).
  VBench Quality (red background) violates the tier structure observed by humans and \worldjen \ .
  Badges on the right show each model's rank shift ($\Delta = $ VBench$-$Human rank).
  Summary strips at the bottom report pairwise concordance and Spearman~$\hat{\rho}$.}
\label{fig:vbench_bump}
\end{figure}

\begin{figure}[t]
\centering
\includegraphics[width=\linewidth]{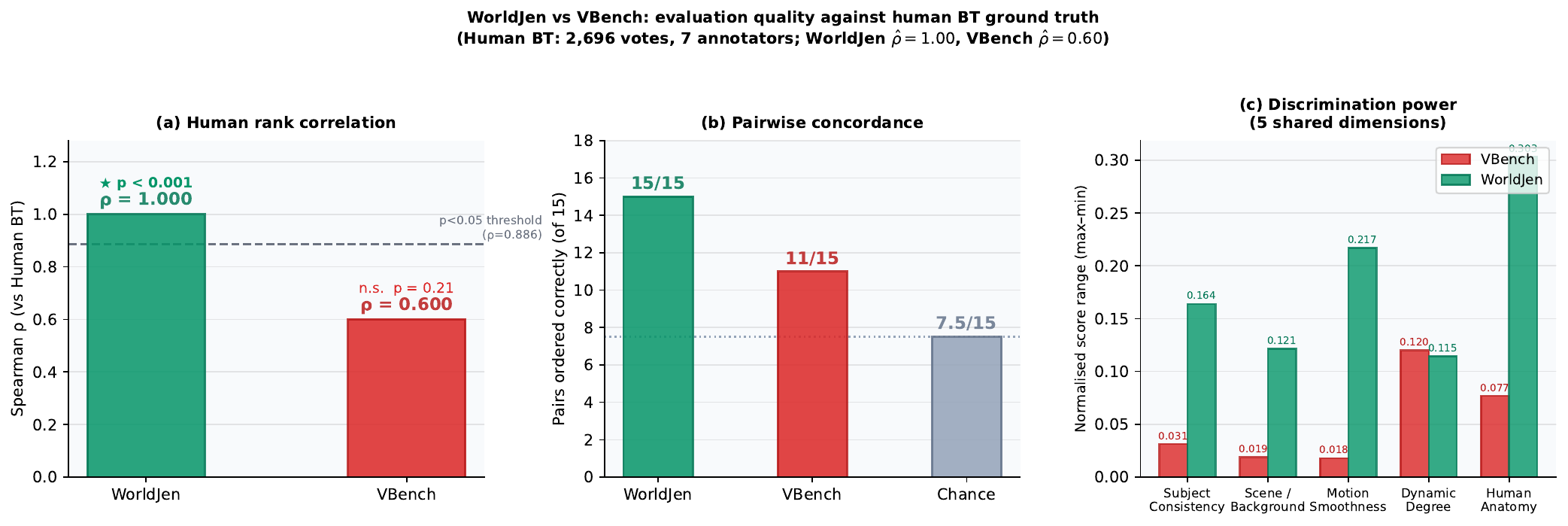}
\caption{\textbf{Three independent measures of evaluation quality}
  \textbf{(a) Pairwise concordance:} \worldjen\ correctly orders all 15 of 15 model pairs
  relative to human judgment (perfect score); VBench orders only 11 of 15, closer to
  the 7.5 chance baseline.
  \textbf{(b) Human rank correlation:} \worldjen\ achieves Spearman $\hat{\rho}=+1.000$
  (perfect tier concordance; $p=0.0014$ at $n=6$); VBench Quality reaches
  $\hat{\rho}=+0.60$ ($p=0.21$, not significant at $n=6$).
  The dashed line marks the $\hat{\rho}=0.886$ threshold required for $p<0.05$ at $n=6$.
  \textbf{(c) Discrimination power:} on the five shared conceptual dimensions,
  \worldjen's normalised score ranges are 1--12$\times$ wider than VBench's (Dynamic Degree being the one exception at $\approx$1$\times$; Motion Smoothness leads at 12$\times$),
  indicating that VBench scores saturate near ceiling and cannot distinguish models
  on the dimensions that matter most to human annotators.}
\label{fig:vbench_evidence}
\end{figure}

We run VBench~\citep{huang2024vbench} on the same 50 \worldjen\ prompts and evaluate
both benchmarks against the human Bradley--Terry ranking as ground truth
(\S~\ref{sec:human_study}; Tables~\ref{tab:leaderboard} and~\ref{tab:rankcompare}).
Table~\ref{tab:vbench} reports all raw VBench per-dimension scores with no
post-hoc adjustment; Figures~\ref{fig:vbench_bump} and~\ref{fig:vbench_evidence}
present the human-alignment comparison; the analysis follows below.

\paragraph{Dimension selection and ranking formula.}
VBench supports a \texttt{custom\_input} mode that evaluates arbitrary user-provided
videos without requiring VBench's standard prompt suite.  The dimensions compatible
with custom input in VBench~v1 are exactly six of the set. We evaluate all six of
\emph{Subject Consistency}, \emph{Background Consistency}, \emph{Motion Smoothness},
\emph{Dynamic Degree}, \emph{Aesthetic Quality}, and \emph{Imaging Quality}.
From VBench~2.0, the custom-input-compatible dimensions include \emph{Human Anatomy},
which we include as the most relevant to our model set.
VBench~2.0 introduces physics-related dimensions (Mechanics, Thermodynamics, Materials)
but these currently do \emph{not} support custom-input evaluation, making them unavailable for
our prompt set.

The VBench quality score and rank (\textbf{VB}$\uparrow$ column in Table~\ref{tab:vbench})
follow the formula described in VBench paper~\citep{huang2024vbench}:
\begin{equation}
  Q_{\mathrm{VB}} = \frac{\displaystyle\sum_{d} w_d\,\hat{s}_d}{\displaystyle\sum_{d} w_d},
  \quad
  \hat{s}_d = \frac{s_d - \min_d}{\max_d - \min_d},
  \label{eq:vbench-quality}
\end{equation}
where $s_d$ is the raw per-dimension score, $(\min_d, \max_d)$ are the normalisation
bounds taken directly from VBench's \texttt{scripts/constant.py} (e.g.\
$[0.706,\,0.998]$ for Motion Smoothness, $[0.146,\,1.0]$ for Subject Consistency),
and the dimension weight $w_d = 0.5$ for Dynamic Degree and $w_d = 1$ for all others.
Models are ranked by $Q_{\mathrm{VB}}$; we apply no post-hoc adjustments. The results are summarized in table \ref{tab:vbench}.

\begin{table}[t]
\centering
\caption{VBench raw per-dimension scores on the 50 \worldjen\ prompts
  (6 models; VBench v1 + VBench 2.0 Human Anatomy).
  \textbf{Bold} = best per column; \underline{underline} = worst.
  \emph{Range} row = max$-$min across models (discrimination proxy).
  Rightmost three columns show the VBench quality rank (by $Q_{\mathrm{VB}}$,
  Eq.~\ref{eq:vbench-quality}), \worldjen\ BT rank, and human BT rank.}
\label{tab:vbench}
\small
\setlength{\tabcolsep}{4pt}
\begin{tabular}{lcccccccrrr}
\toprule
 & \multicolumn{3}{c}{\textit{Temporal consistency ($\Delta<0.04$)}} & \multicolumn{4}{c}{\textit{Quality / motion ($\Delta>0.07$)}} & & & \\
\cmidrule(lr){2-4}\cmidrule(lr){5-8}
\textbf{Model} & \textbf{SubjC} & \textbf{BgC} & \textbf{MotS} & \textbf{DynD} & \textbf{AestQ} & \textbf{ImagQ} & \textbf{HumAn} & \textbf{VB\,$\uparrow$} & \textbf{WJ\,$\uparrow$} & \textbf{Hum\,$\uparrow$} \\
\midrule
\veo     & 0.882 & \textbf{0.929} & 0.983 & \underline{0.880} & \textbf{0.651} & \textbf{0.675} & 0.875 & \textbf{1} & \textbf{1} & \textbf{1} \\
\kling   & 0.881 & \textbf{0.929} & 0.981 & 0.900           & 0.621         & 0.578         & \underline{0.802} & 4 & 2 & 2 \\
\wanb    & 0.875 & 0.928           & \underline{0.971} & 0.960 & 0.599 & 0.615 & \textbf{0.879} & 3 & 3 & 3 \\
\ltx     & \underline{0.853} & \underline{0.911} & 0.976 & \textbf{1.000} & \underline{0.577} & 0.578 & 0.838 & 5 & 4 & 4 \\
\hunyuan & \textbf{0.884} & 0.919 & \textbf{0.987} & 0.980 & 0.592 & 0.590 & 0.844 & 2 & 5 & 5 \\
\wans    & 0.875 & 0.916           & 0.969           & 0.960 & 0.566 & \underline{0.539} & 0.832 & \underline{6} & \underline{6} & \underline{6} \\
\midrule
\textit{Range ($\Delta$)} & \textit{0.031} & \textit{0.019} & \textit{0.018} & \textit{0.120} & \textit{0.086} & \textit{0.136} & \textit{0.077} & & & \\
\bottomrule
\end{tabular}
\end{table}

\begin{table}[t]
\centering
\caption{Normalised score range (max$-$min across 6 models, divided by scale span) for shared
  concepts between VBench and \worldjen.
  VBench range is divided by 1 (scale $[0,1]$); \worldjen\ range is divided by 4 (scale $[1,5]$),
  so both columns are directly comparable on $[0,1]$.
  VBench's near-ceiling scores compress model differences to under 3\,pp on most
  dimensions, eliminating discrimination precisely where \worldjen\ finds the largest spreads.}
\label{tab:vbench_spread}
\small
\setlength{\tabcolsep}{5pt}
\begin{tabular}{lcccc}
\toprule
\textbf{Concept} & \textbf{VBench dim} & \textbf{VB norm.\ range} & \textbf{\worldjen\ dim} & \textbf{WJ norm.\ range} \\
\midrule
Subject consistency   & Subject Consistency    & 0.031 & Subject Consistency & 0.164 \\
Background / scene    & Background Consistency & 0.019 & Scene Consistency   & 0.121 \\
Motion smoothness     & Motion Smoothness      & 0.018 & Motion Smoothness   & 0.217 \\
Motion quantity       & Dynamic Degree         & 0.120 & Dynamic Degree      & 0.115 \\
Human quality         & Human Anatomy (VB2)    & 0.077 & Human Fidelity      & 0.304 \\
\bottomrule
\end{tabular}
\end{table}

\paragraph{Score compression and ranking inversion on shared dimensions.}
Five of the seven evaluated VBench dimensions have direct \worldjen\ counterparts.
Table~\ref{tab:vbench_spread} compares score ranges across the 6~models for each
shared concept.
VBench's reference-free statistics cluster near ceiling. The highest-ranked model and the lowest-ranked model differ by at most 0.031
on a [0,1] scale.
\worldjen's prompt-specific VQA questions on the same concepts span 0.12--0.30
(normalised to $[0,1]$ for comparison), achieving 1--12$\times$ greater discrimination.
When model scores cluster within such a narrow band, per-model differences
fall below measurement noise, making the resulting ranking sensitive to
small fluctuations rather than genuine quality gaps. This offers a plausible explanation for the cross tier inversions observed relative to both \worldjen\ and human judges (see table \ref{tab:vbench} and fig \ref{fig:vbench_bump}).

\paragraph{Analysis.}
Against human BT ratings as ground truth, \worldjen\ BT ratings correctly order
\textbf{15 out of 15} pairwise model comparisons and reproduces all three
human-established tiers without a single tier-boundary violation
($\hat{\rho}=+1.00$, interpreted as tier concordance given $n{=}6$;
Kendall $\hat{\tau}=+1.00$, $p=0.003$).
VBench quality achieves only $\hat{\rho}=+0.60$ ($p=0.21$, not significant
even at $n{=}6$; Kendall $\hat{\tau}=+0.47$, $p=0.27$) and correctly orders
only \textbf{11 out of 15} pairs, approaching the chance level of 7.5
(Figure~\ref{fig:vbench_evidence}).
Critically, VBench's two worst errors are tier-boundary violations, not
within-tier reorderings.

The root cause of VBench's failure is structural, not incidental.
VBench's temporal-consistency dimensions (SubjC, BgC, MotS) are reference-free
statistics computed per frame or per adjacent-frame pair; for modern video generators
all three saturate near 0.97--0.99. When scores are this compressed, ranking noise dominates signal.

An additional contributing factor is input resolution: VBench's underlying metrics
(DINO~\citep{caron2021dino}, CLIP~\citep{radford2021clip}, RAFT~\citep{teed2020raft})
explicitly resize frames to $224{\times}224$ pixels before feature extraction.
By contrast, \worldjen\ extracts frames at the video's native resolution
and passes them directly to the auditor without any
downsampling, the auditor tiles large images at its native resolution rather than squashing
them to a fixed small footprint. This resolution gap is a second structural reason why VBench misses fine-grained
temporal artifacts that are clearly visible to \worldjen's VLM auditor.

% ─────────────────────────────────────────────────────────────────────────────
%\FloatBarrier
% ─────────────────────────────────────────────────────────────────────────────
%\FloatBarrier
% ─────────────────────────────────────────────────────────────────────────────
%\FloatBarrier
\section{Limitations and Future Work}
\label{sec:limitations}
% ─────────────────────────────────────────────────────────────────────────────

\paragraph{Human evaluation scale.}
Human annotation in this work focused on pairwise annotations \S~\ref{sec:human_study}. Dimension specific signals are harder to extract from this data as was observed in \S~\ref{sec:phas} Table~\ref{tab:phas_weights_calibrated}. Dimension-specific human evaluation, where annotators score each video on individual quality
dimensions (e.g.\ motion smoothness, subject consistency) would further enable fine-grained validation of individual \worldjen\
dimensions against human judgment. However, such annotation is considerably more cognitively demanding for annotators,
as it requires repeated viewing of each video and independent rating across multiple dimensions,
making it impractical to scale without specialist recruitment or dedicated crowd-sourcing infrastructure.
Nevertheless, we hope to report on this in the near future.

\paragraph{Single seed per (model, prompt)}
For this work, the primary objective was to validate the \worldjen\ \emph{evaluation framework}
itself establishing that VLM-based scoring reliably tracks human preference and
outperforms reference-free metrics. A single seed per prompt is sufficient for framework validation, the correlation and
concordance metrics that constitute our main claims are robust to per-prompt noise when
aggregated across 50 prompts and 6 models. That said, we all agree that
video generation is inherently stochastic. The same prompt can yield a compelling output on
one seed and a degenerate one on another.
This can inflate variance in the Bradley-Terry matchups and is a potential source
of uncertainty in the current BT ratings. The 95\% bootstrap CI half-widths in Table~\ref{tab:leaderboard} ($\pm$55--69\,pts)
reflect this. Models within a tier that are genuinely close in quality cannot be confidently separated with one seed per prompt.
\textbf{Generating 3--5 seeds per (model, prompt) and averaging scores before the
pairwise matchup} is a natural extension for the next experimental round.
With 3 seeds, the per-prompt score estimate variance drops by $\approx3\times$,
roughly halving the CI half-widths and making the intra-tier rank ordering far more reliable. 

\paragraph{Heterogeneous video duration.}
Veo~3.1 Fast produces ${\approx}8$\,s clips while all other five models produce
${\approx}5$\,s clips a 60\% temporal gap that creates asymmetric biases.

\noindent\textit{Bias against Veo (consistency penalty).}
Every additional second of generation increases the probability of subject mutation,
background warp, or semantic drift.
Holding temporal coherence for 60\% longer is a harder task.
That Veo still leads on Subject Consistency (4.61) and Semantic Drift (4.83)
suggests its margins may be \emph{understated}. In particular, if all models were forced to generate
8\,s, the shorter-clip models would likely see their consistency scores fall.

\noindent\textit{Bias in favour of Veo (narrative advantage).}
Longer videos do give the model more time to execute complex multi-step prompt
instructions, and our data provide partial support for this (see Table~\ref{tab:dimensions}).
Veo ranks \textbf{1st on Semantic Adherence} across all 50 prompts having a
mean 4.53 vs.\ 4.39 compared to Kling. This is consistent with Veo's extra runtime allowing it to 
complete the described action sequence more fully.
Veo also ranks 1st on Semantic Drift with a score of 4.83, further supporting a narrative-completion 
advantage for the longer format on instruction-following dimensions specifically.

\noindent\textit{Built-in control: Micro mode.}
Importantly, our dimension-aware sampling already mitigates the most critical
form of duration bias for the physics and motion dimensions at the centre of our
analysis. \textbf{Micro mode} (§\ref{sec:vlm}) evaluates only the first ${\approx}2$\,s of
every video regardless of total length, sampling every 5th raw frame up to index~60
(${\approx}12$ frames at 30\,fps). Inertial Consistency, Physical Mechanics, Temporal Flickering, and Motion
Smoothness are therefore assessed over an identical temporal window for every
model, hence 8\,s or 5\,s makes no difference. The reported physics gap (top models below 3.5/5) is robust to this.

\noindent\textit{Residual bias in Holistic and Sampled modes.}
For Holistic mode (32 frames) and Sampled mode (16 frames), frames are drawn at
fixed \emph{percentages} of total length.
For a 5\,s video this means one frame every ${\approx}4$ source frames, for an
8\,s video, one every ${\approx}7$.
The larger inter-frame gap for Veo means the VLM sees fewer frames per unit time,
potentially missing high-frequency micro-flickering or sudden jitter that occurs
between sampled frames. This could modestly inflate Veo's smoothness score on holistic dimensions.

\noindent\textit{The truncation dilemma.}
Naively cutting Veo to 5\,s would remove the very footage used to assess semantic
completion, penalising Veo for planning a longer narrative arc, which is a genuine
model capability. We decided against truncation since instruction adherence is a key factor to measure in video benchmarking.
Standardising at 8\,s for all models is an alternative but is unavailable for
models that do not support that duration. Standardising video duration across all compared models remains an open challenge
for multi-model video benchmarking in general.

\paragraph{Benchmark scale and model coverage.}
The current benchmark evaluates 50 prompts across six models.
Ideally 50--200 prompts lower the CI spread significantly for robust BT rating estimation; the scope of this study achieves the lower bound.
With $n{=}6$ models, the Spearman rank correlation has limited statistical power for raw data and perfect rank order agreement in general. Hence $\hat{\rho}=1.000,~p=0.0014 $ is conservatively interpreted as evidence of tier-level concordance between VLM and human evaluations, despite the fact that the underlying scores are derived from incredibly dense data of \textbf{47{,}160} Likert responses.

\paragraph{Feature extensions} New modalities and evaluation paradigms
that broaden the scope of the framework:

\begin{enumerate}[label=(\roman*),leftmargin=1.8em,topsep=3pt,itemsep=2pt]
\item \textbf{Conditional generation modalities (TI2V, TIA2V).}
The current framework targets text-to-video (T2V).
Natural extensions include text-image-to-video (TI2V), where a reference image
anchors the visual style or subject, and text-image-audio-to-video (TIA2V),
where both image and audio conditioning must be respected.
Each modality introduces new evaluation dimensions (e.g.\ image fidelity,
audio-visual synchrony) that require dedicated VQA question sets.

\item \textbf{Video-to-video (V2V) and editing evaluation.}
Models that accept an input video and perform style transfer, inpainting, or
motion retargeting present a distinct evaluation challenge: faithfulness to
the source must be balanced against the quality of the edit.
\worldjen's dimension-aware VQA approach can be adapted to score
edit-specific axes (e.g.\ content preservation, edit adherence).

\item \textbf{World model and action recognition benchmarking.}
As generative video models evolve towards interactive world models capable of
simulating physical environments and responding to agent actions,
evaluation requires new research. The model wide physics inconsistency noted by the VLM judge is possibly related to this and deserves further investigation.

\end{enumerate}

% ─────────────────────────────────────────────────────────────────────────────
%\FloatBarrier
\section{Conclusion}
\label{sec:conclusion}
% ─────────────────────────────────────────────────────────────────────────────

\worldjen\ is an end-to-end benchmark for generative video models built around
prompt-specific VQA questionnaires graded by a VLM judge across 16 quality dimensions,
with rankings derived from Bradley-Terry maximum-likelihood estimation. Applied across six state-of-the-art models, 
\worldjen\ independently reproduces a three-tier BT ranking structure, derived by human annotators. The agreement between the two evaluation modes, perfect rank concordance and significant Spearman correlation validates the VLM-as-a-judge approach as a reliable proxy for human preference.

The Predicted Human Alignment Score (PHAS) complements the BT rating as a lightweight,
interpretable aggregate that distils multi-dimension scores into a single number correlated
with human preference. The VLM judge reveals an industry-wide gap in physical plausibility, even the strongest current models fall well short of flawless execution on physics and motion dimensions, pointing to a clear frontier for future progress.

Cross-VLM ablations confirm that the tier structure is respected by alternative
auditor models. A head-to-head comparison with VBench on the same prompt set demonstrates that \worldjen\ substantially better tracks human judgment, a gap that traces directly to the
discrimination power lost when reference-free metrics cluster near their ceiling.

We release the full curated prompt dataset and evaluation code to support reproducible, human-aligned video generation benchmarking. We expect that scaling our framework will continue to produce stable scores, and that benchmarking costs will go down due to reduction of the number of videos required for statistical significance. An immediate consequence is to apply it in the training pipelines to choose between model checkpoints with similar loss values, or optimization validation. 

% ─────────────────────────────────────────────────────────────────────────────
\section*{Acknowledgements}
% ─────────────────────────────────────────────────────────────────────────────

We thank \textbf{Emir Soyturk}, \textbf{Jeremy Felder}, and \textbf{Roman Palkin},
for engineering contributions to the \worldjen\ product platform. We thank the moonmath.ai research team 
Yuval Domb, Tomer Solberg, and Omer Shlomovits for consistent encouragement, brainstorming ideas and support in this
effort. We are grateful to all annotators who contributed their valuable time and effort to the human evaluation study. 

\paragraph*{LLM/VLM model index.}
Table~\ref{tab:model_index} maps every shorthand model name used in this paper
to its exact API identifier and role in the pipeline.

\begin{table}[t]
\centering
\caption{Complete index of LLM and VLM models used in this work.
  Shorthands used in the paper body, exact API identifiers, and the role each
  model plays in the \worldjen\ pipeline.}
\label{tab:model_index}
\small
\setlength{\tabcolsep}{4pt}
\begin{tabular}{llll}
\toprule
\textbf{Shorthand in paper} & \textbf{API model ID} & \textbf{Provider} & \textbf{Role} \\
\midrule
Gemini 3.1 Flash-Lite & \texttt{gemini-3.1-flash-lite-preview} & Google & Phase A: suitability scoring \& prompt enhancement \\
Gemini 3 Flash        & \texttt{gemini-3-flash-preview}         & Google & Phase B: VQA question generation \\
Gemini 3 Flash        & \texttt{gemini-3-flash-preview}         & Google & Phase B: primary VLM evaluator (main results) \\
Claude Sonnet 4.6     & \texttt{claude-sonnet-4-6}             & Anthropic & A4 ablation: alternative VLM auditor \\
Gemma 4               & \texttt{gemma-4-31b-it}                & Google & A6 ablation: open-source VLM auditor \\
Gemini 2.5 Pro        & \texttt{gemini-2.5-pro}                & Google & Writing \& statistical assistance only \\
Claude Sonnet 4.6     & \texttt{claude-sonnet-4-6}             & Anthropic & Writing \& code assistance only \\
\bottomrule
\end{tabular}
\end{table}

\paragraph*{AI assistance statement.}
The authors used large language and vision-language models as assistants during the preparation of this work.
Specifically, \textbf{Claude Sonnet~4.6} (Anthropic) and \textbf{Gemini~2.5~Pro} (Google DeepMind)
were used for writing assistance, statistical cross-validation, consistency checking of reported numbers,
figure generation and iteration, and LaTeX editing.
\textbf{Gemini~3~Flash} (\texttt{gemini-3-flash-preview}; Google DeepMind) served as the primary VLM evaluator in the \worldjen\
benchmark pipeline (\S~\ref{sec:vlm}), and \textbf{Claude~Sonnet~4.6} (\texttt{claude-sonnet-4-6}) and \textbf{Gemma~4}
were used as alternative VLM evaluators in ablation studies A4 and A6 respectively.
All experimental design decisions, scientific claims, interpretation of results, and final editorial
judgments were made by the human authors.
No AI system is listed as an author; the authors take full responsibility for the content of this work.
% ─────────────────────────────────────────────────────────────────────────────
\bibliographystyle{plainnat}
\bibliography{worldjen}
% ─────────────────────────────────────────────────────────────────────────────
\appendix
% \FloatBarrier

\section{Human Evals}
\label{app:human}
% ══════════════════════════════════════════════════════════════════════════════
\subsection{Annotation Protocol}
\label{app:human_protocol}

This section documents the exact procedure for the human preference study described
in \S~\ref{sec:human_study}.  The study covers all 50 prompts (300 videos) and is deployed
via a Google Apps Script web application.

% ── Step 0: Preparation (Coordinator) ────────────────────────────────────────
\subsubsection*{Step 0 — Preparation (Coordinator)}

\begin{enumerate}[leftmargin=1.8em, topsep=3pt, itemsep=4pt]

  \item \textbf{Confirm assets.}
  VLM evaluation is complete for all \textbf{50~prompts $\times$ 6~models $=$ 300
  videos}.  All 300 video files are stored on Google Drive; the Apps Script backend
  maps each \texttt{(prompt\_id, model)} tuple to a Drive file ID.

  \item \textbf{Pair generation.}
  For each prompt, the interface automatically enumerates all $\binom{6}{2}=15$
  model pairings, yielding \textbf{750 total pairs}.  Left/right assignment is
  randomised independently per pair per session.  Already-completed pairs (stored
  in the annotator's Google Sheet) are filtered out on resume so no pair is shown
  twice to the same annotator.

  \item \textbf{Session design.}
  Each annotator's queue contains all 750~pairs they have not yet personally judged,
  sorted by ascending global coverage (least-reviewed pairs first).
  A break overlay appears every 50~pairs.
  An annotator who completes all their remaining pairs sees an ``All Done'' screen.

  \item \textbf{Access control.}
  Annotators identify themselves by entering their email address at session start.
  The interface normalises emails to lowercase for consistent history lookup.
  Drive folder access is granted at the folder level so the script can serve
  video blobs.  \emph{Note: the publicly released dataset uses anonymized annotator
  IDs (A1--A7) in place of email addresses}.

\end{enumerate}

\begin{figure}[H]
  \centering
  \includegraphics[width=0.95\linewidth]{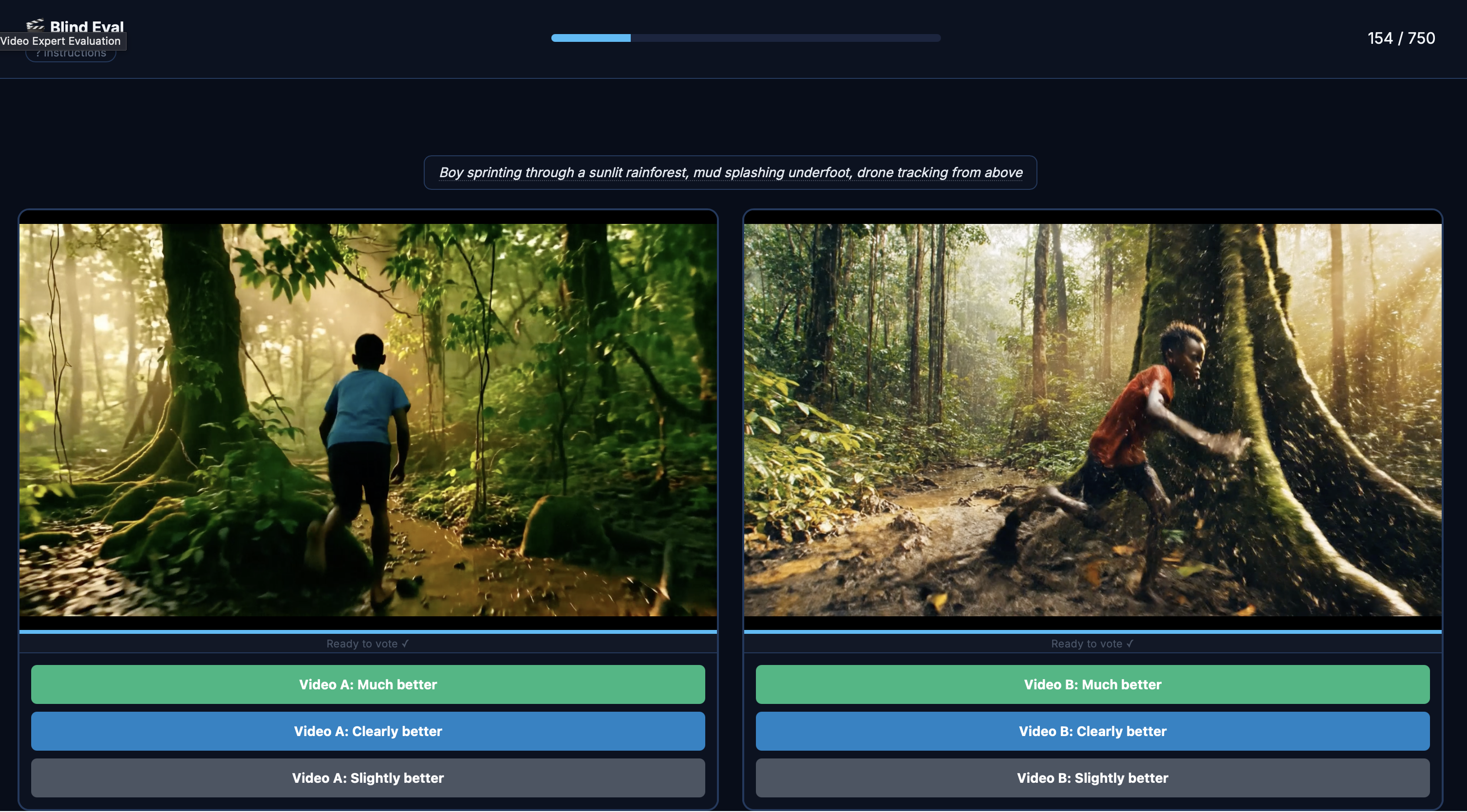}\\[6pt]
  \includegraphics[width=0.95\linewidth]{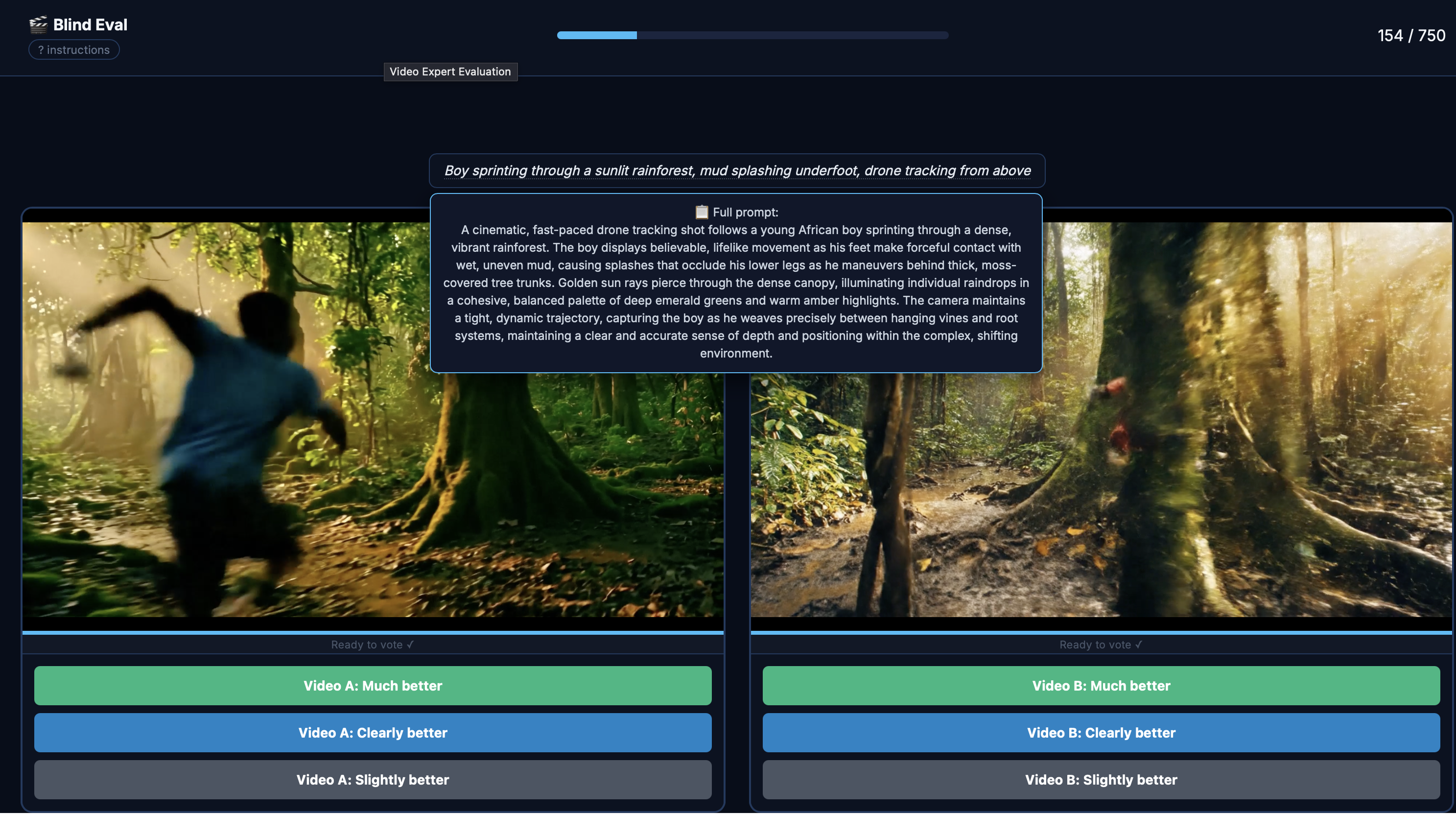}
  \caption{\textbf{Top:} The \worldjen\ blind pairwise evaluation interface.
    Model identities are masked (``Video~A'' / ``Video~B''; left/right assignment
    is re-randomised per pair).
    A short \emph{prompt summary} is displayed as an italic tag above the arena;
    hovering the tag reveals the full enhanced prompt in a tooltip overlay
    (\textbf{Bottom}), allowing annotators to verify exact wording and spatial
    relationships without the full text cluttering the comparison view.
    A 10-second watch timer (progress bar beneath each video) enforces a minimum
    viewing window before vote buttons unlock, preventing premature judgements.
    Vote buttons offer three confidence levels (\emph{Much} / \emph{Clearly} /
    \emph{Slightly better}).
    The progress bar (top centre) and pair counter (top right) track
    per-annotator lifetime progress.
    Responses are appended in real time to Google~Sheets.}
  \label{fig:evalui}
\end{figure}
% ── Step 1: Annotator onboarding ─────────────────────────────────────────────
\subsubsection*{Step 1 — Annotator Onboarding}

On first visit a full-screen instruction card appears automatically.
Annotators must click \emph{``I Understand — Start Evaluating''} to proceed.
The card can be re-opened at any time via the \texttt{? instructions} button in the
header.  The complete text shown to annotators is reproduced below.

%\medskip
\begin{small}
\begin{tabular}{|p{0.93\linewidth}|}
\hline
\vspace{2pt}
\textbf{\textsf{Annotation Instructions}} \\[4pt]
\textbf{1.~Read the prompt first.}
Base your decision on prompt-faithfulness, not visual polish.\\[3pt]
\textbf{2.~Prioritise core action over background.}
Rank requirements mentally: Core Action / Physics $\to$ Characters $\to$ Background.\\[3pt]
\textbf{3.~Symmetric artifacts cancel.}
If both videos flicker or both clip, ignore that and judge what \emph{differs}.\\[3pt]
\textbf{4.~Forced choice — no skips.}
Choose the more faithful attempt even if neither is perfect.
Use \emph{Slightly better} when the margin is thin.\\[6pt]

\textbf{Decision Tree}\\[3pt]
\hspace*{1em}Can you tell the videos apart on prompt faithfulness?\\[1pt]
\hspace*{1.5em}$\mid$\\
\hspace*{1.5em}$\vdash$\; One succeeds where the other clearly fails
  \hfill $\to$ \textbf{Much better}\\
\hspace*{2.4em}{\footnotesize(e.g.\ core action done vs.\ not done; physics violated vs.\ correct)}\\[2pt]
\hspace*{1.5em}$\mid$\\
\hspace*{1.5em}$\vdash$\; Noticeable gap, but both have some flaws
  \hfill $\to$ \textbf{Clearly better}\\
\hspace*{2.4em}{\footnotesize(e.g.\ one gets the action right but background drifts)}\\[2pt]
\hspace*{1.5em}$\mid$\\
\hspace*{1.5em}$\llcorner$\; Marginal / too close to call
  \hfill $\to$ \textbf{Slightly better}\\
\hspace*{2.4em}{\footnotesize(both equally good \emph{or} equally bad — still pick one)}\\[6pt]

\textbf{Confidence Scale}\\[3pt]
\begin{tabular}{@{\hspace{1em}}lcp{0.55\linewidth}@{}}
\toprule
\textbf{Choice} & \textbf{Wt.} & \textbf{When to use} \\
\midrule
Much better     & 3 & One clearly succeeds where the other clearly fails on a core element. \\
Clearly better  & 2 & Noticeable advantage on $\geq\!1$ prompt element; minor flaws on both. \\
Slightly better & 1 & Marginal or no visible difference — forced best-guess.
                      Covers ``barely better'' \emph{and} ``essentially tied.'' \\
\bottomrule
\end{tabular}\\[6pt]

\textbf{Example}\\[3pt]
\textit{Prompt: ``A basketball player spins and dunks while the crowd cheers,
stadium lights flashing.''}\\[3pt]
\textbf{Video A} — smooth motion, but the player clips through the rim
(object permanence failure).\\
\textbf{Video B} — slight flicker, but the dunk lands correctly and the crowd is visible.\\[3pt]
$\Rightarrow$ \textbf{Choose Video B — Clearly better.}
The rim clip violates the core physical action; the flicker is a rendering artifact
that does not contradict the prompt.\\[6pt]

\texttt{Space} = pause/play both videos \quad \texttt{R} = restart both from the beginning\\[2pt]
\hline
\end{tabular}
\end{small}

% ── Step 2: Annotation session ────────────────────────────────────────────────
\subsubsection*{Step 2 — Annotation Session}

\begin{enumerate}[leftmargin=1.8em, topsep=3pt, itemsep=4pt]

  \item \textbf{For each comparison pair:}
  \begin{enumerate}[label=\alph*., topsep=1pt, itemsep=2pt]
    \item Read the \emph{prompt summary} tag shown above both video cards.
          Hover the tag to reveal the full enhanced prompt in a tooltip if needed.
    \item Wait for the loading overlay to disappear — both videos start playing
          simultaneously only after both have fully buffered; vote buttons are
          locked until this point.
    \item Watch both videos for at least \textbf{10~seconds} (enforced by a
          progress bar beneath each video; buttons unlock automatically).
    \item Watch Video~A and Video~B (each loops until a choice is made).
    \item Use \texttt{Space} to pause and compare, \texttt{R} to restart.
    \item Click the button matching your confidence level under the preferred video.
    \item A ``Choice Saved!'' toast confirms the submission; the next pair loads
          automatically.
  \end{enumerate}

  \item \textbf{Session breaks.}
  After every 50~pairs a break overlay appears with the pair count.
  Annotators click \emph{Continue} when ready.  Progress is saved continuously
  to Google Sheets, so closing the browser is safe.

  \item \textbf{Slow connections.}
  If a video does not buffer within 5~seconds, a ``Loading\ldots (slow connection)''
  hint is shown.  After 12~seconds the client automatically re-fetches a fresh
  copy of the video from Drive.

\end{enumerate}

% ── Step 3: Data export and aggregation ──────────────────────────────────────
\subsubsection*{Step 3 — Data Export and Aggregation}

\begin{enumerate}[leftmargin=1.8em, topsep=3pt, itemsep=4pt]

  \item \textbf{Google Sheets format.}
  Each vote is appended as one row with columns:
  \texttt{timestamp, email, prompt\_id, model\_a, model\_b, winner, loser, confidence, source}.
  In the released anonymized dataset, \texttt{timestamp} is
  dropped and \texttt{email} is replaced by an opaque annotator ID (\texttt{A1}--\texttt{A7}).

  \item \textbf{Inter-annotator agreement.}
  For pairs judged by $\geq 2$ annotators, compute mean IAA and Krippendorff's~$\alpha$.

  \item \textbf{Human BT rating.}
  Pool all comparisons and fit an unweighted Bradley-Terry model (each vote
  contributes one win/loss; confidence labels are used only in the PHAS step below)
  to obtain per-model Human BT rating anchored at 1500.
  Report 95\%~bootstrap CIs (1{,}000 resamples).

  \item \textbf{Calibrated PHAS weights.}
  Using the 30-prompt calibration split (1{,}653 annotations; disjoint from the
  20-prompt validation set), fit a non-negative constrained ridge logistic regression:
  \begin{itemize}[topsep=1pt, itemsep=1pt]
    \item Feature vector $\mathbf{x} \in \mathbb{R}^{16}$: per-dimension VLM score
      difference $x_d = s_{m_A,p,d} - s_{m_B,p,d}$ for each applicable dimension;
      null-suitability dimensions are excluded (not zeroed) per prompt.
    \item Label $y \in \{0,1\}$: $y{=}1$ if the annotator preferred $m_A$, $y{=}0$
      otherwise; sample weight = confidence (Much/Clearly/Slightly $\to$ 3/2/1).
    \item Inverse regularisation strength $C{=}1/\lambda{=}0.1$ chosen by 5-fold
      cross-validation from $\{0.001,0.01,0.1,1,10,100\}$.
    \item Output: calibrated non-negative dimension weights $w_d$ (normalised to
      sum to 1), used directly in Eq.~\eqref{eq:phas_prompt}.
  \end{itemize}

  \item \textbf{Validation.}
  Evaluate the calibrated weights on the held-out 20-prompt validation set
  (1{,}043 annotations): report pairwise prediction accuracy and PHAS model ranking
  (Spearman $\hat{\rho}$ vs.\ Human BT rating).

\end{enumerate}

\section{Prompt Curation}
\label{app:prompt_curation}
\label{app:curation}
% ══════════════════════════════════════════════════════════════════════════════
\subsection{Definitions}
\label{app:defs}
% ══════════════════════════════════════════════════════════════════════════════
Table~\ref{tab:dim_defs} provides full definitions for all 16 evaluation dimensions.
\begin{table}[t]
\centering
\caption{Complete dimension taxonomy used in \worldjen.}
\label{tab:dim_defs}
\small
\setlength{\tabcolsep}{5pt}
\begin{tabular}{lll}
\toprule
\textbf{Group} & \textbf{Dimension} & \textbf{Definition} \\
\midrule
\multirow{5}{*}{A: Motion \& Stability}
  & Subject Consistency   & Main character/object maintains consistent shape, colour, identity \\
  & Scene Consistency     & Environment stability without warping or melting during motion/camera \\
  & Motion Smoothness     & Fluid, stutterless motion without jitter or frame skipping \\
  & Temporal Flickering   & Lighting/texture stability without unwanted flashes or brightness changes \\
  & Inertial Consistency  & Objects follow momentum laws with natural acceleration and deceleration \\
\midrule
\multirow{4}{*}{B: Logic \& Physics}
  & Physical Mechanics  & Gravity, friction, collision behave consistently with physical laws \\
  & Object Permanence   & Objects maintain identity when reappearing after occlusion \\
  & Human Fidelity      & Humans rendered without anatomical artifacts or impossible poses \\
  & Dynamic Degree      & Actual object/character motion (not just camera movement or static scene) \\
\midrule
\multirow{3}{*}{C: Instruction Adherence}
  & Semantic Adherence   & Video contains exactly what was requested with correct attributes \\
  & Spatial Relationship & Objects positioned correctly relative to each other as specified \\
  & Semantic Drift       & Absence of unintended deviation from the prompt concept \\
\midrule
\multirow{4}{*}{D: Aesthetic Quality}
  & Composition \& Framing  & Shot composition, rule-of-thirds, focal interest, visual balance \\
  & Lighting \& Volumetric  & Lighting quality, volumetric effects, shadow coherence \\
  & Color Harmony           & Colour palette coherence, tonal balance, aesthetic saturation \\
  & Structural Gestalt      & Overall visual coherence, depth, foreground/background separation \\
\bottomrule
\end{tabular}
\end{table}
\subsection{VidProM Filtering Pipeline}
\label{app:filtering}

The filtering pipeline applies the following sequential stages to the $\sim$1.7M
VidProM prompts:
\begin{enumerate}[leftmargin=1.5em,topsep=2pt,itemsep=0pt]
  \item \textbf{NSFW/safety filter:} VidProM's built-in classifier removes sexually
    explicit, violent, and hateful content.
  \item \textbf{Deduplication:} Exact-hash deduplication followed by MinHash/LSH near-duplicate
    removal with a Jaccard threshold of 0.8.
  \item \textbf{Length filter:} Prompts $< 30$ characters or $> 500$ characters are removed.
  \item \textbf{Complexity score:} An LLM-estimated score rewards prompts involving
    physics interactions, multi-subject scenes, temporal events, and spatial relationships.
    Prompts in the bottom quartile are discarded.
  \item \textbf{Blacklist:} Prompts containing URLs, political figures, named celebrities,
    or trademarked properties are flagged and removed.
  \item \textbf{Spam detection:} Repetitive, malformed, or auto-generated prompts are
    removed via an n-gram-based classifier.
\end{enumerate}
These stages retain approximately 5,000 prompts ($\sim$0.3\% of the original corpus).
Subsequent LLM judging further flags 276 (7.4\%) for copyright/safety review, yielding
the final set of 3,754 unique prompts.
\FloatBarrier

\subsection{Prompt Enhancement}
\label{sec:enhancement}

\paragraph{Targeted enhancement.}
For each dimension where a prompt scores below 7 on suitability, we prompt Gemini~3.1~Flash-Lite
to strengthen that specific aspect while preserving the prompt's core theme, style, and
subject. Enhancement operates in two passes, the first addressing dimension weaknesses,
the second cleaning grammar and phrasing for video model compatibility, followed by
automated re-scoring to validate improvement. The system prompt is reported in \S~\ref{app:LLMenhance}.

\begin{figure}[tbp]
  \centering
  \includegraphics[width=0.65\linewidth]{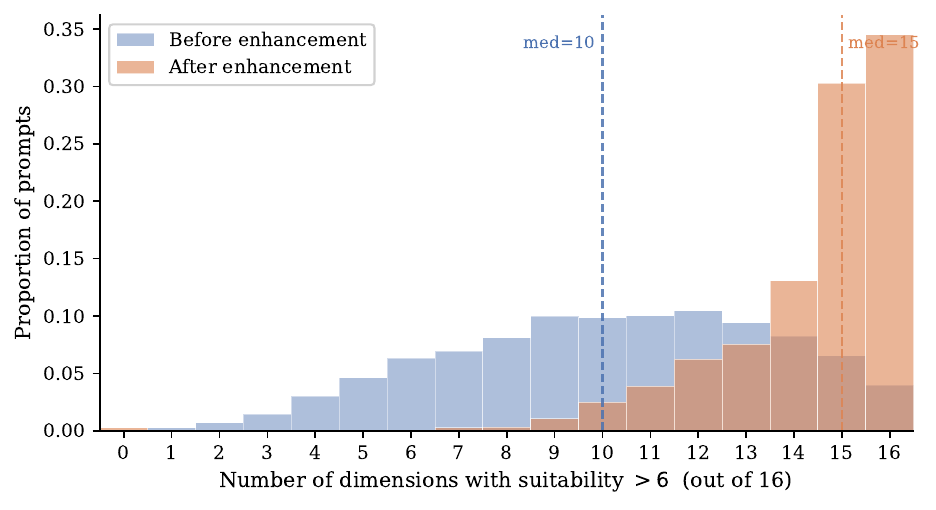}
  \caption{Distribution of prompts by number of dimensions with suitability $>6$
    (out of 16) before (blue, shaded) and after (orange) LLM enhancement.
    Before enhancement the distribution peaks around 10 dimensions (median=10);
    after enhancement it shifts decisively right (median=15), indicating near-complete
    multi-dimensional coverage for most prompts.}
  \label{fig:coverage}
\end{figure}

\paragraph{Enhancement statistics.}
Table~\ref{tab:enhancement} shows the before/after \emph{suitability} and
\emph{difficulty} scores for all 16 dimensions.
Suitability measures how well a prompt exercises a given dimension;
difficulty measures how hard that dimension is to execute correctly.
Enhancement raises both simultaneously: suitability improves the average scores by $+1.10$
(7.30 $\to$ 8.40) and difficulty by $+1.05$ (6.44 $\to$ 7.49), confirming
that enhanced prompts are not merely more appropriate but also more demanding.
The largest suitability gains occur in \textit{Object Permanence}
($+2.46$, 5.62 $\to$ 8.07) and \textit{Inertial Consistency}
($+2.17$, 6.38 $\to$ 8.56), accompanied by difficulty gains of
$+2.00$ and $+1.87$ respectively. Thus these represent the two dimensions most underspecified in
original samples from VidProM.
The smallest gains in both metrics are in \textit{Semantic Adherence}
(suitability $+0.28$, difficulty $+0.61$) and
\textit{Human Fidelity} (suitability $+0.37$, difficulty $+0.43$),
both already near saturation before enhancement. Human-authored prompts
are naturally specific about subjects and content, leaving little room
for the LLM to add value on these axes. 
Group~B (Logic \& Physics) achieves the largest absolute group gain
($+1.35$ suitability, $+1.21$ difficulty), consistent with the A1 ablation
finding that enhanced prompts most strongly stress physics and structural
generation capabilities (\S~\ref{sec:abl_enhance}).
The distribution shift in multi-dimensional coverage at threshold $k=6$ is
shown in Figure~\ref{fig:coverage}. Thus there are multiple possibilities to either test 
the entirety of the spectrum of dimensions or a smaller limited set. An example comparison of
unenhanced vs.\ enhanced prompts and their scores are provided in
Appendix~\S~\ref{app:prompt_example_clean}.

\begin{table}[t]
\centering
\caption{Per-dimension \textbf{suitability} and \textbf{difficulty} scores before and after
  LLM enhancement (1--10 scale, mean over 3{,}754 prompts).
  \emph{Suitability} measures how well the prompt exercises the dimension;
  \emph{difficulty} measures how hard the dimension is to execute correctly given the prompt.
  Both metrics increase with enhancement, confirming that enhanced prompts are simultaneously
  more appropriate \emph{and} more demanding for each dimension.}
\label{tab:enhancement}
\small
\setlength{\tabcolsep}{4pt}
\renewcommand{\arraystretch}{1.05}
\begin{tabular}{llccc@{\hskip 10pt}ccc}
\toprule
 & & \multicolumn{3}{c}{\textbf{Suitability}} & \multicolumn{3}{c}{\textbf{Difficulty}} \\
\cmidrule(lr){3-5}\cmidrule(lr){6-8}
\textbf{Group} & \textbf{Dimension} & \textbf{Bef} & \textbf{Aft} & \textbf{$\Delta$} & \textbf{Bef} & \textbf{Aft} & \textbf{$\Delta$} \\
\midrule
\multirow[t]{5}{*}{A: Motion \& Stability}
  & Subject Consistency   & 7.37 & 8.26 & $+0.88$ & 6.85 & 7.67 & $+0.82$ \\
  & Scene Consistency     & 7.07 & 7.92 & $+0.85$ & 6.31 & 7.08 & $+0.77$ \\
  & Motion Smoothness     & 7.79 & 8.48 & $+0.70$ & 6.82 & 7.65 & $+0.82$ \\
  & Temporal Flickering   & 7.28 & 8.81 & $+1.53$ & 6.54 & 8.01 & $+1.46$ \\
  & Inertial Consistency  & 6.38 & 8.56 & $\mathbf{+2.17}$ & 5.88 & 7.75 & $\mathbf{+1.87}$ \\
\midrule
\rowcolor{gray!10}\multirow[t]{4}{*}{B: Logic \& Physics}
  & Physical Mechanics  & 6.71 & 8.36 & $+1.65$ & 6.49 & 7.85 & $+1.37$ \\
\rowcolor{gray!10}
  & Object Permanence   & 5.62 & 8.07 & $\mathbf{+2.46}$ & 5.33 & 7.33 & $\mathbf{+2.00}$ \\
\rowcolor{gray!10}
  & Human Fidelity      & 8.57 & 8.94 & $+0.37$ & 8.05 & 8.48 & $+0.43$ \\
\rowcolor{gray!10}
  & Dynamic Degree      & 7.83 & 8.73 & $+0.90$ & 6.64 & 7.66 & $+1.02$ \\
\midrule
\multirow[t]{3}{*}{C: Instruction Adh.}
  & Semantic Adherence  & 8.67 & 8.96 & $+0.28$ & 7.09 & 7.71 & $+0.61$ \\
  & Spatial Relationship& 7.23 & 8.31 & $+1.08$ & 6.21 & 7.30 & $+1.09$ \\
  & Semantic Drift      & 6.50 & 7.48 & $+0.98$ & 5.51 & 6.43 & $+0.93$ \\
\midrule
\multirow[t]{4}{*}{D: Aesthetic Quality}
  & Composition \& Framing & 7.50 & 8.32 & $+0.82$ & 6.21 & 6.90 & $+0.70$ \\
  & Lighting \& Volumetric & 7.82 & 9.05 & $\mathbf{+1.23}$ & 6.96 & 8.13 & $\mathbf{+1.16}$ \\
  & Color Harmony          & 7.06 & 7.96 & $+0.90$ & 5.91 & 6.69 & $+0.77$ \\
  & Structural Gestalt     & 7.36 & 8.34 & $+0.98$ & 6.53 & 7.47 & $+0.94$ \\
\midrule
\multicolumn{2}{l}{\textbf{Overall (all 16 dimensions)}}
  & 7.30 & 8.40 & $\mathbf{+1.10}$
  & 6.44 & 7.49 & $\mathbf{+1.05}$ \\
\bottomrule
\end{tabular}
\end{table}

\FloatBarrier
% \iffalse % System prompts commented out; available in code release
\subsection{System Prompts}
\label{app:system_prompts_curation}

The following system prompts are used verbatim in the prompt curation pipeline.
Model: \texttt{gemini-3.1-flash-lite-preview} for both phases.

\subsubsection*{Phase~1 — LLM Judge (Suitability \& Difficulty Scoring)}
\label{app:LLMscoring}
\begin{small}
\begin{verbatim}
You are a video generation benchmarking expert. Analyze the prompt and
provide scores on multiple dimensions.

A prompt can test MULTIPLE groups. Determine which groups apply and score
all relevant dimensions.

**Group A: Motion & Stability**
Applicable if: prompt involves camera movement, moving subjects, or
temporal consistency challenges.

For each dimension, score BOTH suitability and difficulty (1-10 each):

1. subject_consistency: Does the main character/object change shape,
   color, or identity during the video?
   - suitability: Does this prompt create conditions where subject
     inconsistency would be exposed?
   - difficulty: How hard for a video model to keep identity consistent?

2. scene_consistency: Does the environment stay stable or warp/melt?
   - suitability: Does this prompt expose scene warping during camera
     motion?
   - difficulty: How hard to keep scene stable during camera motion?

3. motion_smoothness: Does the video have stuttering or jitter?
   - suitability: Does the prompt expose frame skips? (fast/complex
     motion scores high)
   - difficulty: How hard for a model to render this motion smoothly?
   - NOTE: Rendering quality only — not physics.

4. temporal_flickering: Are there flashes or brightness artifacts?
   - suitability: Score high only for complex textures (water, hair,
     fire, smoke, fine patterns). Intentional lighting changes are NOT
     flickering.
   - difficulty: How hard to avoid unwanted flickering?

5. inertial_consistency: Do objects follow laws of momentum?
   - suitability: Focus on velocity changes (falling, stopping,
     throwing, catching, sliding to a stop).
   - difficulty: How hard to render physically accurate inertia?
   - NOTE: Physics (velocity changes) only — not rendering smoothness.

**Group B: Logic & Physics**
Applicable if: prompt involves physical interactions, gravity,
collisions, humans/animals, or object persistence.

1. physical_mechanics: Do gravity, friction, and collisions look
   realistic?
2. object_permanence: If an object goes behind a wall, does it look
   the same when it reappears?
3. human_fidelity: Are humans rendered without alien artifacts (extra
   fingers, distorted faces, impossible body twisting)?
   Set to null if no humans in the prompt.
4. dynamic_degree: Is there actual movement, or just a still image
   with zoom?

**Group C: Instruction Adherence**
Applicable if: prompt has specific objects, colors, spatial
relationships, or precise requirements.

1. semantic_adherence: Does the video contain exactly what was asked?
2. spatial_relationship: Are objects in the right relative positions?
3. semantic_drift: Does the AI start following the prompt but "forget"
   it halfway through?

**Group D: Aesthetic Quality**
Applicable if: prompt involves specific artistic styles, high-detail
environments, or cinematic descriptions.

1. composition_framing: Is the shot well-balanced?
2. lighting_volumetric: Is the lighting realistic with depth?
3. color_harmony: Are the colors pleasing and consistent?
4. structural_gestalt: Do elements look like they belong in the same
   world?

Scoring guidelines:
- Suitability: 1 = poor test, 5 = decent test, 10 = excellent/ideal.
- Difficulty: 1 = easy for model, 5 = moderate, 10 = extremely hard.
- Set scores to null for non-applicable dimensions.
- Flag needs_review = true for harmful, policy-violating, or
  copyright-sensitive content.

Return ONLY valid JSON with all four groups in the structure shown in
the documentation. No extra text, no markdown.
\end{verbatim}
\end{small}

\subsubsection*{Phase~2 — LLM Enhancer}
\label{app:LLMenhance}
\begin{small}
\begin{verbatim}
You are a video generation prompt expert. Enhance the given video
prompt by:

1. Fixing language: Correct grammar, spelling, improve coherence.
2. Addressing weak dimensions: Add specific elements to boost weak
   dimensions (listed in the user message).
3. Preserving core theme: Keep the main subject and concept EXACTLY
   as intended.

Guidelines for weak dimensions (use specific, stress-testing details):
- motion_smoothness: Add fast/complex motion (running, spinning,
  fast-moving objects).
- temporal_flickering: Add complex textures (water, fire, hair,
  reflective surfaces). Specify high-frequency details like "individual
  water droplets", "strands of silk hair", or "fine mesh textures".
- inertial_consistency: Add velocity changes (falling, stopping,
  throwing/catching, sliding). Describe transitions such as "initial
  drag", "sudden deceleration", or "natural pendulous swing".
- physical_mechanics: Add falling, bouncing, colliding, or fluid
  interactions.
- object_permanence: Add occlusion (objects going behind/under things).
- human_fidelity: Add close-ups of hands/faces, complex poses (only
  if humans already in prompt). Focus on anatomical stress points:
  "knuckle articulation during gripping", "subtle facial muscle
  movements", "realistic foot-to-ground contact".
- dynamic_degree: Add state transformations (melting, dancing,
  running) not just camera movement.
- semantic_adherence: Add specific colors, attributes, objects.
- spatial_relationship: Specify precise positions (on top of, inside,
  holding, under).
- semantic_drift: Add multi-stage or sustained activities.
- composition_framing: Specify shot type (close-up, wide, POV).
- lighting_volumetric: Add lighting detail (neon, sunset, volumetric
  fog, multiple lights, shadows).
- color_harmony: Add color palette (monochromatic, vibrant, pastel).
- structural_gestalt: Add detail for unified elements (consistent
  style, coherent world).

Constraints:
- DO NOT add new characters/entities not in the original.
- Only enhance elements already present or naturally implied.
- Keep it concise (3-5 sentences, under 1000 characters).

Return a JSON object with exactly one key: "enhanced_prompt" (string).
Example: {"enhanced_prompt": "Your enhanced prompt text here."}
\end{verbatim}
\end{small}
% \fi % end of commented-out Phase A system prompts

% ══════════════════════════════════════════════════════════════════════════════
%\FloatBarrier
\section{VLM Questionnaire}
\label{app:vlm}
% ══════════════════════════════════════════════════════════════════════════════

% \iffalse % System prompts commented out ; available in code release
\subsection{System Prompts}
\label{app:system_prompts_vlm}

\subsubsection*{VQA Question Generator}

Model: \texttt{gemini-3-flash-preview}.
Dimensions with \texttt{null} suitability for a given prompt are omitted entirely.
The call is split by group (A–D); each call sends the system prompt below plus a
user message of the form:

\begin{small}
\begin{verbatim}
PROMPT: {enhanced_prompt_text}

DIMENSIONS TO ANALYZE:
- {dimension_name}: {dimension_definition}
...

Generate 10 questions per dimension.
\end{verbatim}
\end{small}

\noindent where each \texttt{\{dimension\_definition\}} is drawn from the following
table (reproduced verbatim from the source code):

\begin{small}
\begin{tabular}{@{}lp{0.72\linewidth}@{}}
\toprule
\textbf{Dimension} & \textbf{Definition passed to the model} \\
\midrule
subject\_consistency   & Does the main character/object change shape, color, or identity during the video? \\
scene\_consistency     & Does the environment (trees, buildings, background) stay stable or ``warp''/``melt'' as the camera moves? \\
motion\_smoothness     & Does the video have ``stuttering,'' ``jitter,'' or frames that look like they're skipping? \\
temporal\_flickering   & Are there flashes of light or sudden brightness changes (unwanted flickering/artifacts)? \\
inertial\_consistency  & Do objects follow the laws of momentum --- speeding up and slowing down naturally? \\
physical\_mechanics    & Do gravity, friction, and collisions look realistic? \\
object\_permanence     & If an object goes out of view or behind a wall, does it look exactly the same when it reappears? \\
human\_fidelity        & Are humans rendered without ``alien'' artifacts like extra fingers, distorted faces, etc.? \\
dynamic\_degree        & Is there actual movement, or is it just a still image with zoom? \\
semantic\_adherence    & Does the video contain exactly what was asked for in the prompt? \\
spatial\_relationship  & Are objects in the right place relative to each other? \\
semantic\_drift        & Does the AI start following the prompt but ``forget'' it and change the scene halfway through? \\
composition\_framing   & Is the shot well-balanced, or does it feel like a random crop? \\
lighting\_volumetric   & Is the lighting realistic with depth, or does it look flat and ``CGI-like''? \\
color\_harmony         & Are the colors pleasing and consistent, or is there ``digital bleeding''? \\
structural\_gestalt    & Do the elements look like they belong in the same world, or like stickers pasted on? \\
\bottomrule
\end{tabular}
\end{small}

\medskip
\noindent\textbf{System prompt:}

\begin{small}
\begin{verbatim}
You are a video generation benchmarking expert. Your task is to generate
10 unique, probing VQA (Video Question Answering) questions for EACH of
the dimensions listed below, specifically for a video generated from the
provided PROMPT.

For each DIMENSION of the PROMPT:
1. Generate 10 questions that specifically probe that dimension as it
   relates to this prompt.
2. Questions should cover:
   - Expected events and details mentioned in the prompt.
   - Potential failure modes (e.g., "Does the character's face distort
     when they turn?").
   - Success modes (e.g., "Is the reflection on the water consistent
     with the light source?").
   - Adversarial probing (checking for subtle inconsistencies).
3. For each question, define a 1-5 scoring rubric:
   - 1: Major failure / Completely incorrect.
   - 2: Notable artifacts / Significant issues.
   - 3: Mediocre / passable but flawed.
   - 4: Good / minor imperfections only.
   - 5: Perfect / Flawless execution.

Return ONLY a JSON object where keys are the dimension names and values
are lists of 10 question objects. Each question object must have
"question" and "rubric_description".
\end{verbatim}
\end{small}

\subsubsection*{VLM Evaluator}

Model: \texttt{gemini-3-flash-preview}.
Frames are extracted per the dimension-aware sampling strategy
and passed alongside the prompt below.
The user message is constructed per dimension, per video.

\begin{small}
\begin{verbatim}
You are evaluating a video generated from a prompt.
Dimension: {dimension_name}

Questions to answer:
1. {question_1}
   Rubric: {rubric_description_1}
2. {question_2}
   Rubric: {rubric_description_2}
...
10. {question_10}
    Rubric: {rubric_description_10}

Answer each question with a score (1-5) and a short justification.
Return ONLY a JSON list of objects:
[{"score": X, "justification": "..."}, ...]
\end{verbatim}
\end{small}
% \fi % end of commented-out Phase B system prompts

% ══════════════════════════════════════════════════════════════════════════════
%\FloatBarrier
\section{Case Studies}
\label{app:casestudies}
% ══════════════════════════════════════════════════════════════════════════════

This appendix presents two contrasting case studies.
\textbf{Prompt\_1732} (\S~\ref{app:casestudy_clean}) is a \emph{clean example}:
both judges (Gemini/Claude) agree, rankings mirror the global BT ranking order, and prompt enhancement
substantially raises physics suitability without unlocking edge-case artifacts. \textbf{Prompt\_1102} (\S~\ref{app:casestudy_hallucination}) is a \emph{hallucination
example}: the VLM judges produce inverted Inertial Consistency rankings,
providing a concrete instance of dimension-level VLM over-rating
(\S~\ref{sec:abl_reliability}).

\subsection{Clean Evaluation Example: Prompt\_1732}
\label{app:casestudy_clean}
% ─────────────────────────────────────────────────────────────────────────────

Prompt\_1732 (humanoid dragon lord of terror) illustrates a well-functioning
evaluation: Veo~3.1 ranks first in both Gemini (mean~4.68) and Claude (mean~3.52),
Inertial Consistency is consistently ordered across both judges (Veo~3.1: 4.4 / 3.3),
and the ranking Veo $>$ Kling $\approx$ Wan~A14B $>$ LTX-2 $>$ Hunyuan $\approx$
Wan~1.3B closely mirrors the global BT rating leaderboard.
No anomalous over-rating is observed: the weaker models (Hunyuan, Wan~1.3B) score
at the bottom on both judges for IC, Physical Mechanics, and overall mean.

\subsubsection{Prompt and Enhancement}
\label{app:prompt_example_clean}

\paragraph{Original prompt (prompt\_1732):}
\emph{``A humanoid dragon, lord of terror, full body, blue and red fire color,
magic temple background, futuristic armor, holding magic sword, long shot.''}

\paragraph{Enhanced prompt:}
\emph{``A full-body shot of a hulking humanoid dragon lord of terror, clad in sleek
futuristic armor that reflects the swirling blue and red inferno engulfing his form.
As he lunges forward, his massive frame exhibits natural weight and momentum with a
sudden, heavy stop before he slashes a glowing magic sword, causing the blade to
collide and spark against the stone floor of a crumbling magic temple. The blue and
red flames dance with intricate, shimmering detail, casting chaotic, realistic shadows
that shift dynamically as he moves behind and around broken temple pillars, remaining
consistently visible and correctly obscured as he weaves through the environment
throughout the intense action sequence.''}

\paragraph{Dimension gains (prompt\_1732):}
The largest gains are Inertial Consistency ($5 \to 10$, $+5$),
Object Permanence ($5 \to 10$, $+5$), and Physical Mechanics ($6 \to 10$, $+4$),
reflecting the explicit addition of weight, momentum, and object-collision cues in
the enhanced prompt.
Full per-dimension scores are in Table~\ref{tab:enhancement_prompt1732}.

\begin{table}[h]
\centering
\caption{Per-dimension suitability and difficulty before/after enhancement for
  \textbf{prompt\_1732}. All groups were applicable before and after;
  no group unlocking occurs.}
\label{tab:enhancement_prompt1732}
\footnotesize
\begin{tabular}{llcccc}
\toprule
\multirow{2}{*}{\textbf{Grp}} & \multirow{2}{*}{\textbf{Dimension}} &
  \multicolumn{2}{c}{\textbf{Suitability}} & \multicolumn{2}{c}{\textbf{Difficulty}} \\
  \cmidrule(lr){3-4}\cmidrule(lr){5-6}
  & & Bef & Aft ($\Delta$) & Bef & Aft ($\Delta$) \\
\midrule
A & Subject Consistency  & 9 & 9\hfill$(0)$            & 9 & 9\hfill$(0)$ \\
A & Scene Consistency    & 7 & 9\hfill$(+2)$           & 7 & 9\hfill$(+2)$ \\
A & Motion Smoothness    & 6 & 9\hfill$(+3)$           & 6 & 8\hfill$(+2)$ \\
A & Temporal Flickering  & 9 & 9\hfill$(0)$            & 9 & 9\hfill$(0)$ \\
A & Inertial Consistency & 5 & 10\hfill$\mathbf{(+5)}$ & 6 & 9\hfill$(+3)$ \\
\midrule
B & Physical Mechanics   & 6 & 10\hfill$\mathbf{(+4)}$ & 7 & 9\hfill$(+2)$ \\
B & Object Permanence    & 5 & 10\hfill$\mathbf{(+5)}$ & 6 & 9\hfill$(+3)$ \\
B & Human Fidelity       & 8 & 8\hfill$(0)$            & 9 & 9\hfill$(0)$ \\
B & Dynamic Degree       & 7 & 10\hfill$(+3)$          & 7 & 9\hfill$(+2)$ \\
\midrule
C & Semantic Adherence   & 9 & 9\hfill$(0)$  & 8 & 8\hfill$(0)$ \\
C & Spatial Relationship & 7 & 9\hfill$(+2)$ & 7 & 8\hfill$(+1)$ \\
C & Semantic Drift       & 6 & 9\hfill$(+3)$ & 6 & 8\hfill$(+2)$ \\
\midrule
D & Composition \& Framing & 7 & 8\hfill$(+1)$  & 6 & 7\hfill$(+1)$ \\
D & Lighting \& Volumetric & 10 & 10\hfill$(0)$ & 9 & 9\hfill$(0)$ \\
D & Color Harmony          & 9 & 8\hfill$(-1)$  & 8 & 7\hfill$(-1)$ \\
D & Structural Gestalt     & 8 & 10\hfill$(+2)$ & 8 & 9\hfill$(+1)$ \\
\bottomrule
\end{tabular}
\end{table}

\subsubsection{VQA Question Set: Physical Mechanics}
\label{app:questions_clean}

For prompt\_1732 evaluated on \textbf{Physical Mechanics}:

\begin{enumerate}[leftmargin=1.5em,topsep=2pt,itemsep=2pt]
  \item \emph{Does the dragon lord show a believable deceleration (inertia) when he
    comes to the sudden, heavy stop after his lunge?}
  \item \emph{When the sword collides with the stone floor, do the resulting sparks
    appear to originate from the point of contact?}
  \item \emph{Does the dragon lord's weight cause any visible deformation or physical
    impact to the stone floor upon his lunge and stop?}
  \item \emph{Is there friction or slippage between the character's feet and the stone
    temple floor during the lunge movement?}
  \item \emph{Do the blue and red flames react to the dragon lord's movement as if they
    are being displaced by his body?}
  \item \emph{Is the arc of the sword slash physically consistent with the character's
    arm range and momentum?}
  \item \emph{Do the sparks dissipate and fall according to gravity, or do they linger
    unnaturally in the air?}
  \item \emph{Does the armor material look `heavy' and reactive to the sudden stop, or
    does it jitter/vibrate inappropriately?}
  \item \emph{Is the contact point of the sword blade firmly hitting the stone, or does
    it clip through the floor surface?}
  \item \emph{Do the shifting shadows accurately reflect the path of the flames and the
    character's movement relative to the pillars?}
\end{enumerate}

\subsubsection{Scoring Rubric: Physical Mechanics}
\label{app:rubric_clean}

Table~\ref{tab:pm_rubric} lists the per-question scoring rubric for the Physical
Mechanics dimension as generated by the VQA Question Generator for prompt\_1732.
Each question is scored on a 1--5 Likert scale; the rubric anchors define the two
extremes (1 = major failure, 5 = perfect execution).  These per-question scores are
averaged to produce the dimension score.

\begin{table}[h]
\centering
\caption{Per-question scoring rubric for \textbf{Physical Mechanics},
  prompt\_1732 (\emph{humanoid dragon lord, lunging halt and sword slash}).
  Score 1 = major failure; Score 5 = perfect execution.
  Intermediate scores (2--4) are interpolated by the VLM based on the
  severity of artifacts observed.}
\label{tab:pm_rubric}
\small
\setlength{\tabcolsep}{5pt}
\begin{tabular}{cllp{4.8cm}p{4.8cm}}
\toprule
\textbf{Q\#} & \textbf{Focus} & & \textbf{Score 1 (Failure)} & \textbf{Score 5 (Perfect)} \\
\midrule
1 & Dragon deceleration
  && Character stops instantly with no inertial carry-through
  & Believable heavy deceleration arc after lunge \\[2pt]
2 & Spark origin
  && Sparks appear unconnected to or away from the contact point
  & Sparks clearly emanate from the sword--floor contact point \\[2pt]
3 & Floor deformation
  && No physical impact on the stone floor; character floats
  & Visible deformation, crack, or dust response to impact \\[2pt]
4 & Foot friction / slippage
  && Feet glide frictionlessly or clip through the floor
  & Visible friction, scrape, or resistance during lunge \\[2pt]
5 & Flame displacement
  && Flames are static or move opposite to the character
  & Flames bend / displace in response to the character's motion \\[2pt]
6 & Sword arc momentum
  && Sword arc violates arm anatomy or moves with no momentum
  & Arc is anatomically consistent and momentum-driven \\[2pt]
7 & Spark gravity
  && Sparks float upward or linger unnaturally in the air
  & Sparks follow a gravity-consistent dissipation arc \\[2pt]
8 & Armor reactivity
  && Armor looks weightless; jitters or teleports on sudden stop
  & Armor looks heavy and settles naturally after the stop \\[2pt]
9 & Sword--floor contact
  && Sword visibly clips through or hovers above the floor surface
  & Sword makes firm, clean contact with the floor surface \\[2pt]
10 & Shadow accuracy
  && Shadows are absent, frozen, or inconsistent with light sources
  & Shadows accurately track flame position and character movement \\
\bottomrule
\end{tabular}
\end{table}

\subsubsection{Model Scores}
\label{app:casestudy_clean_scores}

\begin{figure}[tbp]
  \centering
  \includegraphics[width=\linewidth]{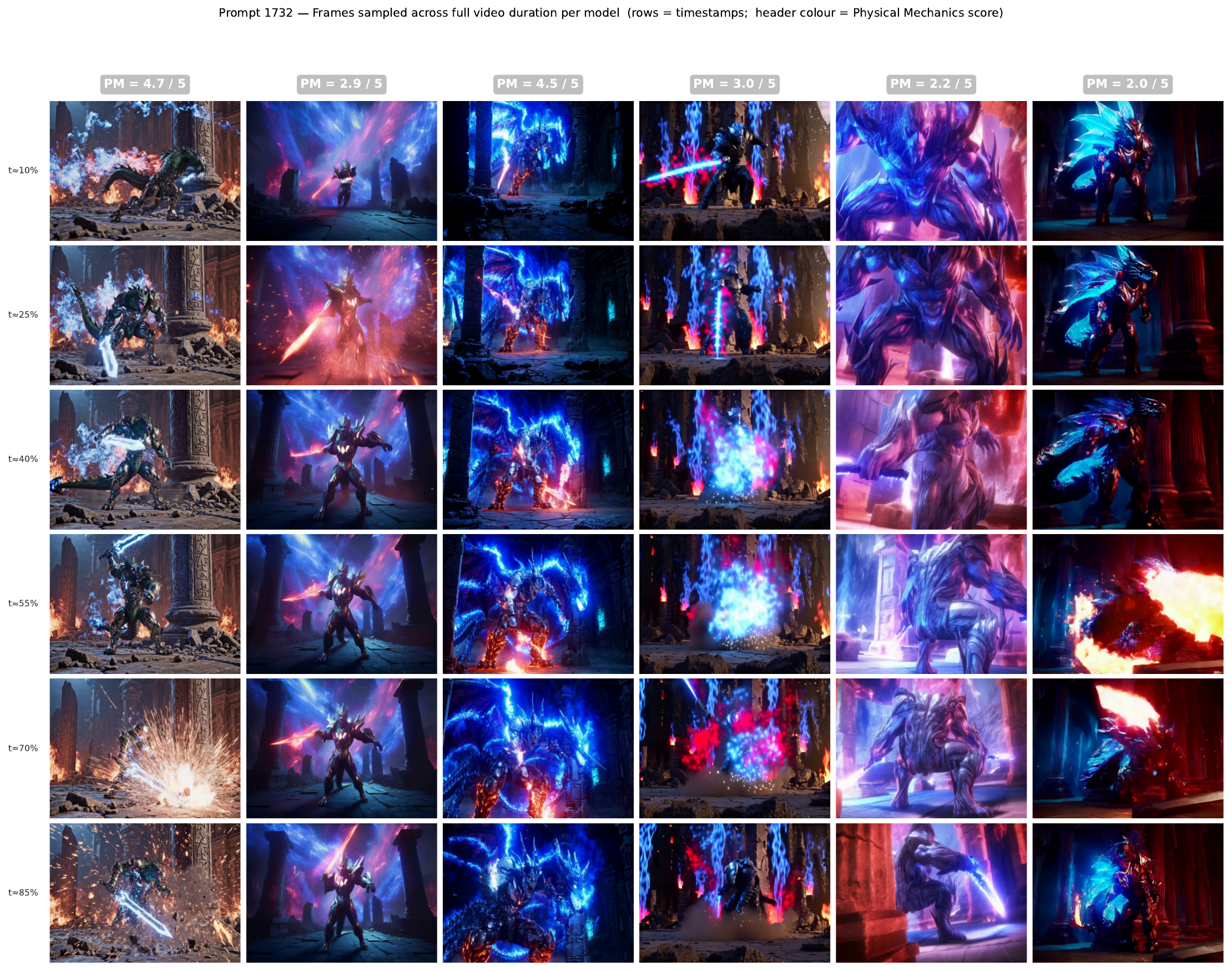}
  \caption{Six frames sampled uniformly across the full video duration
    ($t\approx10\%$--$85\%$) for each of the six evaluated models on prompt\_1732
    (\emph{humanoid dragon lord, lunging halt and sword slash}).
    \textbf{Columns} = models ordered by Physical Mechanics (PM) score (left-to-right);
    \textbf{rows} = timestamps capturing the action arc:
    approach ($t\approx10\%$), mid-lunge, impact, post-impact, hold, and settle
    ($t\approx85\%$).
    Column headers are colour-coded by the model's mean PM score (1--5,
    RdYlGn scale; red = poor, green = excellent).
    \veo\ (PM~4.7) and \wanb\ (PM~4.5) show the most physically coherent
    sword-collision and flame-displacement; \wans\ (PM~2.0) and \hunyuan\ (PM~2.2)
    exhibit the weakest physical grounding, consistent with their lower BT rating.}
  \label{fig:dragon_grid}
\end{figure}

\begin{figure}[tbp]
  \centering
  \includegraphics[width=\linewidth]{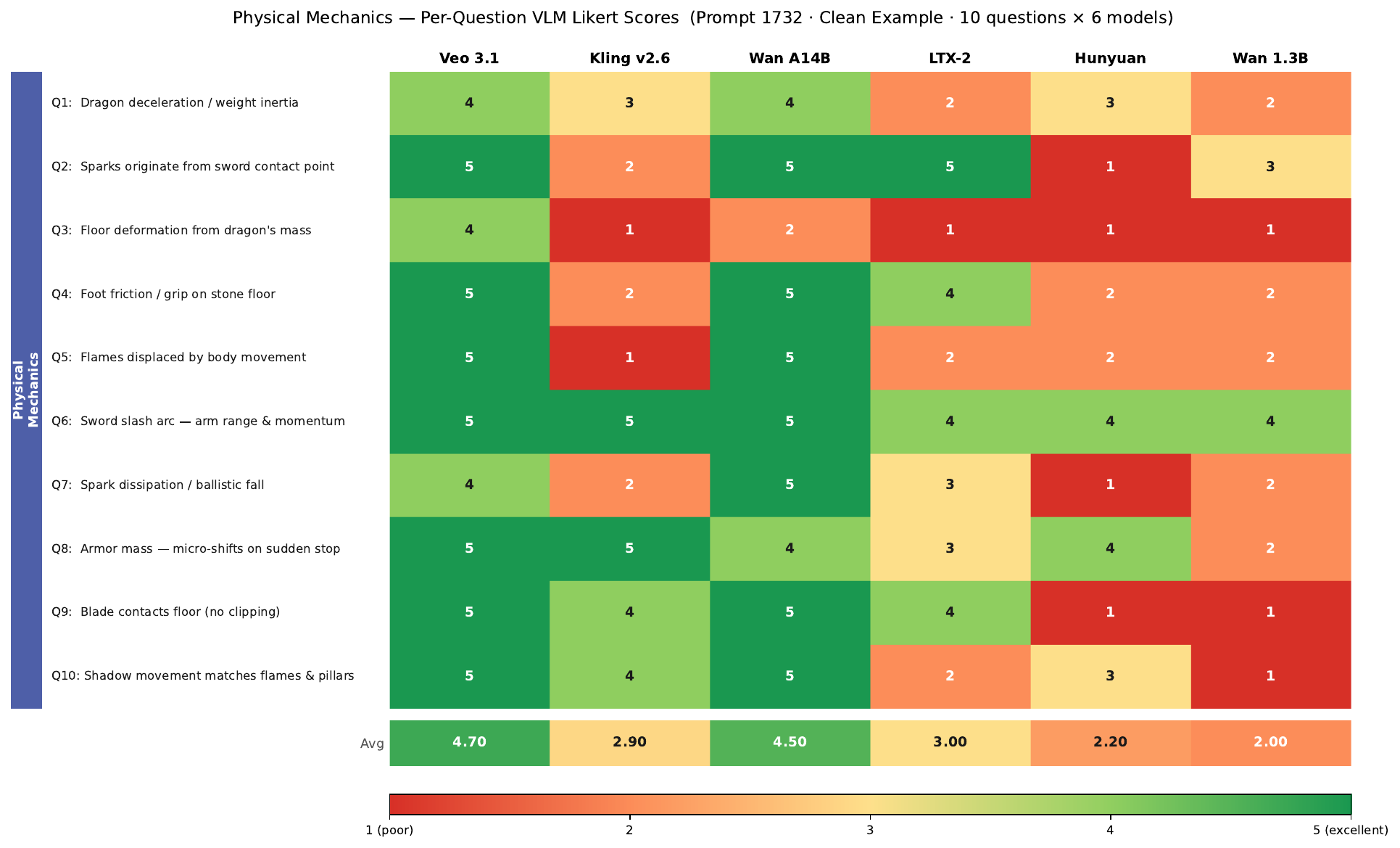}
  \caption{Per-question VLM Likert scores (1--5) for \textbf{Physical Mechanics} on
    prompt\_1732 (\emph{humanoid dragon lord, lunging halt and sword slash}),
    all six models.
    Rows correspond to the 10 questions listed in
    \S~\ref{app:questions_clean}.
    The bottom row shows the per-model average.
    Contrast with Figure~\ref{fig:assassin_scores}: here Veo~3.1 (avg~4.70)
    leads cleanly across nearly all questions, with Wan~A14B (4.50) as a
    close second, while Hunyuan and Wan~1.3B consistently score at the bottom.
    This monotone top-to-bottom gradient mirrors the global BT ranking and
    illustrates the expected behaviour of a well-discriminating prompt.}
  \label{fig:dragon_scores}
\end{figure}

Figures~\ref{fig:dragon_grid} and~\ref{fig:dragon_scores} illustrate the scores along with the video screenshots. As can be seen from Table~\ref{tab:casestudy_clean_scores} Veo~3.1 leads Physical Mechanics in
Gemini (4.70) and the weakest models (Hunyuan, Wan~1.3B) score at the bottom in both. The Gemini per-question scores show a clear monotone gradient from Veo~3.1
(mostly 4--5) down to Wan~1.3B (mostly 1--2), consistent with the global BT ranking.

\begin{table}[h]
\centering
\caption{Per-model scores (1--5 Likert) for selected dimensions on prompt\_1732.
  \textbf{G} = Gemini~3~Flash, \textbf{C} = Claude Sonnet.
  Models ordered by global VLM BT ranking. Both judges agree: Veo~3.1 leads IC,
  Hunyuan and Wan~1.3B are consistently weakest.}
\label{tab:casestudy_clean_scores}
\small
\setlength{\tabcolsep}{4pt}
\begin{tabular}{lcccccccc}
\toprule
 & \multicolumn{2}{c}{\textbf{Inertial Cons.}} & \multicolumn{2}{c}{\textbf{Phys.\ Mech.}}
 & \multicolumn{2}{c}{\textbf{Object Perm.}} & \multicolumn{2}{c}{\textbf{Sem.\ Adher.}} \\
\cmidrule(lr){2-3}\cmidrule(lr){4-5}\cmidrule(lr){6-7}\cmidrule(lr){8-9}
\textbf{Model} & G & C & G & C & G & C & G & C \\
\midrule
Veo 3.1   & \textbf{4.4} & \textbf{3.3} & \textbf{4.7} & 2.8 & \textbf{5.0} & 3.1 & \textbf{5.1} & \textbf{4.4} \\
Kling     & 3.9 & 2.9 & 2.9 & 2.6 & 3.7 & 2.8 & 4.1 & 3.9 \\
Wan A14B  & 3.7 & 2.4 & 4.5 & 2.7 & \textbf{5.0} & 2.9 & 4.6 & 3.8 \\
LTX-2     & 3.4 & 2.6 & 3.0 & 2.5 & \textbf{5.0} & 2.5 & \textbf{5.0} & 3.8 \\
Hunyuan   & 3.1 & 2.5 & 2.2 & 2.4 & 4.5 & \textbf{3.0} & 4.0 & 3.2 \\
Wan 1.3B  & 3.9 & 2.7 & 2.0 & 2.6 & 2.2 & 2.1 & \textbf{5.0} & 3.4 \\
\bottomrule
\end{tabular}
\end{table}

% ══════════════════════════════════════════════════════════════════════════════
\FloatBarrier
% ─────────────────────────────────────────────────────────────────────────────
\subsection{Hallucination Example: Prompt\_1102}
\label{app:casestudy_hallucination}
% ─────────────────────────────────────────────────────────────────────────────
\FloatBarrier
Prompt\_1102 is selected because the two VLM judges (Gemini/Claude) produce diametrically opposed Inertial Consistency rankings: Gemini assigns the highest IC to LTX-2 and Wan~1.3B
(both 4.5) while ranking Veo~3.1 last (2.7), whereas Claude ranks Veo~3.1 first
(3.2) and Wan~A14B last (1.4).  The rank correlation between judges on IC for this
prompt is $\rho=-1.0$ (perfectly inverted). 

\subsubsection{Prompt and Enhancement}
\label{app:prompt_example}

\paragraph{Original prompt (prompt\_1102):}
\emph{``A female assassin riding a lavender horse through a fairytale medieval forest.''}

\paragraph{Enhanced prompt:}
\emph{``A cinematic wide shot of a female assassin riding a lavender horse at a full gallop
through a lush, fairytale medieval forest. As the horse abruptly decelerates into a sudden
sliding halt, the assassin's long black hair whips dynamically in the wind, catching
volumetric sunlight filtering through the dense canopy. Her form remains perfectly consistent
as she dips behind thick, moss-covered oak trees, staying correctly hidden from view as she
passes behind the trunks, while the horse's hooves kick up realistic clumps of mud and
grass that collide with the environment. The scene is bathed in golden-hour lighting,
creating sharp, realistic shadows that ground the characters within the unified,
high-fantasy landscape.''}

\paragraph{Dimension gains (prompt\_1102):}
Enhancement unlocks the Logic \& Physics group entirely for this prompt: prior to
enhancement, Physical Mechanics and Object Permanence had suitability scores of 5 and 4
respectively, below the applicable threshold, so the group was excluded from evaluation.
Post-enhancement both reach 10.
The largest individual gains are Object Permanence ($4 \to 10$, $+6$), Physical Mechanics
($5 \to 10$, $+5$), and Inertial Consistency ($6 \to 10$, $+4$).
Full per-dimension scores are in Table~\ref{tab:enhancement_prompt1102}.

\begin{table}[h]
\centering
\caption{Per-dimension suitability and difficulty before/after enhancement for
  \textbf{prompt\_1102}. Group~B was not applicable pre-enhancement;
  $\dagger$ marks the newly unlocked group.}
\label{tab:enhancement_prompt1102}
\footnotesize
\begin{tabular}{llcccc}
\toprule
\multirow{2}{*}{\textbf{Grp}} & \multirow{2}{*}{\textbf{Dimension}} &
  \multicolumn{2}{c}{\textbf{Suitability}} & \multicolumn{2}{c}{\textbf{Difficulty}} \\
  \cmidrule(lr){3-4}\cmidrule(lr){5-6}
  & & Bef & Aft ($\Delta$) & Bef & Aft ($\Delta$) \\
\midrule
A & Subject Consistency  & 9 & 9\hfill$(0)$             & 8 & 8\hfill$(0)$ \\
A & Scene Consistency    & 7 & 9\hfill$(+2)$            & 6 & 8\hfill$(+2)$ \\
A & Motion Smoothness    & 8 & 9\hfill$(+1)$            & 7 & 8\hfill$(+1)$ \\
A & Temporal Flickering  & 8 & 8\hfill$(0)$             & 7 & 7\hfill$(0)$ \\
A & Inertial Consistency & 6 & 10\hfill$\mathbf{(+4)}$  & 6 & 9\hfill$(+3)$ \\
\midrule
B$^\dagger$ & Physical Mechanics & 5 & 10\hfill$\mathbf{(+5)}$ & 5 & 9\hfill$(+4)$ \\
B$^\dagger$ & Object Permanence  & 4 & 10\hfill$\mathbf{(+6)}$ & 5 & 9\hfill$(+4)$ \\
B$^\dagger$ & Human Fidelity     & 9 & 9\hfill$(0)$            & 8 & 8\hfill$(0)$ \\
B$^\dagger$ & Dynamic Degree     & 8 & 10\hfill$(+2)$          & 7 & 9\hfill$(+2)$ \\
\midrule
C & Semantic Adherence   & 9 & 9\hfill$(0)$  & 7 & 8\hfill$(+1)$ \\
C & Spatial Relationship & 8 & 9\hfill$(+1)$ & 7 & 8\hfill$(+1)$ \\
C & Semantic Drift       & 7 & 8\hfill$(+1)$ & 6 & 7\hfill$(+1)$ \\
\midrule
D & Composition \& Framing & 8 & 10\hfill$(+2)$ & 7 & 8\hfill$(+1)$ \\
D & Lighting \& Volumetric & 6 & 9\hfill$(+3)$  & 5 & 8\hfill$(+3)$ \\
D & Color Harmony          & 8 & 8\hfill$(0)$   & 7 & 7\hfill$(0)$ \\
D & Structural Gestalt     & 7 & 9\hfill$(+2)$  & 6 & 8\hfill$(+2)$ \\
\bottomrule
\end{tabular}
\end{table}

\subsubsection{VQA Question Set: Inertial Consistency}
\label{app:questions}

For prompt\_1102 evaluated on \textbf{Inertial Consistency}:

\begin{enumerate}[leftmargin=1.5em,topsep=2pt,itemsep=2pt]
  \item \emph{Rate the realism of the horse's deceleration as it transitions from a full
    gallop into the sliding halt, focusing on whether the momentum and ground-contact forces
    are consistent with the described abruptness.}
  \item \emph{Assess how accurately the assassin's hair movement reflects the inertia of
    a sudden stop---does it whip forward realistically as momentum would carry it?}
  \item \emph{Evaluate whether the mud and grass clumps kicked up by the horse's hooves
    follow believable projectile trajectories and settle at appropriate speeds.}
  \item \emph{Rate the degree to which the assassin's upper body exhibits natural
    forward lurch when the horse brakes suddenly, as Newton's First Law would predict.}
  \item \emph{Assess the temporal consistency of the transition from galloping to halted:
    does the speed ramp-down feel continuous or does it exhibit frame-skip discontinuities?}
  \item \emph{Rate whether the physical forces shown in the sliding halt---ground friction,
    weight transfer, grass displacement---are mutually consistent or contradictory.}
  \item \emph{Evaluate whether the horse's body posture during deceleration matches the
    characteristic ``haunches-down'' biomechanics of a sliding stop.}
  \item \emph{How convincingly does the horse's mane and tail continue to move forward
    after the body stops, as inertia would carry them?}
  \item \emph{Rate the plausibility of any secondary objects (saddlebags, cloak) in
    exhibiting delayed inertial motion relative to the horse's halt.}
  \item \emph{Assess the overall physical coherence of the deceleration sequence: do all
    elements (horse, rider, hair, mud, foliage) behave as a physically consistent system?}
\end{enumerate}

\subsubsection{Scoring Rubric: Inertial Consistency}
\label{app:rubric}

Table~\ref{tab:ic_rubric} lists the per-question scoring rubric for the Inertial
Consistency dimension as generated by the VQA Question Generator for prompt\_1102.
Each question is scored on a 1--5 Likert scale; the rubric anchors define the two
extremes (1 = major failure, 5 = perfect execution). 

\begin{table}[h]
\centering
\caption{Per-question scoring rubric for \textbf{Inertial Consistency},
  prompt\_1102 (\emph{female assassin on lavender horse, sliding halt}).
  Score 1 = major failure; Score 5 = perfect execution.
  Intermediate scores (2--4) are interpolated by the VLM based on the
  severity of artifacts observed.}
\label{tab:ic_rubric}
\small
\setlength{\tabcolsep}{5pt}
\begin{tabular}{cllp{4.8cm}p{4.8cm}}
\toprule
\textbf{Q\#} & \textbf{Focus} & & \textbf{Score 1 (Failure)} & \textbf{Score 5 (Perfect)} \\
\midrule
1 & Mud / hoof inertia
  && Mud stops instantly mid-air
  & Mud follows physical arc and inertia \\[2pt]
2 & Assassin body lean
  && Assassin stays rigid / unaffected by physics
  & Realistic inertial lean backwards \\[2pt]
3 & Mane \& tail delay
  && Hair stops instantly with the horse
  & Hair continues forward, then settles \\[2pt]
4 & Deceleration distance
  && Physics-defying instant stop
  & Believable slide distance \\[2pt]
5 & Leg tension \& weight
  && Legs appear weightless or stiff
  & Correct bracing posture \\[2pt]
6 & Hair whip (long black hair)
  && Hair moves the wrong direction or not at all
  & Realistic hair whip and settle \\[2pt]
7 & Mud/grass projectile path
  && Mud flies in erratic, gravity-defying paths
  & Realistic projectile physics \\[2pt]
8 & Perceived weight
  && Seems like a low-gravity environment
  & Clearly weighted characters \\[2pt]
9 & Camera follows weight
  && Camera movement ignores character physics
  & Camera supports the inertial narrative \\[2pt]
10 & Clothing lag / drag
  && Clothes stay perfectly pinned to body
  & Clothing flutters with momentum \\
\bottomrule
\end{tabular}
\end{table}

\subsubsection{Model Scores}
\label{app:casestudy}
\begin{figure}[h]
  \centering
  \includegraphics[width=\linewidth]{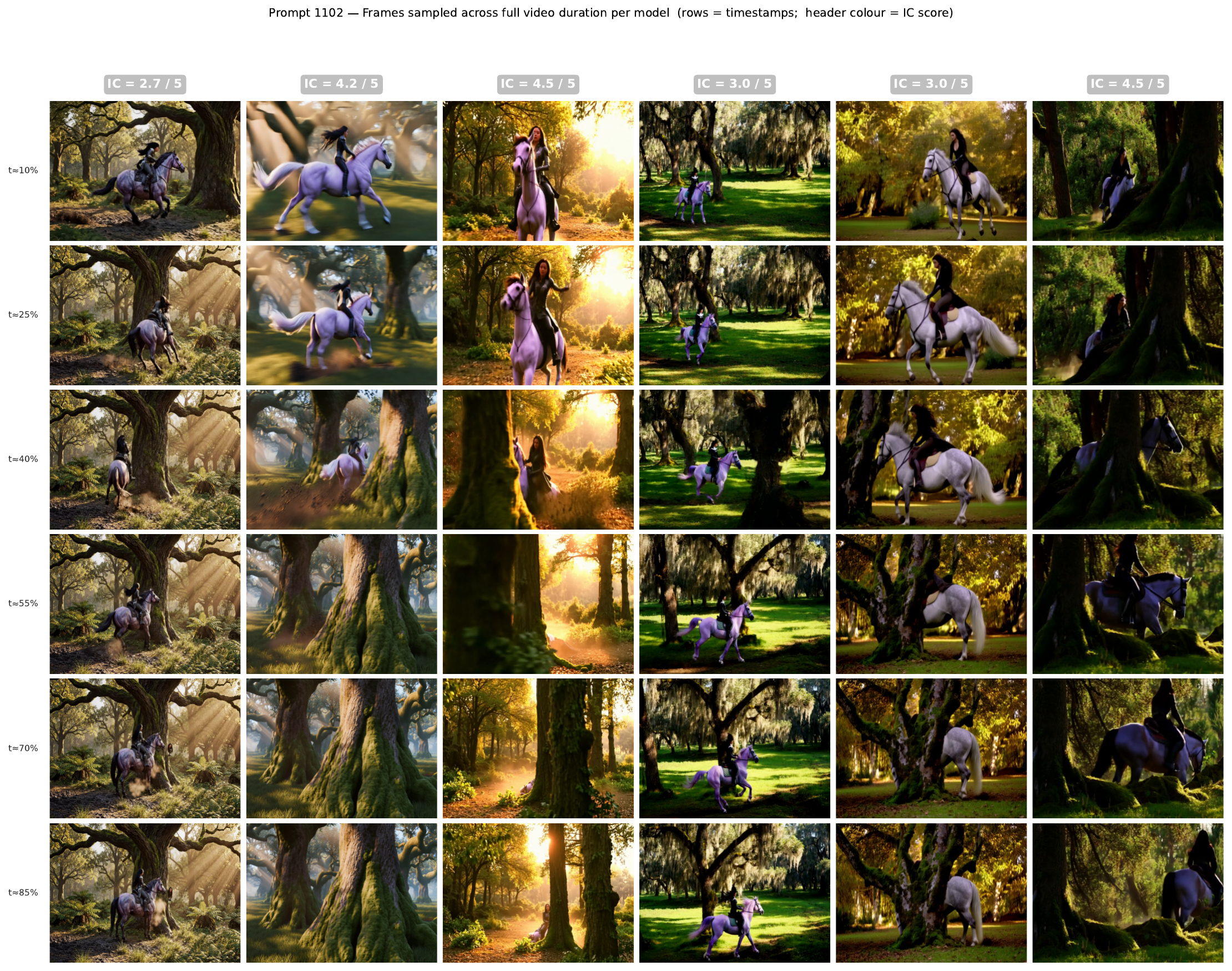}
  \caption{Six frames sampled uniformly across the full video duration
    ($t\approx10\%$--$85\%$) for each of the six evaluated models on prompt\_1102
    (assassin on horseback decelerating to a halt).
    \textbf{Columns} = models (left-to-right by VLM BT rating);
    \textbf{rows} = timestamps capturing the full deceleration arc:
    full gallop ($t\approx10\%$), early deceleration, mid-deceleration with forward
    lurch, late deceleration with haunches drop, near-halt, and post-halt settling
    ($t\approx85\%$).
    Column headers are colour-coded by the model's mean Inertial Consistency (IC)
    score (1--5, RdYlGn scale; red = poor, green = excellent).
    \ltx~and \wans~achieve the highest VLM-rated IC scores (both 4.5), followed
    by \kling~(4.2); \veo~(IC~2.7) and \hunyuan~(IC~3.0) score lower on this prompt
    despite strong overall rankings, illustrating per-prompt physics variability.
    The high IC rating for \wans, a smaller and weaker model is a notable anomaly also
    visible in Figure~\ref{fig:assassin_scores}, and represents a potential case of
    VLM over-rating; see \S~\ref{sec:abl_reliability}.}
  \label{fig:assassin_grid}
\end{figure}

\begin{figure}[h]
  \centering
  \includegraphics[width=\linewidth]{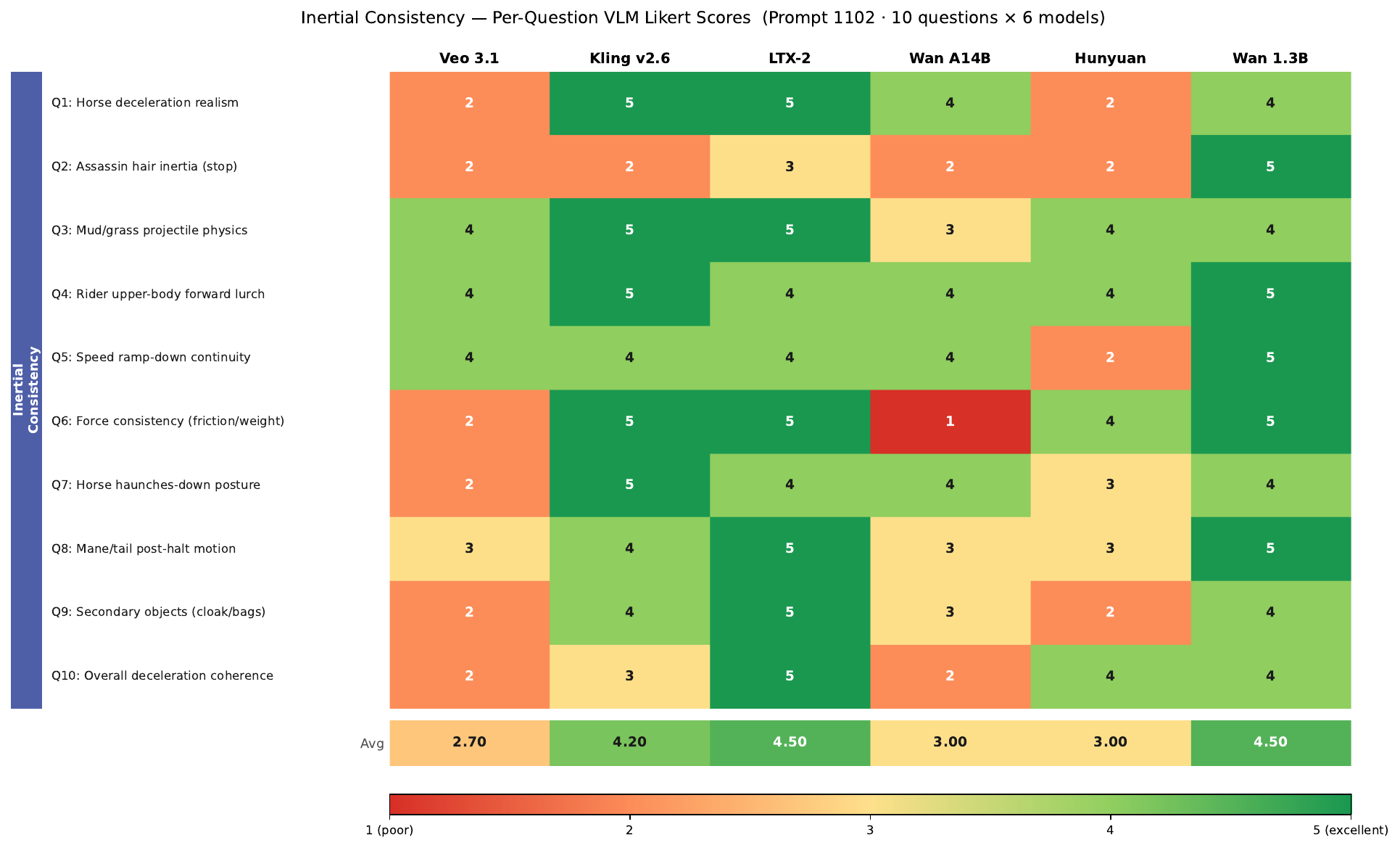}
  \caption{Per-question VLM Likert scores (1--5) for Inertial Consistency on
    prompt\_1102, all six models.
    Rows correspond to the 10 questions listed in Appendix~\ref{app:questions}.
    The rightmost column shows the per-model average.
    Notable patterns: Q2 (assassin hair inertia at stop) shows the largest
    inter-model spread, \wans~receives a 5 while all other models score 2--3,
    a likely VLM hallucination given the model's weaker overall performance.
    Q6 (force consistency) and Q10 (overall deceleration coherence) show moderate
    agreement, while Q3 (mud/grass projectile physics) exposes large variance,
    exactly the discriminative behavior the difficulty-tiered questions are
    designed to elicit.}
  \label{fig:assassin_scores}
\end{figure}

Table~\ref{tab:casestudy_hallucinated_scores} reports the head to head dimensions scores for prompt\_1102 across relevant dimensions. Figures~\ref{fig:assassin_grid} and~\ref{fig:assassin_scores} together illustrate
the full \worldjen\ evaluation pipeline on the Prompt\_1102
\begin{table}[h]
\centering
\caption{Per-model scores (1--5 Likert) for selected dimensions on prompt\_1102.
  \textbf{G} = Gemini~3~Flash, \textbf{C} = Claude Sonnet.
  Models ordered by global VLM BT ranking. Complete inversion across both judges, which is a sign for VLM hallucination.}
\label{tab:casestudy_hallucinated_scores}
\small
\setlength{\tabcolsep}{4pt}
\begin{tabular}{lcccccccc}
\toprule
 & \multicolumn{2}{c}{\textbf{Inertial Cons.}} & \multicolumn{2}{c}{\textbf{Phys.\ Mech.}}
 & \multicolumn{2}{c}{\textbf{Object Perm.}} & \multicolumn{2}{c}{\textbf{Sem.\ Adher.}} \\
\cmidrule(lr){2-3}\cmidrule(lr){4-5}\cmidrule(lr){6-7}\cmidrule(lr){8-9}
\textbf{Model} & G & C & G & C & G & C & G & C \\
\midrule
Veo 3.1   & 2.7 & \textbf{3.2} & \textbf{4.6} & \textbf{2.8} & 4.8 & 2.9 & \textbf{5.0} & \textbf{4.7} \\
Kling     & 4.2 & 2.8 & 4.5 & \textbf{2.8} & \textbf{5.0} & \textbf{3.1} & \textbf{5.0} & 4.3 \\
Wan A14B  & 3.0 & 1.4 & 3.4 & 2.1 & \textbf{5.0} & 2.9 & 3.9 & 3.1 \\
LTX-2     & \textbf{4.5} & 2.4 & 2.9 & 2.3 & 4.6 & 2.9 & 4.6 & 3.9 \\
Hunyuan   & 3.0 & 2.3 & 2.3 & 2.0 & 2.8 & 2.5 & 3.4 & 2.9 \\
Wan 1.3B  & \textbf{4.5} & 2.3 & 2.2 & 2.4 & \textbf{5.0} & 2.8 & 4.5 & 3.1 \\
\bottomrule
\end{tabular}
\end{table}

% ══════════════════════════════════════════════════════════════════════════════
\section{Ablation studies A1-A3}\label{app:ablation}
\subsection{A1 — Prompt Enhancement vs.\ No Enhancement}
\label{sec:abl_enhance}

\textbf{Motivation.}
This ablation quantifies whether LLM-based prompt enhancement changes model rankings and
what its per-dimension effect is across all 16 evaluation axes.
A central question is whether enhancement merely rewords prompts or
\emph{fundamentally raises the bar}, thereby increasing difficulty 
and exposing capability gaps that sparse, short prompts conceal.

\textbf{Method.}
We generate $20 \times 6 = 120$ new videos from the original (unenhanced)
VidProM prompts and regenerate VQA question sets from the unenhanced text.
Question regeneration is essential, using enhanced-prompt questions to
evaluate unenhanced-prompt videos would penalise models for content they
were never instructed to produce, conflating prompt quality with video quality.
The full VLM evaluation pipeline is then run identically for both conditions,
and BT ratings and per-model Likert scores are compared on the same
20 prompts.

\begin{table}[h]
\centering
\small
\caption{A1: BT rating and average Likert score with vs.\ without prompt enhancement
  (20 prompts $\times$ 6 models; BT rating re-computed on this 20-prompt subset).
  $\Delta = \text{Enhanced} - \text{Unenhanced}$.
  Rank order is identical in both conditions ($\hat{\rho}=1.000,~p=0.0014$).}
\label{tab:abl_enhance}
\setlength{\tabcolsep}{5pt}
\begin{tabular}{@{}lrrrrrrr@{}}
\toprule
\textbf{Model} & \multicolumn{3}{c}{\textbf{BT rating}} & & \multicolumn{3}{c}{\textbf{Avg Score}} \\
\cmidrule(lr){2-4}\cmidrule(lr){6-8}
 & Enh & Unenh & $\Delta$ & & Enh & Unenh & $\Delta$ \\
\midrule
\veo      & 1673.1 & 1734.1 & $-61.0$ & & 4.279 & 4.448 & $-0.169$ \\
\kling    & 1604.5 & 1629.0 & $-24.5$ & & 4.205 & 4.338 & $-0.132$ \\
\wanb     & 1518.0 & 1591.9 & $-73.9$ & & 4.085 & 4.326 & $-0.241$ \\
\ltx      & 1511.5 & 1490.8 & $+20.7$ & & 4.083 & 4.143 & $-0.060$ \\
\hunyuan  & 1405.2 & 1346.4 & $+58.8$ & & 3.924 & 3.961 & $-0.037$ \\
\wans     & 1287.7 & 1207.6 & $+80.1$ & & 3.741 & 3.765 & $-0.025$ \\
\midrule
Overall   &        &        &         & & 4.053 & 4.164 & $-0.111$ \\
\bottomrule
\end{tabular}
\end{table}

\begin{table}[h]
\centering
\small
\caption{A1: Per-model, per-dimension score difference $\Delta = \text{Enhanced} - \text{Unenhanced}$
  (pooled over 20 prompts per cell). Rows sorted by pooled $\Delta$; the rule separates
  positive from negative pooled values. \textbf{Bold} = largest magnitude per dimension.
  Per-model values can differ substantially from the pool, and individual cells should be
  read independently of the pool sign.}
\label{tab:abl_enhance_dims}
\setlength{\tabcolsep}{4pt}
\begin{tabular}{@{}lrrrrrrrr@{}}
\toprule
\textbf{Dimension} & \textbf{Pool} & \textbf{Veo} & \textbf{Kling} & \textbf{Wan v2.2} & \textbf{LTX-2} & \textbf{Hunyuan} & \textbf{Wan 1.3B} \\
\midrule
dynamic\_degree       & $+0.45$ & $+0.56$ & $+0.69$ & $+0.15$ & $+0.42$ & $+0.49$ & $+0.38$ \\
color\_harmony        & $+0.07$ & $+0.04$ & $+0.13$ & $-0.07$ & $+0.03$ & $+0.02$ & $+0.28$ \\
lighting\_volumetric  & $+0.04$ & $-0.18$ & $-0.03$ & $-0.15$ & $+0.16$ & $+0.22$ & $+0.24$ \\
subject\_consistency  & $+0.02$ & $-0.14$ & $+0.04$ & $-0.32$ & $-0.01$ & $+0.12$ & $+0.45$ \\
\midrule
composition\_framing  & $-0.05$ & $-0.23$ & $-0.09$ & $-0.06$ & $+0.12$ & $+0.21$ & $-0.26$ \\
semantic\_drift       & $-0.07$ & $+0.05$ & $-0.02$ & $-0.08$ & $-0.21$ & $-0.14$ & $-0.03$ \\
physical\_mechanics   & $-0.11$ & $+0.10$ & $\mathbf{-0.34}$ & $-0.21$ & $-0.08$ & $\mathbf{-0.49}$ & $+0.38$ \\
temporal\_flickering  & $-0.13$ & $-0.23$ & $-0.09$ & $-0.35$ & $+0.17$ & $+0.06$ & $-0.34$ \\
scene\_consistency    & $-0.13$ & $-0.03$ & $-0.13$ & $\mathbf{-0.66}$ & $-0.02$ & $-0.06$ & $+0.14$ \\
human\_fidelity       & $-0.16$ & $-0.41$ & $-0.16$ & $-0.47$ & $+0.06$ & $+0.07$ & $-0.06$ \\
inertial\_consistency & $-0.17$ & $\mathbf{-0.41}$ & $-0.35$ & $+0.03$ & $-0.27$ & $-0.32$ & $+0.28$ \\
semantic\_adherence   & $-0.19$ & $-0.11$ & $-0.20$ & $-0.01$ & $-0.36$ & $-0.32$ & $-0.14$ \\
motion\_smoothness    & $-0.23$ & $\mathbf{-0.52}$ & $-0.10$ & $-0.45$ & $-0.09$ & $+0.04$ & $-0.27$ \\
spatial\_relationship & $-0.23$ & $-0.26$ & $-0.13$ & $-0.05$ & $-0.29$ & $-0.40$ & $-0.24$ \\
object\_permanence    & $\mathbf{-0.46}$ & $-0.63$ & $\mathbf{-0.82}$ & $-0.45$ & $-0.19$ & $-0.06$ & $-0.57$ \\
structural\_gestalt   & $\mathbf{-0.49}$ & $-0.33$ & $\mathbf{-0.58}$ & $\mathbf{-0.76}$ & $-0.47$ & $-0.07$ & $\mathbf{-0.73}$ \\
\bottomrule
\end{tabular}
\end{table}

\textbf{Rank order is fully preserved.}
Spearman $\hat{\rho}=1.000,~p=0.0014$ between enhanced and un-enhanced BT ratings shows that
all six models rank identically under both conditions Table~\ref{tab:abl_enhance}, confirming that
enhancement does not introduce architectural bias.

\textbf{Enhancement is a difficulty amplifier, not a ranking modifier.}
As can be seen from table \ref{tab:abl_enhance_dims} 12 of 16 dimensions score lower under enhanced prompts and the overall mean
Likert score drops from 4.164 to 4.053 ($\Delta{=}{-0.111}$).
The effect is strongly tier-dependent: top models (Veo $\Delta{=}{-0.169}$,
Kling $\Delta{=}{-0.132}$, \wanb\ $\Delta{=}{-0.241}$) absorb the full
difficulty increase because they \emph{attempt} the richer instructions and
partially fail, while weaker models (\wans\ $\Delta{=}{-0.025}$,
\hunyuan\ $\Delta{=}{-0.037}$) under perform at both levels and are largely
unaffected. This is the behavior a discriminative benchmark should exhibit, harder prompts
widen the gap between capable and limited models.

% ─────────────────────────────────────────────────────────────────────────────
\subsection{A2 — Number of Questions per Dimension}
\label{app:abl_qcount}

\textbf{Motivation.}
We use 10 Likert questions per dimension per video.  Fewer questions would
reduce API cost linearly; more might add noise.  This ablation finds the
minimum $Q$ that preserves rank stability.

\textbf{Results.}
We subsample $Q \in \{1, 2, 3, 5, 7, 10\}$ questions per dimension per video
(200 random draws each) and recompute the model-level average score, then measure
Spearman $\hat{\rho}$ of the resulting rank order against the $Q{=}10$ gold
ranking.

\begin{table}[h]
\centering
\small
\caption{A2: Question count ablation (50 prompts $\times$ 6 models, 200 bootstrap draws).
  Mean Spearman $\hat{\rho}$ vs.\ 10-question baseline and average per-model score
  variance across bootstrap draws.}
\label{tab:abl_qcount}
\begin{tabular}{@{}ccc@{}}
\toprule
$Q$ (questions / dim) & Spearman $\hat{\rho}$ vs.\ $Q{=}10$ & Avg score variance \\
\midrule
1  & 1.0000 & 0.001243 \\
2  & 1.0000 & 0.000491 \\
3  & 1.0000 & 0.000295 \\
5  & 1.0000 & 0.000128 \\
7  & 1.0000 & 0.000053 \\
10 & 1.0000 & 0.000000 \\
\bottomrule
\end{tabular}
\end{table}

\begin{figure}[h]
  \centering
  \includegraphics[width=\linewidth]{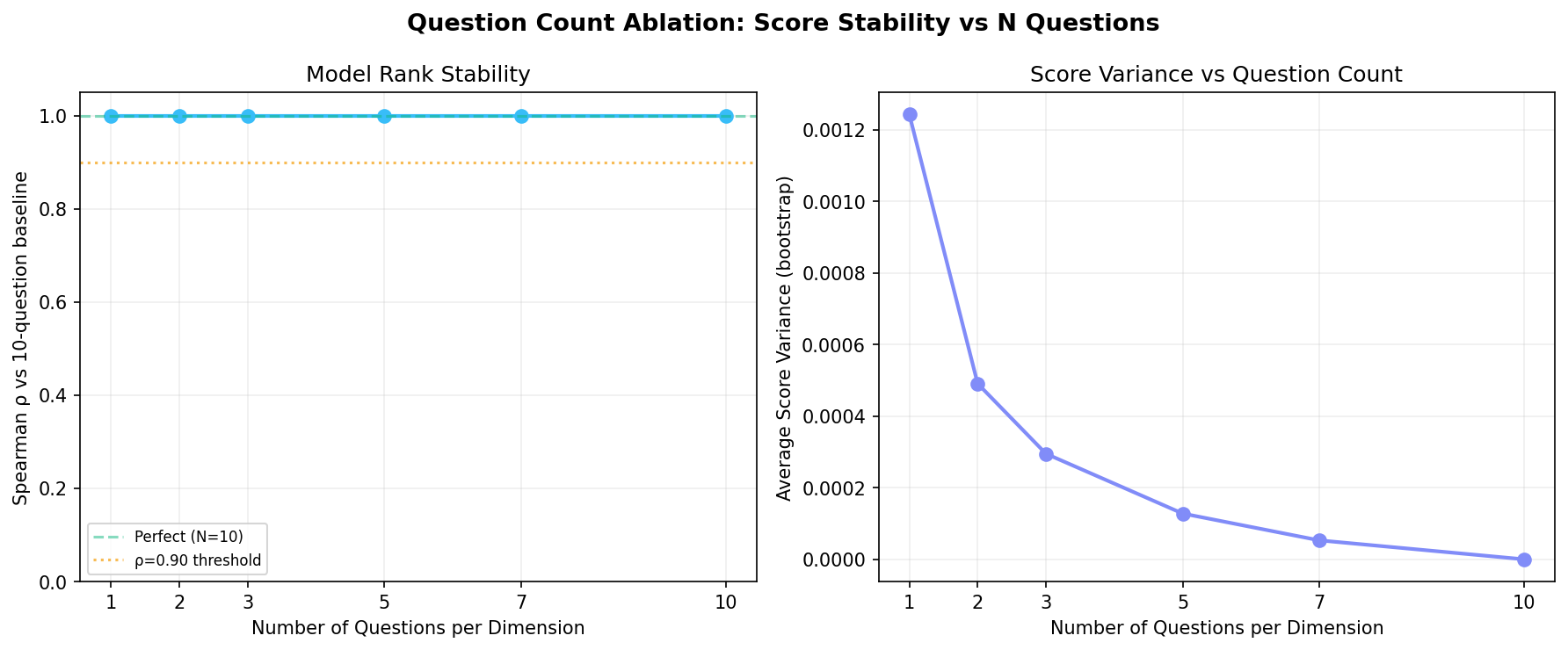}
  \caption{A2: Question count ablation (50 prompts $\times$ 6 models, 200 bootstrap draws).
    \textbf{Left:} Spearman $\hat{\rho}$ vs.\ 10-question baseline as a function of $Q$.
    Rank order is perfectly stable at all tested values of $Q$ ($\hat{\rho}=1.0$).
    \textbf{Right:} Average bootstrap score variance drops sharply with $Q$,
    falling ${\approx}23\times$ from $Q{=}1$ to $Q{=}7$.}
  \label{fig:abl_qcount}
\end{figure}

\textbf{Finding.}
Figure~\ref{fig:abl_qcount} shows rank stability (left) and score variance (right)
as a function of $Q$.  Rank order is perfectly preserved at \emph{all} tested values
of $Q$ ($\hat{\rho}=1.000,~p=0.0014$ for $Q \geq 1$), reflecting the clear
quality separation between the six models on the full 50-prompt dataset.
Score variance drops ${\approx}23\times$
from $Q{=}1$ to $Q{=}7$ (right panel), confirming that variance reduction is the
primary benefit of higher $Q$.  We retain $Q{=}10$ in the benchmark because (i) it
provides more reliable \emph{per-dimension} scores (not just overall rank),
(ii) the marginal API cost per video is small, and
(iii) the framework is designed for future evaluations where models are
more closely matched: the stability at $Q{=}1$ is a property of the
\emph{current} six-model landscape (wide tier gaps, massive prompt-level averaging),
not a general guarantee.
A cost-constrained deployment on the current model set
could safely use $Q{=}3$--$5$ without rank degradation, but we caution against
$Q{=}1$ for benchmarks where top models differ by $\leq 0.1$ Likert points,
or where fewer than 30 prompts are used.
The score-variance reduction (23$\times$ from $Q{=}1$ to $Q{=}7$) remains
the primary argument for higher $Q$ in precision-sensitive settings.

\textbf{Why is $\hat{\rho}$ so stable?}
Four structural factors jointly explain the near-universal $\hat{\rho}=1.0$:
\begin{enumerate}[topsep=2pt,itemsep=1pt]
  \item \textbf{Wide quality gap.}
    The top and bottom models (Veo~3.1 vs.\ Wan~2.1-1.3B) differ by $>$1.5 Likert
    points, a gap large enough that even very noisy single-question scores cannot
    reverse their ordering.
  \item \textbf{Massive averaging.}
    Even at $Q{=}1$, each model's final score is the mean over
    $50~\text{prompts} \times 16~\text{dimensions} = 800$ individual responses.
    The Central Limit Theorem strongly stabilises the estimate.
  \item \textbf{Score ceiling on easy dimensions.}
    Dimensions like Color Harmony (range 4.59--4.86) and Semantic Drift (4.61--4.83)
    are near ceiling for all models, contributing minimal discriminative signal
    regardless of $Q$.
\end{enumerate}

\textbf{Caveat: what would break this.}
In a future evaluation where top models score within $\pm 0.1$ of each other,
or where fewer prompts are used, or where questions are genuinely independent
(e.g.\ from different annotators), $Q{=}1$ would likely \emph{not} preserve rank.
Additional risk factors include:
\begin{itemize}[topsep=2pt,itemsep=1pt]
  \item \emph{Harder/more diverse prompts} that expose dimension-specific failures
    not captured by the current set.
  \item \emph{VLM scoring bias} (Gemini~3~Flash's generosity) compressing the
    inter-model gap, reducing the signal-to-noise ratio for low $Q$.
  \item \emph{Correlated dimensions} (e.g.\ Scene Consistency and Temporal
    Flickering both capture temporal stability) reducing the effective number of
    independent evaluation axes below 16.
\end{itemize}
We recommend $Q{\ge}5$ as a conservative default for future benchmarks with
a tighter quality band.

% ─────────────────────────────────────────────────────────────────────────────
\subsection{A3 — BT Rating Stability vs.\ Number of Prompts}
\label{app:abl_nprompts}

\textbf{Motivation.}
Is 50 prompts sufficient for a stable leaderboard, and how does rank order
evolve as $N$ grows?  This ablation characterises the minimum $N$ needed and
quantifies residual uncertainty through bootstrap confidence intervals.

\textbf{Method.}
For each $N \in \{10, 20, 30, 40, 50\}$ we draw 500 random subsets of $N$
prompts from the full 50-prompt benchmark.  For each draw, the BT rating is
recomputed from the pairwise win/loss matrix induced by the VLM Likert scores,
and we record (i) the Spearman $\hat{\rho}$ of the resulting rank vs.\ the
full-50 gold ranking and (ii) whether it exactly matches the gold rank order.
Separately, we compute 95\% bootstrap confidence intervals on each model's BT rating
by prompt-level resampling ($N{=}1000$ draws) using all 50 prompts; CI
half-widths are reported in Table~\ref{tab:leaderboard}.

\begin{table}[h]
\centering
\small
\caption{A3: BT rating stability vs.\ number of prompts (500 subsampling draws each,
  gold ranking is at N=50 full-benchmark ranking).}
\label{tab:abl_nprompts}
\begin{tabular}{@{}cccc@{}}
\toprule
$N$ prompts & Mean $\hat{\rho}$ vs.\ $N{=}50$ & 95\% CI of $\hat{\rho}$ & Perfect rank \% \\
\midrule
10 & 0.875 & [0.657, 1.000] & 13.2\% \\
20 & 0.944 & [0.829, 1.000] & 33.2\% \\
30 & 0.959 & [0.886, 1.000] & 39.0\% \\
40 & 0.972 & [0.886, 1.000] & 54.4\% \\
50 & 1.000 & [1.000, 1.000] & 100.0\% \\
\bottomrule
\end{tabular}
\end{table}

\begin{figure}[h]
  \centering
  \includegraphics[width=\linewidth]{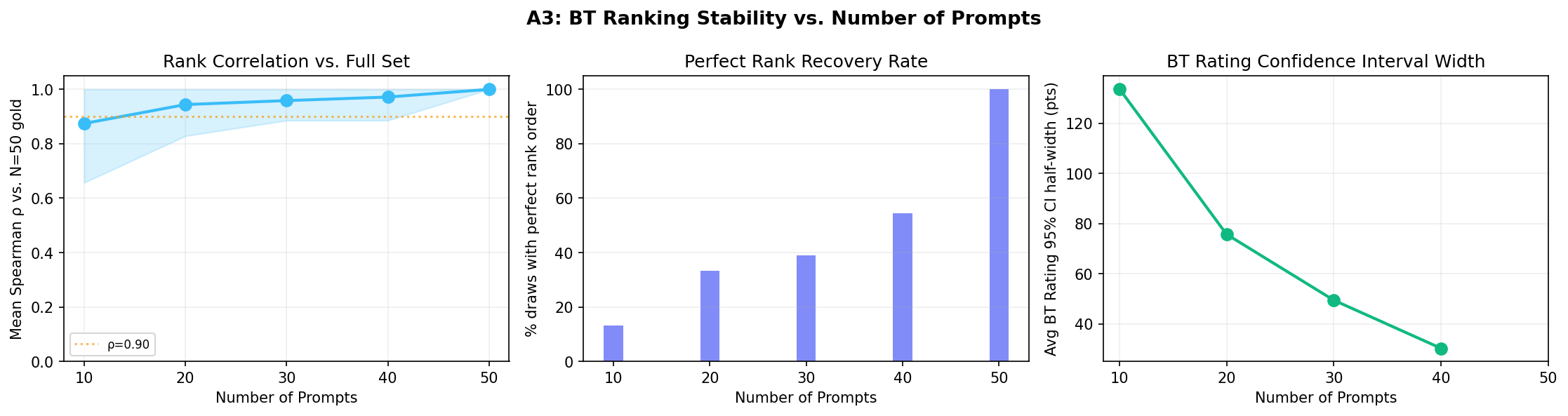}
  \caption{A3: BT rating stability vs.\ number of prompts (50-prompt benchmark, 500 draws each).
    \textbf{Left:} Mean Spearman $\hat{\rho}$ vs.\ N=50 gold ranking rises steadily with $N$,
    reaching 0.972 at $N{=}40$.
    \textbf{Centre:} Perfect rank recovery rate — 54.4\% of draws at $N{=}40$,
    100\% at $N{=}50$.
    \textbf{Right:} Average BT rating 95\% CI half-width narrows substantially as $N$ grows.}
  \label{fig:abl_nprompts}
\end{figure}

\textbf{Finding.}
Figure~\ref{fig:abl_nprompts} and Table~\ref{tab:abl_nprompts} show that rank
stability improves steadily with $N$.  At $N{=}50$ (the full benchmark),
the order is perfectly stable ($\hat{\rho}=1.0$, 100\% perfect draws).
At $N{=}40$, mean $\hat{\rho}$ is already 0.972 with a tight CI [0.886, 1.000]
and 54.4\% perfect-rank recovery.  At $N{=}20$, $\hat{\rho}=0.944$ but the CI
widens to [0.829, 1.000] and perfect rank is recovered in only 33.2\% of draws,
confirming that the current 50-prompt scale is a meaningful improvement over a
20-prompt baseline.

The 50-prompt bootstrap CI half-widths (Table~\ref{tab:leaderboard}) range from
$\pm$55 pts (Wan~1.3B) to $\pm$69 pts (Veo~3.1 Fast).
The Veo/Kling bands ([1590, 1728] vs.\ [1571, 1697]) remain overlapping,
consistent with their narrow 24-point gap,
and confirm the ranking is stable with the full 50-prompt dataset.

% ─────────────────────────────────────────────────────────────────────────────
% ─────────────────────────────────────────────────────────────────────────────

% ══════════════════════════════════════════════════════════════════════════════
%\FloatBarrier
\end{document}